\let\latexdocument\document
\let\latexenddocument\enddocument

\documentclass{clv3}

\usepackage{savesym}
\savesymbol{numdef}

\let\document\latexdocument
\let\enddocument\latexenddocument

\usepackage{url}

\usepackage{acro}

\usepackage{xcolor}

\usepackage{xspace}

\usepackage{tikz}

\usepackage{booktabs}

\usepackage{colortbl}

\usepackage{ltablex}

\usepackage{multirow}

\usepackage{multicol}

\usepackage{caption}
\usepackage{subcaption}

\usepackage{pdflscape}
\usepackage{afterpage}

\usepackage{nicefrac}

\usepackage{longtable}

\usepackage[colorinlistoftodos,prependcaption]{todonotes}

\presetkeys%
    {todonotes}%
    {inline,backgroundcolor=yellow}{}

\usepackage{siunitx}

\newcolumntype{d}{S[table-format=1.3]}

\usepackage{amssymb}
\usepackage{textcomp}
\usepackage{tipa}

\usepackage{amsmath}

\usepackage{rotating}

\usepackage{enumitem}

\usepackage{subcaption}

\usepackage{textgreek}

\usepackage{adjustbox}

\usepackage{placeins}

\definecolor{darkblue}{rgb}{0, 0, 0.5}
\usepackage{hyperref}
\hypersetup{colorlinks=true,citecolor=darkblue, linkcolor=darkblue, urlcolor=darkblue}

\usepackage[noabbrev, capitalise]{cleveref}

\makeatletter
\newcommand{\crefnames}[3]{%
    \@for\next:=#1\do{%
        \expandafter\crefname\expandafter{\next}{#2}{#3}%
    }%
}
\makeatother

\crefnames{part,chapter,section}{\S}{\S\S}

\definecolor{ao}{rgb}{0.0, 0.5, 0.0}
\definecolor{jck-gray}{gray}{0.9}

\newcommand{\jhi}[1]{\emph{#1}}
\newcommand{\jexample}[1]{\textit{#1}}
\newcommand{\jlabel}[1]{\textsc{#1}}

\newcommand{\him}[1]{\textsc{\lowercase{#1}}\xspace}  %
\newcommand{\himl}[1]{\textit{#1}\xspace} %

\newcommand{\hids}[1]{\textsc{\lowercase{#1}}\xspace} %

\definecolor{forestgreen}{rgb}{0.13, 0.55, 0.13}
\colorlet{jgreen}{white!70!forestgreen}
\colorlet{jred}{white!70!red}
\colorlet{jorange}{orange}

\definecolor{retagbest}{HTML}{1B6AA5}
\definecolor{retagworst}{HTML}{E8110F}
\definecolor{jflagger}{HTML}{F1A340}
\definecolor{jscorer}{HTML}{998EC3}
\definecolor{jpositive}{HTML}{1A85FF}
\definecolor{jnegative}{HTML}{D41159}

\usetikzlibrary{shapes.geometric}

\newcommand{\tikzcircle}[2][red,fill=red]{\tikz[baseline=-0.5ex]\draw[#1,radius=#2] (0,0) circle ;}%
\newcommand{\flaggercircle}{\tikzcircle[fill=jflagger]{4pt}\xspace}
\newcommand{\scorercircle}{\tikzcircle[fill=jscorer]{4pt}\xspace}
\newcommand{\positivecircle}{\tikzcircle[fill=jpositive]{4pt}\xspace}
\newcommand{\negativecircle}{\tikzcircle[fill=jnegative]{4pt}\xspace}
\newcommand{\modelcircle}[1]{\tikzcircle[fill=#1]{4pt}\xspace}

\newcommand{\supportsy}{{\checkmark}}
\newcommand{\supportsn}{{\textperiodcentered}}

\makeatletter
\renewcommand{\appendixsection}[1]{\refstepcounter{section}%
    \setcounter{table}{0}
    \setcounter{figure}{0}
    \setcounter{equation}{0}
    \section*{Appendix \Alph{section}: #1}%
    \def\cref@currentlabel{[appendix][\arabic{section}][]\Alph{section}}%
    \def\@currentlabelname{#1}%
}
\makeatother

\renewcommand{\appendix}{%
    \setcounter{section}{0}
    \renewcommand{\theequation}{\Alph{section}.\arabic{equation}}
    \renewcommand{\thefigure}{\Alph{section}.\arabic{figure}}
    \renewcommand{\thetable}{\Alph{section}.\arabic{table}}
    \renewcommand{\theHequation}{\Alph{section}.\arabic{equation}}
    \renewcommand{\theHfigure}{\Alph{section}.\arabic{figure}}
    \renewcommand{\theHtable}{\Alph{section}.\arabic{table}}
    \renewcommand{\theHsection}{\Alph{section}.\arabic{table}}
    \renewcommand{\theHsection}{\Alph{section}.\arabic{section}}
}

\defcitealias{plankLearningPartofspeechTaggers2014}{Plank et al.~\citeyearpar{plankLearningPartofspeechTaggers2014}}
\defcitealias{hemphillATISSpokenLanguage1990}{Hemphill et al.~\citeyearpar{hemphillATISSpokenLanguage1990}}
\defcitealias{northcuttPervasiveLabelErrors2021}{Northcutt et al.~\citeyearpar{northcuttPervasiveLabelErrors2021}}
\defcitealias{northcuttConfidentLearningEstimating2021}{Northcutt et al.~\citeyearpar{northcuttConfidentLearningEstimating2021}}
\defcitealias{hollensteinInconsistencyDetectionSemantic2016}{Hollenstein et al.~\citeyearpar{hollensteinInconsistencyDetectionSemantic2016}}
\defcitealias{amiriSpottingSpuriousData2018}{Amiri et al.~\citeyearpar{amiriSpottingSpuriousData2018}}
\defcitealias{hendrycksBaselineDetectingMisclassified2017}{Hendrycks et al.~\citeyearpar{hendrycksBaselineDetectingMisclassified2017}}
\defcitealias{dickinsonDetectingErrorsPartofSpeech2003}{Dickinson et al.~\citeyearpar{dickinsonDetectingErrorsPartofSpeech2003}}
\defcitealias{dligachReducingNeedDouble2011}{Dligach et al.~\citeyearpar{dligachReducingNeedDouble2011}}

\DeclareAcronym{nlp}{short = NLP, long = natural language processing}
\DeclareAcronym{aed}{short = AED, long = annotation error detection}
\DeclareAcronym{pos}{short = POS, long = part-of-speech}
\DeclareAcronym{cv}{short = CV, long = cross-validation}

\title{Annotation Error Detection: Analyzing the Past and Present for a More Coherent Future}
\runningtitle{Annotation Error Detection}
\runningauthor{Klie et al.}

\author{Jan-Christoph Klie\thanks{Corresponding author}}
\affil{Ubiquitous Knowledge Processing Lab\\Department of Computer Science\\Technical University of Darmstadt\\\href{ www.ukp.tu-darmstadt.de }{ www.ukp.tu-darmstadt.de }}

\author{Bonnie Webber}
\affil{School of Informatics, University of Edinburgh}

\author{Iryna Gurevych}
\affil{UKP Lab / TU Darmstadt}

\newif\ifarxiv

\arxivtrue

\ifarxiv
\renewcommand{\footmark}{}

\jname{}
\jinfo{}
\else
\pageonefooter{Action editor: Nianwen Xue. Submission received: 7 June 2022; revised version received: 29 August 2022; accepted for publication: 23 September 2022.}
\fi

\begin{document}
    \maketitle

    \begin{abstract}
        Annotated data is an essential ingredient in natural language processing for training and evaluating machine learning models.
It is therefore very desirable for the annotations to be of high quality.
Recent work, however, has shown that several popular datasets contain a surprising amount of annotation errors or inconsistencies.
To alleviate this issue, many methods for annotation error detection have been devised over the years.
While researchers show that their approaches work well on their newly introduced datasets, they rarely compare their methods to previous work or on the same datasets.
This raises strong concerns on methods' general performance and makes it difficult to asses their strengths and weaknesses.
We therefore reimplement 18 methods for detecting potential annotation errors and evaluate them on 9 English datasets for text classification as well as token and span labeling.
In addition, we define a uniform evaluation setup including a new formalization of the annotation error detection task, evaluation protocol and general best practices.
To facilitate future research and reproducibility, we release our datasets and implementations in an easy-to-use and open source software package.\footnote{\url{https://github.com/UKPLab/nessie}}

    \end{abstract}

    \section{Introduction}
\label{sec:introduction}

Annotated corpora are an essential component in many scientific disciplines, including \ac{nlp}~\citep{gururanganDonStopPretraining2020, petersTuneNotTune2019}, linguistics~\citep{haselbachApproximatingTheoreticalLinguistics2012}, language acquisition research~\citep{behrensCorporaLanguageAcquisition2008}, and the digital humanities~\citep{schreibmanCompanionDigitalHumanities2004}.
Corpora are used to train and evaluate machine learning models, to deduce new knowledge, and to suggest appropriate revisions to existing theories.
Especially in machine learning, high quality datasets play a crucial role in advancing the field~ \cite{sunRevisitingUnreasonableEffectiveness2017}.
It is often taken for granted that gold standard corpora do not contain errors --- but alas, this is not always the case.
Datasets are usually annotated by humans who can and do make mistakes~\citep{northcuttPervasiveLabelErrors2021}.
Annotation errors can even be found in corpora used for shared tasks such as  \hids{CoNLL-2003}~\citep{tjongkimsangIntroductionCoNLL2003Shared2003}.
For instance, \jexample{Durban} is annotated there as \jlabel{per} (person) and \jexample{S.AFRICA} as \jlabel{misc} (miscellaneous), but both should be annotated as \jlabel{loc} (location).

Gold standard annotation is also subject to inconsistency, where words or phrases that are intended to refer to the same type of thing (and so should be labelled in the same way) are nevertheless assigned different labels~\citep[see e.g.,][]{hollensteinInconsistencyDetectionSemantic2016}.
For example, in \hids{CoNLL-2003}, when \jexample{Fiorentina} was used to refer to the local football club, it was annotated as \jlabel{org}, but when \jexample{Japan} was used to refer to the Japanese national football team, it was inconsistently annotated as \jlabel{loc}.
One reason for annotation inconsistencies is that tokens can be ambiguous, either because they have multiple senses (e.g., the word \jexample{club} can refer to an organization or to a weapon), or because metonymy allows something to be referred to by one of its parts or attributes (e.g., the Scottish curling team being referred to as \jexample{Scotland}, as in \jexample{Scotland beat Canada in the final    match}).
We further define errors as well as inconsistencies and also discuss ambiguity in detail in \cref{sec:amb_and_disag}.

Such annotation errors or inconsistencies can negatively impact a model's performance or even lead to erroneous conclusions \cite{manningPartofSpeechTagging972011, northcuttPervasiveLabelErrors2021, larsonInconsistenciesCrowdsourcedSlotFilling2020, zhangCrowdsourcingLearningDomain2021}.
A deployed model that learned errors during training can potentially cause harm, especially in critical applications like medical or legal settings.
High-quality labels are needed to evaluate machine learning methods even if they themselves are robust to label noise~\citep[e.g.,][]{songLearningNoisyLabels2020}.
Corpus linguistics relies on correctly annotated data to develop and confirm new theories.
Learner corpora containing errors might be detrimental to the language learning experience and teach wrong lessons.
Hence, it is imperative for datasets to have high-quality labels.

Cleaning the labels by hand, however, is expensive and time consuming.
Therefore, many automatic methods for \acf{aed} have been devised over the years.
These methods enable dataset creators and machine learning practitioners to narrow down the instances which need manual inspection and ---if necessary---- correction.
This reduces the overall work needed to find and fix annotation errors~\citep[see e.g.,][]{reissIdentifyingIncorrectLabels2020}.
As an example, \ac{aed} has been used to discover that widely used benchmark datasets contain errors and inconsistencies~\citep{northcuttPervasiveLabelErrors2021}.
Around 2\% of the samples (sometimes even more than 5\%)  have been found incorrectly annotated in datasets like Penn Treebank ~\citep{dickinsonDetectingErrorsPartofSpeech2003}, sentiment analysis datasets like SST, Amazon Reviews, or IMDb~\citep{barnesSentimentAnalysisNot2019, northcuttPervasiveLabelErrors2021}, CoNLL-2003~\citep{wangCrossWeighTrainingNamed2019, reissIdentifyingIncorrectLabels2020}, or relation extraction in TACRED~\citep{altTACREDRevisitedThorough2020, stoicaReTACREDAddressingShortcomings2021}.
\ac{aed} has likewise been used to find ambigous instances, e.g. for \ac{pos} annotation~\citep{dickinsonDetectingErrorsPartofSpeech2003}.
Additionaly, it has been shown that errors in automatically annotated (silver) corpora can also be found and fixed with the help of \ac{aed}~\citep{rehbeinPOSErrorDetection2014, menardTurningSilverGold2019a}.

While \ac{aed} methods have been applied successfully in the past~\citep[e.g.,][]{reissIdentifyingIncorrectLabels2020}, there are several issues that hinder their wide-spread use.
New approaches for \ac{aed} are  often only evaluated on newly introduced datasets that are proprietary or not otherwise available~\citep[e.g.,][]{dligachReducingNeedDouble2011, amiriSpottingSpuriousData2018, larsonOutlierDetectionImproved2019}.
Also, they rarely compare newly introduced methods to previous works or baselines.
These issues make comparisons of \ac{aed} methods very difficult.
In addition to that, there is neither agreement on how to evaluate \ac{aed} methods nor which metrics to use during their development and application.
As a result, it is often not clear how well \ac{aed} works in practice, especially which \ac{aed} methods should be applied to which kind of data and underlying tasks.
To alleviate these issues, we define a unified evaluation setup for  \ac{aed}, conduct a large scale analysis of 18 \ac{aed} methods and apply them to 9 datasets for text classification, token labeling and span labeling.
This work focuses on errors and inconsistencies related to instance labels.
We leave issues such as boundary errors, sentence splitting or tokenization for future work.
The methods presented in this article are particularly suited to the \ac{nlp} community, but many of them can also be adapted to other tasks (e.g. relation classification) and domains (like computer vision).
The research questions we answer are:

\begin{enumerate}[label={\textbf{RQ\arabic*}},itemsep=0pt]
\label{sec:intro_rq}
    \item Which methods work well across tasks and datasets?
    \item Does model calibration help to improve \ac{aed} performance?
    \item To what extent are model and \ac{aed} performance correlated?
    \item What (performance) impact does using cross-validation have?
\end{enumerate}%
The research reported in this article addresses the aforementioned issues by providing the following contributions:

\begin{description}[itemsep=2pt,wide, labelwidth=!, labelindent=0pt]
	\item[Evaluation methodology] To unify its findings and establish comparability, we first define the task of \acf{aed} and a standardized evaluation setup, including an improvement for evaluating span labeling in this context (\cref{sec:aed_formalization}).
	\item[Easy to use reference implementations] We survey past work from the last 25 years and  implement the 18 most common and generally applicable \ac{aed} methods (\cref{sec:methods}). We publish our implementation in a Python package called \textsc{nessie} which is easy to use, thoroughly tested, and extensible to new methods and tasks. We provide abstractions for models, tasks as well as helpers for cross validation to reduce the boilerplate code needed to a minimum. In addition, we provide extensive documentation and code examples. Our package makes it therefore significantly easier to get started with \ac{aed} for researchers and practitioners alike.
	\item[Benchmarking datasets] We identify, vet, and generate datasets for benchmarking \ac{aed} approaches, which results in 9 datasets for text classification, token labeling, and span labeling (\cref{sec:datasets}).
    We also publish the collected datasets to facilitate easy comparision and reproducibility.
	\item[Evaluation and analysis] Using our implementation, we investigate several fundamental research questions regarding \ac{aed} (\cref{sec:experiments}). We specifically focus on how to achieve the best \ac{aed} performance for each task and dataset, taking model calibration, usage of cross-validation as well as model selection into account. Based on our results, we provide recipes and give recommendations on how to best use \ac{aed} in practice (\cref{sec:takeaways}).
\end{description}%

\section{Related work}
\label{sec:related_work}

This section provides a brief overview of annotation error detection and its related tasks.

\paragraph{Annotation Error Detection}

In most works, \ac{aed} is used as a means to improve the quality of an annotated corpus.
As such, the method used is treated as secondary and possible methods are not compared.
\citet{amiriSpottingSpuriousData2018} and \citet{larsonInconsistenciesCrowdsourcedSlotFilling2020} are the few works that implement different methods and baselines, but only use newly introduced datasets.
In other cases, \ac{aed} is just discussed as a minor contribution and not thoroughly evaluated~\citep[e.g.,][]{swayamdiptaDatasetCartographyMapping2020, rodriguezEvaluationExamplesAre2021}.

Therefore, to the best of our knowledge, no large-scale evaluation of \ac{aed} methods exists.
Closest to the current study is \citet{dickinsonDetectionAnnotationErrors2015}, a survey about the history of annotation error detection.
However, they do not reimplement, compare or evaluate existing methods quantitatively.
Their focus is also limited to part-of-speech and dependency annotations.
Our work fills the aforementioned gaps by reimplementing 18 methods for \ac{aed}, evaluating the methods against 9 datasets, and investigating the setups in which they perform best.

\paragraph{Annotation Error Correction}

After potential errors have been detected, the next step is to have them corrected to obtain gold labels.
This is usually done by human annotators who carefully examine those instances that have been detected.
Some \ac{aed} methods can also both detect and correct labels.
Only a few works have studied correction so far, \citep[for instance][]{kvetonSemiAutomaticDetection2002, loftssonCorrectingPOSTaggedCorpus2009, dickinsonDetectingErrorsAutomatically2006, angleAutomatedErrorCorrection2018, qianAnnotationInconsistencyEntity2021}.
In this study, we focus on detection and leave an in-depth treatment of annotation error correction for future work.

\paragraph{Error Type Classification}

Even if errors are not corrected automatically, it may still be worth identifying the type of each error.
For instance, \citet{larsonInconsistenciesCrowdsourcedSlotFilling2020} investigate the different errors for slot filling, e.g. incorrect span boundaries, incorrect labels or omissions.
\citet{altTACREDRevisitedThorough2020} investigate error types for relation classification.
\citet{yaghoub-zadeh-fardStudyIncorrectParaphrases2019} collect tools and methods to find quality errors in paraphrases used to train conversational agents.
\citet{barnesSentimentAnalysisNot2019} analyze the types of errors found in
annotating for sentiment analysis.
While error type classification has not been explicitly addressed in the current study, such classification requires good \ac{aed}, so the results of the current study can contribute to automate error type classification in the future.

\paragraph{Training with Label Noise}

Related to \ac{aed} is the task of training with noise: given a dataset that potentially contains label errors, train a model so that the performance impact due to noisy labels is as low as possible~\citep{songLearningNoisyLabels2020}.
The goal in this setting is not to clean a dataset (labels are left as is), but to obtain a well-performing machine learning model.
An example application is learning directly from crowdsourced data without adjudicating it~\citep{rodriguesDeepLearningCrowds2018}.
Training with label noise and \ac{aed} have in common that they both enable models to be trained when only noisy data is available.
Evaluating these models still requires clean data which \ac{aed} can help to produce.
    \section{Annotation Error Detection}

In this section we introduce \ac{aed} and formalize the concept, then categorize state-of-the-art approaches according to our formalization.

\subsection{Task definition}
\label{sec:aed_formalization}

Given an adjudicated dataset with one label per annotated instance, the goal of \ac{aed} is to find those instances that are likely labeled incorrectly or inconsistently.
These candidate instances then can be given to human annotators for manual inspection or used in annotation error correction methods.
The definition of \jhi{instance} depends on the task and defines the granularity on which errors or inconsistencies are detected.
In this article, we consider \ac{aed} in text classification (where instances are sentences), in token labeling (instances are tokens, e.g., POS tagging) and in span labeling (instances are spans, e.g., named entity recognition).
\ac{aed} can and has been applied to many domains and tasks, for instance sentiment analysis~\citep{barnesSentimentAnalysisNot2019, northcuttPervasiveLabelErrors2021}, relation extraction~\citep{altTACREDRevisitedThorough2020}, POS tagging~\citep{dickinsonDetectingErrorsPartofSpeech2003, loftssonCorrectingPOSTaggedCorpus2009}, image classification~\citep{northcuttPervasiveLabelErrors2021}, named entity recognition~\citep{wangCrossWeighTrainingNamed2019, reissIdentifyingIncorrectLabels2020}, slot filling~\citep{larsonInconsistenciesCrowdsourcedSlotFilling2020}, or speech classification~\citep{northcuttPervasiveLabelErrors2021}.

We consider a label to be \jhi{incorrect} if there is a unique, true label that should be assigned but it differs from the label that has been assigned.
For example, there is a named entity span Durban in \hids{CoNLL-2003} which
has been labelled \jlabel{per}, whereas in context, it refers to a city in South Africa,
so the label should be \jlabel{loc}.

Instances can also be \jhi{ambiguous}, that is, there are at least two different labels that are valid given the context.
For instance, in the sentence \jexample{They were visiting relatives}, \jexample{visiting} can either be a verb or an adjective.
Ambiguous instances itself are often more difficult for machine learning models to learn from and predict.
Choosing one label over another is neither inherently correct nor incorrect.
But ambiguous instances can be annotated \jhi{inconsistently}.
We consider a label \jhi{inconsistent} if there is more than one potential label for an instance, but the choice of resolution was different for similar instances.
E.g. in the sentence \jexample{Stefan Edberg produced some of his vintage best on Tuesday to extend his grand run at the Grand Slams by toppling Wimbledon champion Richard Krajicek}, the entity \jexample{Wimbledon} was annotated as \jlabel{loc}.
But in the headline of this article, \jexample{Edberg extends Grand Slam run , topples Wimbledon champ}, the entity \jexample{Wimbledon} was annotated as \jlabel{misc}.
We discuss the impact of ambiguity on \ac{aed} further in \cref{sec:amb_and_disag}.
An instance that is neither incorrect nor inconsistent is \jhi{correct}.
If not explicitly stated otherwise, then we refer to both incorrect and inconsistent as incorrect or erroneous.

\ac{aed} is typically used after a new dataset has been annotated and adjudicated.
It is assumed that no already cleaned data and no other data having the same annotation scheme is available.

\paragraph{Flaggers vs. scorers} We divide automatic methods for \ac{aed} into two categories which we dub \jhi{flaggers} and \jhi{scorers}.
Flagging means that methods cast a binary judgement whether the label for an instance is correct or incorrect.
Scoring methods give an estimate on how likely it is that an annotation is incorrect.
These correspond to classification and ranking.

While flaggers are explicit as to whether they consider an annotation to be incorrect, they do not indicate the likelihood of that decision.
On the other hand, while scorers provide a likelihood, they require a threshold value to decide when an annotation is considered an error --- for example, those instances with a score above 80\%. Those would then be given to human evaluation.
Scorers can also be used in settings similar to active learning for error correction~\citep{vlachosActiveAnnotation2006}.

This distinction between flaggers and scorers regarding \ac{aed} has not been made in previous work, as typically approaches of one type or the other have been proposed per paper.
But it is key to understanding why different metrics need to be used when evaluating flaggers compared to scorers, similarly to unranked and ranked evaluation from information retrieval (see \cref{sec:metrics}).

\paragraph{Ambiguity}
\label{sec:amb_and_disag}

In certain \ac{nlp} tasks, there exists more than one valid label per instance~\citep{kehlerCoherenceCoreferenceRevisited2007, plankLinguisticallyDebatableJust2014, aroyoTruthLieCrowd2015, pavlickInherentDisagreementsHuman2019, basileWeNeedConsider2021}.
While this might reduce the usefulness of \ac{aed} at first glance, gold labels are not required by \ac{aed}, as it is about uncovering problems independent of their cause and not assigning a gold label.
Instances detected this way are then marked for further processing.
They can be for instance inspected for whether they are incorrectly or inconsistently annotated.
Ambiguous or difficult instances especially deserve additional scrutiny when creating a corpus; finding them is therefore very useful.
Once found, several alternatives are possible: (1) Ambiguous cases can be corrected~\citep[e.g.,][]{altTACREDRevisitedThorough2020, reissIdentifyingIncorrectLabels2020}; (2) they can be removed~\citep[e.g.,][]{jamisonNoiseAdditionalInformation2015}; (3) their annotation guidelines can be adjusted to reduce disagreement~\citep[e.g.][]{pustejovskyNaturalLanguageAnnotation2013}; (4) the task can eventually be redefined to use soft labels~\citep{fornaciariBlackWhiteLeveraging2021} or used to learn from disagreement~\citep{paunComparingBayesianModels2018}.
Finding such instances is hence very desirable and can be achieved by \ac{aed}.
But similarly to past work on \ac{aed}, we focus on detecting errors and inconsistencies as a first step and leave evaluating ambiguity detection performance for future work.

\subsection{Survey of existing AED methods}
\label{sec:methods}

Over the past three decades, several methods have been developed for \ac{aed}.
Here, we group them by how they detect annotation errors and briefly describe each of them.
In this paper, we focus on \ac{aed} for \acl{nlp}, but (as noted earlier in \cref{sec:introduction}), many of the presented methods can be and have been adjusted to different tasks and modalities.
An overview of the different methods is also given in \cref{tab:methods_overview}.

\begin{table*}[!t]
    \centering
    \caption{Annotation error detection methods evaluated in this work. In most scorer methods, scorer output and erroneous labels are positively correlated. Scorers marked with $^\ast$ show negative correlation.}
    \begin{tabular}{@{}llcccr@{}}
    \toprule
    Abbr &Method Name                & \multicolumn{3}{c}{Tasks}           & Proposed by                                                  \\
    \cmidrule(lr){3-5}
    & & Text      & Token               & Span & \\
    \midrule
    \multicolumn{6}{@{}l}{\textbf{Flagger methods}} \\
    CL                 & Confident Learning         & \supportsy & \supportsy & \supportsy &      \citetalias{northcuttConfidentLearningEstimating2021} \\
    CS                 & Curriculum Spotter         & \supportsy & \supportsn & \supportsn &                 \citetalias{amiriSpottingSpuriousData2018} \\
    DE                 & Diverse Ensemble           & \supportsy & \supportsy & \supportsy &              \citet{loftssonCorrectingPOSTaggedCorpus2009} \\
    IRT                & Item Response Theory       & \supportsy & \supportsy & \supportsy &                 \citet{rodriguezEvaluationExamplesAre2021} \\
    LA                 & Label Aggregation          & \supportsy & \supportsy & \supportsy &                 \citetalias{amiriSpottingSpuriousData2018} \\
    LS                 & Leitner Spotter            & \supportsy & \supportsn & \supportsn &                 \citetalias{amiriSpottingSpuriousData2018} \\
    PE                 & Projection Ensemble        & \supportsy & \supportsy & \supportsy &                \citet{reissIdentifyingIncorrectLabels2020} \\
    RE                 & Retag                      & \supportsy & \supportsy & \supportsy &      \citet{vanhalterenDetectionInconsistencyManually2000} \\
    VN                 & Variation N-Grams          & \supportsn & \supportsy & \supportsy &      \citetalias{dickinsonDetectingErrorsPartofSpeech2003} \\
	\addlinespace
    \multicolumn{6}{@{}l}{\textbf{Scorer methods}} \\
	BC & Borda Count                & \supportsy & \supportsy & \supportsy &   \citet{larsonInconsistenciesCrowdsourcedSlotFilling2020} \\
    CU                 & Classification Uncertainty & \supportsy & \supportsy & \supportsy &   \citetalias{hendrycksBaselineDetectingMisclassified2017} \\
    DM$^\ast$          & Data Map Confidence        & \supportsy & \supportsn & \supportsn &           \citet{swayamdiptaDatasetCartographyMapping2020} \\
    DU                 & Dropout Uncertainty        & \supportsy & \supportsy & \supportsy &                 \citetalias{amiriSpottingSpuriousData2018} \\
    KNN                & k-Nearest Neighbor Entropy & \supportsy & \supportsy & \supportsy &                            \citet{grivasNotCuteStroke2020} \\
    LE                 & Label Entropy              & \supportsn & \supportsy & \supportsy & \citetalias{hollensteinInconsistencyDetectionSemantic2016} \\
    MD                 & Mean Distance              & \supportsy & \supportsy & \supportsy &                 \citet{larsonOutlierDetectionImproved2019} \\
    PM$^\ast$          & Prediction Margin          & \supportsy & \supportsy & \supportsy &                 \citetalias{dligachReducingNeedDouble2011} \\
    WD                 & Weighted Discrepancy       & \supportsn & \supportsy & \supportsy & \citetalias{hollensteinInconsistencyDetectionSemantic2016} \\ \bottomrule
\end{tabular}
    \label{tab:methods_overview}
\end{table*}

\subsubsection{Variation-based}

Methods based on the variation principle leverage the observation that similar surface forms are often annotated with only one or at most a few distinct labels.
If an instance is annotated with a different, rarer label, then it is possibly an annotation error or an inconsistency.
Variation-based methods are relatively easy to implement and can be used in settings in which it is difficult to train a machine learning model, such as low-resource scenarios or tasks that are difficult to train models for, e.g., detecting lexical semantic units~\citep{hollensteinInconsistencyDetectionSemantic2016}.
The main disadvantage of variation-based methods is that they need similar surface forms to perform well, which is not the case in settings like text classification or datasets with diverse instances.

\paragraph{Variation n-grams}

The most frequently used method of this kind is variation n-grams, which has been initially developed for \ac{pos} tagging~\citep{dickinsonDetectingErrorsPartofSpeech2003} and later extended to
discontinuous constituents~\citep{dickinsonDetectingErrorsDiscontinuous2005a}, predicate-argument structures~\citep{dickinsonDetectingErrorsSemantic2008}, dependency parsing~\citep{boydDetectingErrorsDependency2008}, or slot filling~\citep{larsonInconsistenciesCrowdsourcedSlotFilling2020}.
For each instance, n-gram contexts of different sizes are collected and compared to each other.
It is considered incorrect if the label for an instance disagrees with labels from other instances with the same n-gram context.

\paragraph{Label Entropy and Weighted discrepancy}

\citet{hollensteinInconsistencyDetectionSemantic2016} derive metrics from the surface form and label counts which are then used as scorers.
These are the entropy over the label count distribution per surface form or the weighted difference between most and least frequent labels.
They apply their methods to find possible annotation errors in datasets for multi-word expressions and super-sense tagging, which are then reviewed manually for tokens that are actual errors.

\subsubsection{Model-Based}
\label{sec:model_based}

Probabilistic classifiers trained on the to-be-corrected dataset can be used to find annotation errors.
Models in this context are usually trained via \ac{cv} and the respective holdout set is used to detect errors.
After all folds have been used as holdout, the complete dataset is analyzed.
Because some methods described below directly use model probabilities, it is of interest whether these are accurately describing the belief of the model.
This is not always true, as models often are overconfident~\citep{guoCalibrationModernNeural2017}.
Therefore, we will evaluate whether \jhi{calibration}, that is, tuning probabilities so that they are closer to the observed accuracy, can improve performance (see \cref{sec:calibration}).
Several ways have been devised for model-based \ac{aed} which are described below.
Note that most model-based methods are agnostic to the task itself and rely only on model predictions and confidences.
This is why they can easily be used with different tasks and modalities.

\paragraph{Re-tagging}

A simple way to use a trained model for \ac{aed} is to use model predictions directly; when the predicted labels are different than the manually assigned ones, instances are flagged as annotation errors~\citep{vanhalterenDetectionInconsistencyManually2000}.
\citet{larsonInconsistenciesCrowdsourcedSlotFilling2020} apply this using a conditional random field (CRF) tagger to find errors in crowdsourced slot-filling annotations.
Similarly, \citet{amiriSpottingSpuriousData2018} use \himl{Retag} for text classification.
\citet{yaghoub-zadeh-fardStudyIncorrectParaphrases2019} train machine learning models to classify whether paraphrases contain errors and if they do, what kind of error it is.
To reduce the need of annotating instances twice for higher quality, \citet{dligachReducingNeedDouble2011} train a model on the labels given by an initial annotator.
If the model disagrees with the instance's labeling, then it is flagged for re-annotation.
For cleaning dependency annotations in a Hindi treebank, \citet{ambatiErrorDetectionTreebank2011} train a logistic regression classifier.
If the model's label does not agree with the original annotation and the model confidence is above a predefined threshold, then the annotation is considered to be incorrect.
\himl{CrossWeigh}~\citep{wangCrossWeighTrainingNamed2019} is similar to \himl{Retag} with repeated \ac{cv}.
During \ac{cv}, \emph{entity disjoint filtering} is used to force more model errors: instances are flagged as erroneous if the probability of their having the correct label falls below the respective threshold.
As it is computationally much more expensive than \himl{Retag} while being very similar, we did not include it in our comparison.

\paragraph{Classification Uncertainty}

Probabilistic classification models assign probabilities which are typically higher for instances that are correctly labeled compared to erroneous ones~\citep{hendrycksBaselineDetectingMisclassified2017}.
Therefore, the class probabilities of the noisy labels can be used to score these for being an annotation error.
Using model uncertainty is basically identical to using the network loss --- as e.g. used by \citet{amiriSpottingSpuriousData2018} --- because the cross-entropy function used to compute the loss is monotonic.
The probability formulation however allows us to use calibration more easily later (see \cref{sec:calibration}), which is why we adapt the former instead of using the loss.

\paragraph{Prediction Margin}

Inspired by active learning,  \himl{Predictive Margin} uses the probabilities of the two highest scoring labels for an instance.
The resulting score is simply their difference \cite{dligachReducingNeedDouble2011}.
The intuition behind this is that samples with smaller margin are more likely to be an annotation error, since the smaller the decision margin is the more unsure the model was.

\paragraph{Confident Learning} This method estimates the joint distribution of noisy and true labels~\citep{northcuttConfidentLearningEstimating2021}. A threshold is then learnt (the average self-confidence) and instances whose computed probability of having the correct label is below the respective threshold are flagged as erroneous.

\paragraph{Dropout Uncertainty}

\citet{amiriSpottingSpuriousData2018} use Monte Carlo dropout~\citep{galDropoutBayesianApproximation2016} to estimate the uncertainty of an underlying model.
There are different acquisition methods to compute uncertainty from the stochastic passes.
A summary can be found in  \citet{shelmanovHowCertainYour2021}.
The work of \citet{amiriSpottingSpuriousData2018} uses the probability variance averaged over classes.

\paragraph{Label Aggregation}

Given $T$ predictions obtained via Monte Carlo dropout, \citet{amiriSpottingSpuriousData2018} use MACE~\citep{hovyLearningWhomTrust2013}, an aggregation technique from crowdsourcing to adjudicate the resulting repeated predictions.

\subsubsection{Training Dynamics}

Methods based on training dynamics use information derived from how a model behaves during training and how predictions change over the course of its training.

\paragraph{Curriculum and Leitner Spotter}

\citet{amiriSpottingSpuriousData2018} train a model via curriculum learning, where the network trains on easier instances during earlier epochs and is then gradually introduced to harder instances.
Instances then are ranked by how hard they were perceived during training.
They also adapt the ideas of the Zettelkasten~\citep{ahrensHowTakeSmart2017} and Leitner queue
networks~\citep{leitnerLerntManLeben1974} to model training.
There, difficult instances are presented more often during training than easier ones.
The assumption behind both of these methods is that instances that are perceived harder or misclassified more frequently are more often annotation errors than easier ones.
These two methods require that the instances can be scheduled independently.
This is  for instance not the case for sequence labeling, as the model trains on complete sentences and not individual tokens or spans.
Even if they have different difficulties, they would end up in the same batch nonetheless.

\paragraph{Data Map Confidence}

\citet{swayamdiptaDatasetCartographyMapping2020} use the class probability for each instance's gold label across epochs as a measure of confidence.
In their experiments, low confidence correlates well with an item having a incorrect label.

\subsubsection{Vector Space Proximity}

Approaches of this kind leverage dense embeddings of tokens, spans, and texts into a vector space and use their distribution therein.
The distance of an instance to semantically similar instances is expected to be smaller than the distance to semantically different ones.
Embeddings are typically obtained by using BERT-type models  for tokens and spans~\citep{devlinBERTPretrainingDeep2019a} or S-BERT for sentences~\citep{reimersSentenceBERTSentenceEmbeddings2019a}.

\paragraph{Mean distance}

\citet{larsonOutlierDetectionImproved2019} compute the centroid of each class by averaging vector embeddings of the respective instances.
Items are then scored by the distance between their embedding vector to their centroid.
The underlying assumption is that semantically similar items should have the same label and be close together (and thereby close to the centroid) in the vector space.
In the original publication, this method was only evaluated on detecting errors in sentence classification datasets, but we extend it to also token and span classification.

\paragraph{k-Nearest-Neighbor Entropy}

In the context of named entity recognition in clinical reports, \citet{grivasNotCuteStroke2020} leverage the work of \citet{khandelwalGeneralizationMemorizationNearest2020} regarding nearest-neighbor language models to find mislabeled named entities.
First, all instances are embedded into a vector space.
Then, the $k$ nearest neighbors of each instance according to their Euclidean distance are retrieved.
Their distances to the instance embedding vector are then used to compute a distribution over labels by applying softmax.
An instance's score is then the entropy of its distance distribution; if it is large, it indicates uncertainty, hinting at being mislabeled.
\citet{grivasNotCuteStroke2020} only used this method qualitatively; we have turned their qualitative approach into a method that can be used to score instances automatically and evaluated it on detecting errors in both named entity recognition and sentence classification -- the latter using S-BERT embeddings.
This method was only evaluated on detecting errors in named entity recognition datasets, but we apply it to sentence classification as well by using S-BERT embeddings.

\subsubsection{Ensembling}

Ensemble methods combine the scores or predictions of several individual flaggers or scorers to obtain better performance than the sum of their parts.

\paragraph{Diverse Ensemble}

Instead of using a single prediction like \himl{Retag} does, the predictions of several, architecturally different models are aggregated.
If most of them disagree on the label for an instance, then it is likely to be an annotation error.
\citet{altTACREDRevisitedThorough2020} use an ensemble of 49 different models to find annotation errors in the TACRED relation extraction corpus.
In their setup, instances are ranked by how often a model suggests a label different than the original one.
\citet{barnesSentimentAnalysisNot2019} use three models to analyze error types on several sentiment analysis datasets; they flag instances for which all models disagree with the gold label.
\citet{loftssonCorrectingPOSTaggedCorpus2009, angleAutomatedErrorCorrection2018} use an ensemble of different taggers to correct \ac{pos} tags.

\paragraph{Projection Ensemble}

In order to correct the \hids{CoNLL-2003} named entity corpus, \citet{reissIdentifyingIncorrectLabels2020} train 17 logistic regression models on different Gaussian projections of BERT embeddings.
The aggregated predictions that disagree with the dataset were then corrected by hand.

\paragraph{Item Response Theory}

\citet{lordStatisticalTheoriesMental1968} developed \himl{Item Response Theory} as a mathematical framework to model relationships between measured responses of test  subjects (e.g. answers to questions in an exam) for an underlying, latent trait  (e.g. the overall grasp on the subject that is tested).
It can also be used to estimate the discriminative power of an item, i.e. how well the response to a question can be used to distinguish between subjects of different ability.
In the context of \ac{aed}, test subjects are trained models, the observations are the predictions on the dataset and the latent trait is task performance.
\citet{rodriguezEvaluationExamplesAre2021} have shown that items which negatively discriminate --- i.e., where a better response indicates being less skilled --- correlate with annotation errors.

\paragraph{Borda Count}

Similarly to combining several flaggers into an ensemble, rankings obtained from different scorers can be combined as well.
For that, \citet{dworkRankAggregationMethods2001} propose to leverage Borda counts, a voting scheme that assigns points based on their ranking.
For each scorer, given scores for $N$ instances, the instance that is ranked the highest is given $N$ points, the second-highest $N-1$ and so on~\citep{szpiroNumbersRuleVexing2010}.
The points assigned by different scorers are then summed up for each instance and form the aggregated ranking.
\citet{larsonOutlierDetectionImproved2019} use this to combine scores for runs of \himl{Mean Distance} with different embeddings and show that this improves overall performance compared to only using individual scores.

\subsubsection{Rule-based}

Several works leverage rules that describe which annotations are valid and which are not.
For example, to find errors in \ac{pos} annotated corpora, \citet{kvetonSemiAutomaticDetection2002} developed a set of conditions that tags have to fulfill in order to be valid, especially n-grams that are impossible based on the underlying  lexical or morphological information of their respective surface forms.
Rule-based approaches for \ac{aed} can be very effective but are hand-tailored to the respective dataset, its domain, language, and task.
Our focus in this article is to evaluate generally applicable methods that can be used for many different tasks and settings.
Therefore, we do not discuss rule-based methods further in the current work.

    \section{Datasets and tasks}
\label{sec:datasets}

In order to compare the performance of \ac{aed} methods on a large-scale, we need datasets with parallel gold and noisy labels.
But even with previous work on correcting noisy corpora, such datasets are hard to find.

We consider three kinds of approaches to obtain datasets that can be used for evaluating \ac{aed}. First, existing datasets can be used whose labels are then randomly perturbed.
Second, there exist adjudicated gold corpora for which the annotations of single annotators exist.
Noisy labels are then the unadjucated annotations.
These kinds of corpora are mainly obtained from crowdsourcing experiments.
Third, there are manually corrected corpora whose both clean and noisy parts have been made public.
Because only a few such datasets are available for \ac{aed}, we have derived several datasets of the first two types from existing corpora.

When injecting random noise we use flipped label noise~\citep{zhengMetaLabelCorrection2021} with a noise level of 5\% which is in a similar range to error rates in previously examined datasets like \hids{Penn Treebank}~\citep{dickinsonDetectingInconsistenciesTreebanks2003} or \hids{CoNLL-2003}~\citep{reissIdentifyingIncorrectLabels2020}.
In our settings, for a random subset of 5\% instances, this kind of noise assigns uniformly a different label from the tagset without taking the original label into account.
While randomly injecting noise is simple and can be applied to any existing gold corpus, errors in these datasets are often easy to spot~\citep{larsonOutlierDetectionImproved2019}.
This is because errors typically made by human annotators vary with the actual label, which is not true for random noise~\citep{hedderichAnalysingNoiseModel2021}.
Note that evaluating \ac{aed} methods does not require knowing true labels: all that is required are potentially noisy labels and whether or not they are erroneous.
It is only correction that needs true labels as well as noisy ones.

As noted earlier, we will address \ac{aed} in three broad \ac{nlp} tasks: text classification, token and span labeling.
These have been the tasks most frequently evaluated in \ac{aed} and on which the majority of methods can be applied.
Also, these tasks have many different machine learning models available to solve them.
This is crucial for evaluating calibration (\cref{sec:calibration}) and assessing whether well-performing models lead to better task performance for model-based methods (\cref{sec:r2_relationship}).
To foster reproducibility and to obtain representative results, we then choose datasets that fulfill the following requirements: 1) they are available openly and free of charge, 2) they are for common and different \ac{nlp} tasks 3) they come from different domains, 4) they have high inter-annotator agreement and very few annotation errors.
Based on these criteria, we select 9 datasets.
They are listed in \cref{tab:datasets_overview} and are described in the following section.
We manually inspected and carefully analyzed the corpora to verify that the given gold labels are of very high quality.

\begin{table}[t]
    \centering
    \caption{Dataset statistics.
    We report the number of instances $|\mathcal{I}|$ and annotations $|\mathcal{A}|$ as well as the number of mislabeled ones ($|\mathcal{I}_\epsilon|$ and $|\mathcal{A}_\epsilon|$), their percentage as well as the number of classes $|\mathcal{C}|$.
    For token and span labeling datasets, $|\mathcal{A}|$ counts the number of annotated tokens and spans, respectively.
    \jhi{Kind} indicates whether the noisy part was created by randomly corrupting labels (R), or by aggregation (A) from individual annotations like crowdsourcing, or whether the gold labels stem from manual correction (M).
    Errors for span labeling are calculated via exact span match. \emph{Source} points to the work that introduced the dataset for use in \ac{aed} if it was created via manual correction and to the work proposing the initial dataset for aggregation or randomly perturbed ones.}
    \resizebox{\textwidth}{!}{
\begin{tabular}{@{}lrrrrrrrcr@{}}
    \toprule
    Name            &   $|\mathcal{I}|$ &   $|\mathcal{I}_\epsilon|$ &   $\frac{|\mathcal{I}_\epsilon|}{|\mathcal{I}|} \%$ &   $|\mathcal{A}|$ & $|\mathcal{A}_\epsilon|$ &   $\frac{|\mathcal{A}_\epsilon|}{|\mathcal{A}|} \%$ &   $|\mathcal{C}|$ & Kind &  Source \\
    \cmidrule(r){1-1}\cmidrule(lr){2-4}\cmidrule(lr){5-7}\cmidrule(lr){8-8}\cmidrule(lr){9-9}\cmidrule(l){10-10}
    \multicolumn{10}{@{}l}{\textbf{Text classification}} \\
    ATIS                                      &  4978 &  238 &  4.78 &   4978 &  238 &  4.78 & 22 &  R &              \citetalias{hemphillATISSpokenLanguage1990} \\
    IMDb                                      & 24799 &  499 &  2.01 &  24799 &  499 &  2.01 &  2 &  M &           \citetalias{northcuttPervasiveLabelErrors2021} \\
    SST                                       &  8544 &  420 &  4.92 &   8544 &  420 &  4.92 &  2 &  R &                    \citet{socherRecursiveDeepModels2013} \\
    \addlinespace
    \multicolumn{10}{@{}l}{\textbf{Token labeling}} \\
    GUM                   &  7397 & 3920 & 52.99 & 137605 & 6835 &  4.97 & 18 & R &                      \citet{zeldesGUMCorpusCreating2017} \\
    Plank                                     &   500 &  373 & 74.60 &   7876 &  931 & 11.82 & 13 & A &        \citetalias{plankLearningPartofspeechTaggers2014} \\
    \addlinespace
    \multicolumn{10}{@{}l}{\textbf{Span labeling}} \\
     CoNLL-2003              &  3380 &         217 &        6.42 &   5505 &         262 &        4.76 &     5 &    M    & \citet{reissIdentifyingIncorrectLabels2020}              \\
    SI Companies                                  &   500 &  224 & 44.80 &   1365 &  325 & 23.81 & 11 & M & \citet{larsonInconsistenciesCrowdsourcedSlotFilling2020} \\
    SI Flights                                &   500 &   43 &  8.60 &   1196 &   49 &  4.10 &  7 & M & \citet{larsonInconsistenciesCrowdsourcedSlotFilling2020} \\
    SI Forex                                  &   520 &   63 & 12.12 &   1263 &   98 &  7.76 &  4 & M & \citet{larsonInconsistenciesCrowdsourcedSlotFilling2020} \\ \bottomrule
\end{tabular}
}
    \label{tab:datasets_overview}
\end{table}

\subsection{Text Classification}

The goal of text classification is to assign a predefined category to a given text sequence (here, a sentence, paragraph, or a document).
Example applications are news categorization, sentiment analysis or intent detection.
For text classification, each individual sentence or document is considered its own instance.

\begin{description}
\item[ATIS] contains transcripts of user interactions with travel inquiry  systems, annotated with intents and slots.
For \ac{aed} on intent classification, we have randomly perturbed the labels.
\item[IMDb] contains movie reviews labeled with sentiment.
\citet{northcuttPervasiveLabelErrors2021} discovered that it contains a non-negligible amount of annotation errors. They applied \himl{Confident Learning} to the test set and let crowdworkers check whether the flags were genuine.
\item[SST] The \hids{Stanford Sentiment Treebank} is a  dataset for sentiment analysis of movie reviews from Rotten Tomatoes.
We use it for binary sentiment classification and randomly perturb the labels.
\end{description}

\subsection{Token Labeling}

The task of token labeling is to assign a label to each token.
The most common task in this category is \ac{pos} tagging.
As there are not many other tasks with easily obtainable datasets, we only use two different \ac{pos} tagging datasets.
For token labeling, each individual token is considered an instance.

\begin{description}
\item[GUM] The \hids{Georgetown University Multilayer Corpus} is an open source corpus annotated with several layers from the Universal Dependencies project~\citep{nivreUniversalDependenciesV22020}.
It has been collected by linguistics students at Georgetown University as part of their course work.
Here, the original labels have been perturbed with random noise.
\item[Plank POS] contains Twitter posts that were annotated by \citet{gimpelPartofspeechTaggingTwitter2011}.
\citet{plankLearningPartofspeechTaggers2014} mapped their labels to Universal POS tags and had 500 tweets reannotated by two new annotators.
We flag an instance as erroneous if its two annotations disagree.
\end{description}

\subsection{Span Labeling}
\label{sec:span_alignment}

Span labeling assigns labels not to single tokens, but to spans of text.
Common tasks that can be modeled that way are named-entity recognition (NER), slot filling or chunking.
In this work, we assume that spans have already been identified, focussing only on finding label errors and leaving detecting boundary errors and related issues for future work.
We use the following datasets:

\begin{description}
\item[CoNLL-2003] is a widely used dataset for named-entity recognition~\citep{tjongkimsangIntroductionCoNLL2003Shared2003}.
It consists of news wire articles from the Reuters Corpus annotated by experts.
\citet{reissIdentifyingIncorrectLabels2020} discovered several annotation errors in the English portion of the dataset.
They developed \himl{Projection Ensembles} and then manually corrected the instances flagged by it.
While errors concerning tokenization and sentence splitting were also corrected, we ignore them here as being out of scope of the current study.
Therefore, we report slightly fewer instances and errors overall in \cref{tab:datasets_overview}.
\citet{wangCrossWeighTrainingNamed2019} also corrected errors in \hids{CoNLL-2003} and named the resulting corpus \hids{CoNLL++}.
As they only re-annotated the test set and found fewer errors, we use the corrected version of \citet{reissIdentifyingIncorrectLabels2020}.
\item[Slot Inconsistencies] is a dataset that was created by \citet{larsonInconsistenciesCrowdsourcedSlotFilling2020} to investigate and classify errors in slot filling annotations.
It contains documents of three domains (\hids{companies}, \hids{forex}, \hids{flights}) which were annotated via crowdsourcing.
Errors were then manually corrected by experts.
\end{description}
Span labeling is typically indicated using Begin-Inside-Out (BIO) tags.\footnote{For simplicity, we describe the BIO tagging format. There are more advanced schemas like BIOES, but our resulting task-specific evaluation is independent of the actual schema used.}
When labeling a span as \texttt{X}, tokens outside the span are labelled \texttt{O},  while the token at the beginning of the span is labeled \texttt{B-X} and tokens within the span are labelled \texttt{I-X}.
Datasets for span labeling are also usually represented in this format.

This raises the issues of (1) boundary differences and (2) split entities.
First, for model-based methods, models might predict different spans and span boundaries from the original annotations.
In many evaluation datasets, boundary issues were also corrected and therefore boundaries for the same span in the clean and noisy data can be different, which makes evaluation difficult.
Second, for scorers it does not make much sense to order BIO tagged tokens independently of their neighbors or to alter only parts of a sequence to a different label.
This can lead to corrections that split entities which is often undesirable.
Therefore, directly using BIO tags as the granularity of detection and correction for span labeling is problematic.

\begin{figure}[hbt]
    \includegraphics[width=\textwidth]{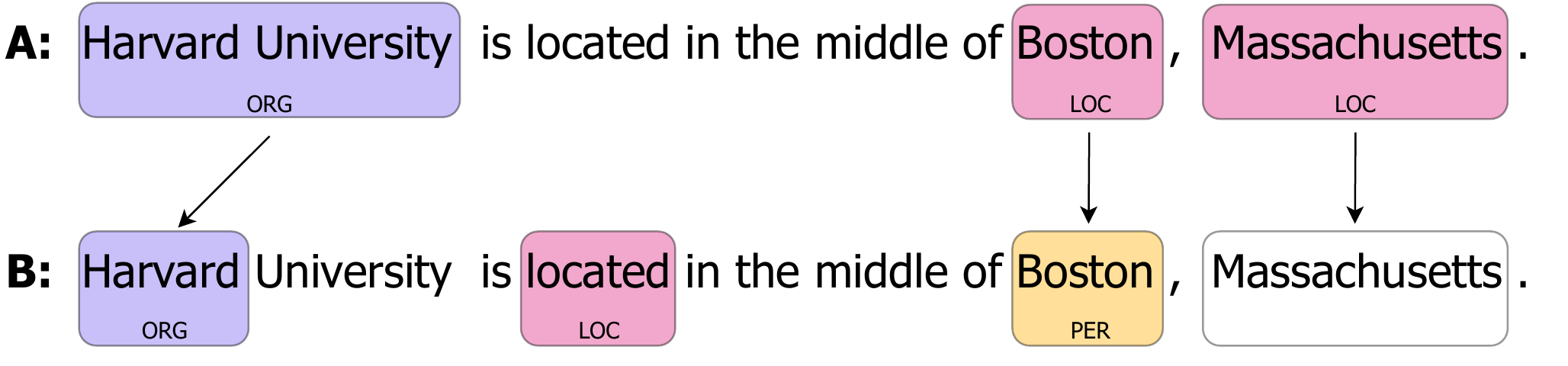}
    \caption{Alignment between original or corrected spans $A$ and noisy or predicted spans $B$. The goal is to find an alignment which maximizes overlap. Spans that are in $A$ but find no match in $B$ are given a match with the same offsets but a special, unique label that is different from all other labels (e.g., \jexample{Massachusetts}). Spans that are in $B$ but find no match in $A$ are dropped (e.g., \jexample{located}).  Spans from $A$ that have no overlapping span in $B$ are considered different and cannot be aligned (e.g., \jexample{Boston} in $A$ and \jexample{Massachusetts} in $B$). Span colors here indicate their labels.}
    \label{fig:span_alignment}
\end{figure}

Hence, we suggest converting the BIO tagged sequences back to a span representation consisting of begin, end and label.
This first step solves the issue of entities potentially being torn apart by detection and correction.
Spans from the original data and from the model predictions then need to be aligned for evaluation in order to reduce boundary issues.
This is depicted in \cref{fig:span_alignment}.

We require a good alignment to (1) maximize overlap between aligned spans so that the most likely spans are aligned, (2) be deterministic, (3) not use additional information like probabilities, and (4) not align spans that have no overlap to avoid aligning things that should not be aligned.
If these properties are not given, then the alignment and resulting confidences or representations that are computed based on this can be subpar.
This kind of alignment is related to evaluation for e.g. named entity recognition in style of MUC-5~\citep{chinchorMUC5EvaluationMetrics1993}, especially for partial matching.
Their alignment does not however satisfy (1) and (3) in the case of multiple predictions overlapping with a single gold entity.
For instance, if the gold entity is \jexample{New York City} and the system predicted \jexample{York} and \jexample{New York}, then in most implementations, the first prediction is chosen and other predictions that also could match are discarded.
What prediction is being first depends on the order of predictions which is non-deterministic.
This also does not choose the optimal alignment with maximum span overlap, which requires a more involved approach.

We thus adopt the following alignment procedure: Given a sequence of tokens, a set of original spans $A$ and predicted/noisy spans $B$, align both sets of spans and thereby allow certain leeway of boundaries.
The goal is to find an assignment that maximizes overlap of spans in $A$ and $B$; only spans of $A$ that overlap in at least one token with spans in $B$ are considered.
This can be formulated as a linear sum assignment problem: given two sets $A, B$ of equal size and a function that assigns a cost to connect an element of $A$ with an element of $B$, find the assignment that minimizes the overall cost~\citep{burkardAssignmentProblemsRevised2012}.
It can happen that not all elements of $A$ are assigned a match in $B$ and vice versa, we assign a special label that indicates missing alignment in the first case and drop spans of $B$ that have no overlap in $A$.
For the latter, it is possible to also assign a special label to indicate that a gold entity is missing; in this work, we focus on correcting labels only and hence leave using this information to detect missing spans for future work.

We are not aware of previous works that propose a certain methodology for this.
While \citet{larsonInconsistenciesCrowdsourcedSlotFilling2020} evaluate \ac{aed} on slot filling, it is not clear on which granularity they measure detection performance or whether and how they align.
To the best of our knowledge, we are the first to propose this span alignment approach for span-level \ac{aed}.
Span alignment requires aggregating token probabilities into span probabilities, which is described in \cref{sec:app_aggregation}.
This alignment approach can also be extended to other tasks like object classification or matching boxes for optical character recognition.
In that case, the metric to optimize is the Jaccard index.

    \section{Experiments}
\label{sec:experiments}

In this section we first define the general evaluation setup, metrics to be used, and the models that are leveraged for model-based \ac{aed}.
Details on how each method was implemented for this work can be found in the appendix.
In \cref{sec:best} through \cref{sec:r3_cross_validation}, we then describe our results for the experiments we perform to answer the research questions raised in \cref{sec:intro_rq}.

\paragraph{Metrics}
\label{sec:metrics}

As described in \cref{sec:aed_formalization}, we differentiate between two kinds of annotation error detectors, \jhi{flaggers} and \jhi{scorers}.
These need different metrics during evaluation, similar to unranked and ranked evaluation from information retrieval~\citep{manningIntroductionInformationRetrieval2008}.
Flagging is a binary classification task.
Therefore, we use the standard metrics for this task which are precision, recall, and  F1.
We also record the percentage of instances flagged \citep{larsonInconsistenciesCrowdsourcedSlotFilling2020}.
Scoring produces a ranking, as in information retrieval.
We use average precision\footnote{Also known as Area Under the Precision-Recall Curve (AUPR/AUPRC). In \ac{aed}, AP is also identical to mean average precision (mAP) used in other works.} (AP), Precision@10\%, and Recall@10\%, similarly to \citet{amiriSpottingSpuriousData2018, larsonOutlierDetectionImproved2019}.
There are reasons why both precision and recall can be considered the more important metric of the two.
A low precision leads to increased cost because many more instances than necessary need to be inspected manually after detection.
Similarly, a low recall leads to problems because there still can be errors left after the application of \ac{aed}.
As both arguments have merit, we will mainly use the aggregated metrics F1 and AP.
Precision and recall at 10\% evaluate a scenario in which a scorer was applied and the first 10\%  with the highest score --- most likely to be incorrectly annotated --- are manually corrected.
We use the \textsc{pytrec-eval} toolkit to compute these ranking  metrics~\footnote{\url{https://github.com/cvangysel/pytrec_eval}}.
Recall relies on knowing the exact number of correctly and incorrectly annotated instances.
While this information may be available when developing and evaluating \ac{aed} methods, it is generally not available when actually applying \ac{aed} to clean real data.
One solution to computing recall then is to have experts carefully annotate a subset of the data and then use it to estimate recall overall.

In contrast to previous work, we explicitly do not use ROC AUC and discourage its use for \ac{aed}, as it heavily overestimates performance when applied to imbalanced datasets~\citep{davisRelationshipPrecisionRecallROC2006, saitoPrecisionRecallPlotMore2015}.
Datasets needing \ac{aed} are typically very imbalanced because there are far more correct labels than incorrect ones.

\paragraph{Models}
\label{sec:models}

We use multiple different neural and non-neural model types  per task for model-based \ac{aed}.
These are used to investigate the relationship between model and method performances, whether model calibration can improve method performances and for creating diverse ensembles.

For text classification we use seven different models: gradient boosting machines~\citep{keLightGBMHighlyEfficient2017a} with either bag-of-word or S-BERT features~\citep{reimersSentenceBERTSentenceEmbeddings2019a}, transformer based on DistilRoBERTa~\citep{sanhDistilBERTDistilledVersion2019}, BiLSTM based on Flair~\citep{akbikFLAIREasytoUseFramework2019}, and FastText~\citep{joulinBagTricksEfficient2017}.
For S-BERT, we use \texttt{all-mpnet-base-v2} as the underlying model, as it has been shown  by their creators to produce sentence embeddings of the highest quality overall.

For token and span labeling, we use four different models: conditional random fields (CRF) with the hand-crafted features as proposed by \citet{gimpelPartofspeechTaggingTwitter2011}, BiLSTM + CRF based on Flair~\citep{akbikFLAIREasytoUseFramework2019}, transformers with CRF based on DistilRoBERTa~\citep{sanhDistilBERTDistilledVersion2019}, and logistic regression (also called maximum entropy model).
For the initialization of Flair-based models we use a combination of Glove~\citep{penningtonGloVeGlobalVectors2014} as well as  Byte-Pair Encoding embeddings~\citep{heinzerlingBPEmbTokenizationfreePretrained2018} and a hidden layer size of 256 for both text classification and sequence labeling.
Note that we do not perform extensive hyper-parameter tuning for model selection because when using \ac{aed} in practice, no annotated in-domain data can be held-out for tuning since all data must be checked for errors.
Also, when comparing models as we do here, it would be prohibitively expensive to carry out hyper-parameter tuning across all datasets and model combinations. Instead, we use default configurations that have been shown to work well on a wide range of tasks and datasets.

When using transformers for sequence labeling we use the probabilities of the first subword token.
We use 10-fold cross-validation to train each model and use the same model weights for all methods evaluated on the same fold.
Thereby, all methods applied to the same fold use the predictions of the same model.

\subsection{RQ1 -- Which methods work well across tasks and datasets?}
\label{sec:best}
We first report the scores resulting from the best setup as a reference to the upcoming experiments.
Then we describe the experiments and results which lead to this setup.
We do not apply calibration to any of the methods for the reported scores because it only marginally improved performance (see \cref{sec:calibration}).
For model-based methods, the best performance for text classification and span labeling was achieved using transformers; for token labeling, best performance was achieved using Flair (see \cref{sec:r2_relationship}).
Not using cross-validation for model-based \ac{aed} was found to substantially reduce recall for model-based AED (see \cref{sec:r3_cross_validation}), so we have used 10-fold cross-validation in comparing model-based methods.

In \cref{tab:best_wide}, we present the overall performance in F1 respectively AP across all datasets and tasks.
Detailed results including scores for all metrics can be found in \cref{appendix:app:best}.

\begin{table*}[b]
    \centering
    \caption{\textbf{F1} and \textbf{AP} for all implemented flaggers \flaggercircle as well as scorers \scorercircle evaluated with the best overall setups.
    We also report the harmonic mean \textbf{H}  \positivecircle computed across all datasets. \textbf{L}abel \textbf{A}ggregation, \textbf{Re}tag, \textbf{D}iverse \textbf{E}nsemble and \textbf{B}orda \textbf{C}ount perform especially well across tasks and datasets. Datasets created via injecting random noise (\hids{ATIS}, \hids{SST} and \hids{GUM}) are comparatively easier to detect errors in.}

    \begingroup
    \renewcommand*{\arraystretch}{1.2}
    \renewcommand*{\tabcolsep}{4pt}
    \begin{small}
    
\begin{tabular}{ccccccccccc}
\toprule
\multicolumn{1}{c}{ } & \multicolumn{3}{c}{Text} & \multicolumn{2}{c}{Token} & \multicolumn{4}{c}{Span} \\
\cmidrule(l{3pt}r{3pt}){2-4} \cmidrule(l{3pt}r{3pt}){5-6} \cmidrule(l{3pt}r{3pt}){7-10}
Method & ATIS & IMDb & SST & GUM & Plank & Comp. & CoNLL & Flights & Forex & H\\
\midrule
\addlinespace[0.3em]
\multicolumn{11}{l}{\textbf{Flagger}}\\
\hspace{1em}CL & \cellcolor[HTML]{FFDEBB}{\textcolor{black}{0.35}} & \cellcolor[HTML]{FFDFBF}{\textcolor{black}{0.33}} & \cellcolor[HTML]{FFDFBE}{\textcolor{black}{0.34}} & \cellcolor[HTML]{F8B567}{\textcolor{black}{0.80}} & \cellcolor[HTML]{FFDCB8}{\textcolor{black}{0.37}} & \cellcolor[HTML]{FFD0A0}{\textcolor{black}{0.50}} & \cellcolor[HTML]{FFE8D0}{\textcolor{black}{0.24}} & \cellcolor[HTML]{FFD8B0}{\textcolor{black}{0.42}} & \cellcolor[HTML]{FECA93}{\textcolor{black}{0.57}} & \cellcolor[HTML]{BFCDFF}{\textcolor{black}{0.39}}\\
\hspace{1em}DE & \cellcolor[HTML]{FBBC77}{\textcolor{black}{0.72}} & \cellcolor[HTML]{FFE3C6}{\textcolor{black}{0.30}} & \cellcolor[HTML]{FFE0BF}{\textcolor{black}{0.33}} & \cellcolor[HTML]{FABA73}{\textcolor{black}{0.74}} & \cellcolor[HTML]{FFD2A4}{\textcolor{black}{0.48}} & \cellcolor[HTML]{FEC992}{\textcolor{black}{0.57}} & \cellcolor[HTML]{FFE4C9}{\textcolor{black}{0.28}} & \cellcolor[HTML]{FFCC97}{\textcolor{black}{0.55}} & \cellcolor[HTML]{FDC486}{\textcolor{black}{0.64}} & \cellcolor[HTML]{B4C5FF}{\textcolor{black}{0.45}}\\
\hspace{1em}IRT & \cellcolor[HTML]{FFFFFF}{\textcolor{black}{0.00}} & \cellcolor[HTML]{FFFEFC}{\textcolor{black}{0.01}} & \cellcolor[HTML]{FFFDFB}{\textcolor{black}{0.02}} & \cellcolor[HTML]{FFFFFF}{\textcolor{black}{0.00}} & \cellcolor[HTML]{FFF4E8}{\textcolor{black}{0.12}} & \cellcolor[HTML]{FFD8B1}{\textcolor{black}{0.41}} & \cellcolor[HTML]{FFE4C7}{\textcolor{black}{0.29}} & \cellcolor[HTML]{FFFEFC}{\textcolor{black}{0.02}} & \cellcolor[HTML]{FDC589}{\textcolor{black}{0.62}} & \cellcolor[HTML]{FFFFFF}{\textcolor{black}{0.00}}\\
\hspace{1em}LA & \cellcolor[HTML]{F7B262}{\textcolor{black}{0.83}} & \cellcolor[HTML]{FFE0BF}{\textcolor{black}{0.33}} & \cellcolor[HTML]{FFDEBC}{\textcolor{black}{0.35}} & \cellcolor[HTML]{FCC07E}{\textcolor{black}{0.68}} & \cellcolor[HTML]{FFD1A2}{\textcolor{black}{0.49}} & \cellcolor[HTML]{FEC88E}{\textcolor{black}{0.59}} & \cellcolor[HTML]{FFE3C5}{\textcolor{black}{0.30}} & \cellcolor[HTML]{FCC282}{\textcolor{black}{0.66}} & \cellcolor[HTML]{FBBE7A}{\textcolor{black}{0.70}} & \cellcolor[HTML]{AFC1FF}{\textcolor{black}{0.48}}\\
\hspace{1em}PE & \cellcolor[HTML]{FFCD98}{\textcolor{black}{0.54}} & \cellcolor[HTML]{FFEEDD}{\textcolor{black}{0.18}} & \cellcolor[HTML]{FFDFBD}{\textcolor{black}{0.34}} & \cellcolor[HTML]{FEC991}{\textcolor{black}{0.58}} & \cellcolor[HTML]{FFD09F}{\textcolor{black}{0.50}} & \cellcolor[HTML]{FECB95}{\textcolor{black}{0.56}} & \cellcolor[HTML]{FFE8D0}{\textcolor{black}{0.25}} & \cellcolor[HTML]{FFE4C7}{\textcolor{black}{0.29}} & \cellcolor[HTML]{FECB95}{\textcolor{black}{0.56}} & \cellcolor[HTML]{C5D1FF}{\textcolor{black}{0.36}}\\
\hspace{1em}RE & \cellcolor[HTML]{F8B467}{\textcolor{black}{0.81}} & \cellcolor[HTML]{FFE0C0}{\textcolor{black}{0.33}} & \cellcolor[HTML]{FFDFBE}{\textcolor{black}{0.34}} & \cellcolor[HTML]{FBBE7C}{\textcolor{black}{0.69}} & \cellcolor[HTML]{FFD1A1}{\textcolor{black}{0.49}} & \cellcolor[HTML]{FDC486}{\textcolor{black}{0.64}} & \cellcolor[HTML]{FFE1C2}{\textcolor{black}{0.32}} & \cellcolor[HTML]{FCC181}{\textcolor{black}{0.67}} & \cellcolor[HTML]{FBBD7A}{\textcolor{black}{0.70}} & \cellcolor[HTML]{AEC1FF}{\textcolor{black}{0.49}}\\
\hspace{1em}VN & \cellcolor[HTML]{FFFFFF}{\textcolor{black}{\supportsn}} & \cellcolor[HTML]{FFFFFF}{\textcolor{black}{\supportsn}} & \cellcolor[HTML]{FFFFFF}{\textcolor{black}{\supportsn}} & \cellcolor[HTML]{FFCB96}{\textcolor{black}{0.55}} & \cellcolor[HTML]{FFE3C5}{\textcolor{black}{0.30}} & \cellcolor[HTML]{FFF4E9}{\textcolor{black}{0.11}} & \cellcolor[HTML]{FFFDFB}{\textcolor{black}{0.02}} & \cellcolor[HTML]{FFE3C7}{\textcolor{black}{0.29}} & \cellcolor[HTML]{FFF2E5}{\textcolor{black}{0.14}} & \cellcolor[HTML]{F2F4FF}{\textcolor{black}{0.08}}\\
\addlinespace[0.3em]
\multicolumn{11}{l}{\textbf{Scorer}}\\
\hspace{1em}BC & \cellcolor[HTML]{9B90C4}{\textcolor{black}{0.98}} & \cellcolor[HTML]{DBD6EA}{\textcolor{black}{0.35}} & \cellcolor[HTML]{CCC5E1}{\textcolor{black}{0.50}} & \cellcolor[HTML]{A196C8}{\textcolor{black}{0.92}} & \cellcolor[HTML]{D8D2E8}{\textcolor{black}{0.38}} & \cellcolor[HTML]{B9B0D6}{\textcolor{black}{0.68}} & \cellcolor[HTML]{F0EEF6}{\textcolor{black}{0.14}} & \cellcolor[HTML]{CDC6E2}{\textcolor{black}{0.49}} & \cellcolor[HTML]{C8C1DF}{\textcolor{black}{0.54}} & \cellcolor[HTML]{BCCAFF}{\textcolor{black}{0.41}}\\
\hspace{1em}CS & \cellcolor[HTML]{9C91C5}{\textcolor{black}{0.97}} & \cellcolor[HTML]{E1DDED}{\textcolor{black}{0.29}} & \cellcolor[HTML]{E9E6F2}{\textcolor{black}{0.21}} & \cellcolor[HTML]{FFFFFF}{\textcolor{black}{\supportsn}} & \cellcolor[HTML]{FFFFFF}{\textcolor{black}{\supportsn}} & \cellcolor[HTML]{FFFFFF}{\textcolor{black}{\supportsn}} & \cellcolor[HTML]{FFFFFF}{\textcolor{black}{\supportsn}} & \cellcolor[HTML]{FFFFFF}{\textcolor{black}{\supportsn}} & \cellcolor[HTML]{FFFFFF}{\textcolor{black}{\supportsn}} & \cellcolor[HTML]{CAD5FF}{\textcolor{black}{0.33}}\\
\hspace{1em}CU & \cellcolor[HTML]{A69CCB}{\textcolor{black}{0.87}} & \cellcolor[HTML]{E2DEEE}{\textcolor{black}{0.28}} & \cellcolor[HTML]{E3DFEF}{\textcolor{black}{0.27}} & \cellcolor[HTML]{9B91C4}{\textcolor{black}{0.98}} & \cellcolor[HTML]{D3CEE5}{\textcolor{black}{0.42}} & \cellcolor[HTML]{B8AFD5}{\textcolor{black}{0.70}} & \cellcolor[HTML]{EEECF5}{\textcolor{black}{0.17}} & \cellcolor[HTML]{B9B0D6}{\textcolor{black}{0.68}} & \cellcolor[HTML]{B7AED5}{\textcolor{black}{0.70}} & \cellcolor[HTML]{BDCBFF}{\textcolor{black}{0.41}}\\
\hspace{1em}DM & \cellcolor[HTML]{9B90C4}{\textcolor{black}{0.98}} & \cellcolor[HTML]{E5E2F0}{\textcolor{black}{0.25}} & \cellcolor[HTML]{CCC6E1}{\textcolor{black}{0.49}} & \cellcolor[HTML]{9E93C6}{\textcolor{black}{0.95}} & \cellcolor[HTML]{E3DFEF}{\textcolor{black}{0.27}} & \cellcolor[HTML]{BBB3D7}{\textcolor{black}{0.66}} & \cellcolor[HTML]{F1EFF7}{\textcolor{black}{0.14}} & \cellcolor[HTML]{DBD6EA}{\textcolor{black}{0.35}} & \cellcolor[HTML]{C1B9DA}{\textcolor{black}{0.61}} & \cellcolor[HTML]{C5D1FF}{\textcolor{black}{0.36}}\\
\hspace{1em}DU & \cellcolor[HTML]{FAF9FC}{\textcolor{black}{0.05}} & \cellcolor[HTML]{F9F8FB}{\textcolor{black}{0.06}} & \cellcolor[HTML]{FAF9FC}{\textcolor{black}{0.05}} & \cellcolor[HTML]{FAF9FC}{\textcolor{black}{0.05}} & \cellcolor[HTML]{E6E3F1}{\textcolor{black}{0.24}} & \cellcolor[HTML]{D3CDE5}{\textcolor{black}{0.43}} & \cellcolor[HTML]{F8F7FB}{\textcolor{black}{0.07}} & \cellcolor[HTML]{ECE9F4}{\textcolor{black}{0.18}} & \cellcolor[HTML]{DED9EC}{\textcolor{black}{0.32}} & \cellcolor[HTML]{F2F4FF}{\textcolor{black}{0.08}}\\
\hspace{1em}KNN & \cellcolor[HTML]{F2F0F7}{\textcolor{black}{0.13}} & \cellcolor[HTML]{FAF9FC}{\textcolor{black}{0.05}} & \cellcolor[HTML]{F4F2F8}{\textcolor{black}{0.11}} & \cellcolor[HTML]{E9E6F2}{\textcolor{black}{0.21}} & \cellcolor[HTML]{DFDBEC}{\textcolor{black}{0.31}} & \cellcolor[HTML]{C1B9DA}{\textcolor{black}{0.61}} & \cellcolor[HTML]{F3F1F8}{\textcolor{black}{0.12}} & \cellcolor[HTML]{F7F6FB}{\textcolor{black}{0.07}} & \cellcolor[HTML]{EEECF5}{\textcolor{black}{0.16}} & \cellcolor[HTML]{ECEFFF}{\textcolor{black}{0.12}}\\
\hspace{1em}LE & \cellcolor[HTML]{FFFFFF}{\textcolor{black}{\supportsn}} & \cellcolor[HTML]{FFFFFF}{\textcolor{black}{\supportsn}} & \cellcolor[HTML]{FFFFFF}{\textcolor{black}{\supportsn}} & \cellcolor[HTML]{C2BADB}{\textcolor{black}{0.60}} & \cellcolor[HTML]{E8E5F2}{\textcolor{black}{0.22}} & \cellcolor[HTML]{D5D0E7}{\textcolor{black}{0.41}} & \cellcolor[HTML]{ECE9F4}{\textcolor{black}{0.19}} & \cellcolor[HTML]{F4F3F9}{\textcolor{black}{0.10}} & \cellcolor[HTML]{F3F2F8}{\textcolor{black}{0.11}} & \cellcolor[HTML]{E2E7FF}{\textcolor{black}{0.18}}\\
\hspace{1em}LS & \cellcolor[HTML]{A297C8}{\textcolor{black}{0.91}} & \cellcolor[HTML]{DFDBEC}{\textcolor{black}{0.31}} & \cellcolor[HTML]{D0C9E3}{\textcolor{black}{0.46}} & \cellcolor[HTML]{FFFFFF}{\textcolor{black}{\supportsn}} & \cellcolor[HTML]{FFFFFF}{\textcolor{black}{\supportsn}} & \cellcolor[HTML]{FFFFFF}{\textcolor{black}{\supportsn}} & \cellcolor[HTML]{FFFFFF}{\textcolor{black}{\supportsn}} & \cellcolor[HTML]{FFFFFF}{\textcolor{black}{\supportsn}} & \cellcolor[HTML]{FFFFFF}{\textcolor{black}{\supportsn}} & \cellcolor[HTML]{B3C4FF}{\textcolor{black}{0.46}}\\
\hspace{1em}MD & \cellcolor[HTML]{F0EEF6}{\textcolor{black}{0.14}} & \cellcolor[HTML]{FCFBFD}{\textcolor{black}{0.03}} & \cellcolor[HTML]{F6F5FA}{\textcolor{black}{0.08}} & \cellcolor[HTML]{F2F1F8}{\textcolor{black}{0.12}} & \cellcolor[HTML]{EFEDF6}{\textcolor{black}{0.16}} & \cellcolor[HTML]{C8C1DF}{\textcolor{black}{0.54}} & \cellcolor[HTML]{F9F8FB}{\textcolor{black}{0.06}} & \cellcolor[HTML]{F8F7FB}{\textcolor{black}{0.07}} & \cellcolor[HTML]{F1EFF7}{\textcolor{black}{0.14}} & \cellcolor[HTML]{F2F4FF}{\textcolor{black}{0.08}}\\
\hspace{1em}PM & \cellcolor[HTML]{F9F8FC}{\textcolor{black}{0.06}} & \cellcolor[HTML]{FAF9FC}{\textcolor{black}{0.05}} & \cellcolor[HTML]{FAF9FC}{\textcolor{black}{0.05}} & \cellcolor[HTML]{FAF9FC}{\textcolor{black}{0.05}} & \cellcolor[HTML]{E7E4F1}{\textcolor{black}{0.23}} & \cellcolor[HTML]{C8C1DF}{\textcolor{black}{0.54}} & \cellcolor[HTML]{F9F8FB}{\textcolor{black}{0.06}} & \cellcolor[HTML]{F3F1F8}{\textcolor{black}{0.12}} & \cellcolor[HTML]{E5E2F0}{\textcolor{black}{0.25}} & \cellcolor[HTML]{F3F5FF}{\textcolor{black}{0.08}}\\
\hspace{1em}WD & \cellcolor[HTML]{FFFFFF}{\textcolor{black}{\supportsn}} & \cellcolor[HTML]{FFFFFF}{\textcolor{black}{\supportsn}} & \cellcolor[HTML]{FFFFFF}{\textcolor{black}{\supportsn}} & \cellcolor[HTML]{C8C1DF}{\textcolor{black}{0.53}} & \cellcolor[HTML]{D7D2E8}{\textcolor{black}{0.39}} & \cellcolor[HTML]{D1CBE4}{\textcolor{black}{0.45}} & \cellcolor[HTML]{EEECF5}{\textcolor{black}{0.16}} & \cellcolor[HTML]{F4F2F9}{\textcolor{black}{0.11}} & \cellcolor[HTML]{F1EFF7}{\textcolor{black}{0.14}} & \cellcolor[HTML]{DFE5FF}{\textcolor{black}{0.20}}\\
\bottomrule
\end{tabular}

        \end{small}
    \endgroup

    \label{tab:best_wide}
\end{table*}

First of all, it can be seen that in datasets with randomly injected noise (\hids{ATIS}, \hids{SST} and \hids{GUM}) errors are easier to find than in aggregated or hand-corrected ones.
Especially in \hids{ATIS}, many algorithms reach close-to-perfect scores, in particular scorer ($> 0.9$ AP).
We attribute this to the artificial noise injected.
The more difficult datasets have usually natural noise patterns that are often harder to solve~\citep{amiriSpottingSpuriousData2018, larsonOutlierDetectionImproved2019, hedderichAnalysingNoiseModel2021}.
The three \hids{Slot Inconsistencies} datasets are also easy compared to \hids{CoNLL-2003}.
On some datasets with real errors --- \hids{Plank} and \hids{Slot Inconsistencies} --- the performance of the best methods is already  quite good with F1 $\approx0.5$ and AP $\approx0.4$ for \hids{Plank} and F1, AP $>0.65$ for \hids{Slot Inconsistencies}.

Overall, methods that work well are \himl{Classification Uncertainty} (CU), \himl{Confident Learning} (CL), \himl{Curriculum Spotter} (CS), \himl{Datamap Confidence} (DM), \himl{Diverse Ensemble} (DE), \himl{Label Aggregation} (LA), \himl{Leitner Spotter} (LS), \himl{Projection Ensemble} (PE), and  \himl{Retag} (RE).
Aggregating scorer judgements via \himl{Borda Count} (BC) can improve performance and deliver the second-best AP score based on the harmonic mean.
The downside here is very high total runtime (the sum of runtimes of individual scores aggregated), as it requires training instances of all scorers beforehand, which already perform very well (H\textsubscript{AP} of \himl{Borda Count} is $0.41$ and the best individual scorer has H\textsubscript{AP} of $0.46$).
While aggregating scores requires well performing scorers (3 in our setup, see~\cref{sec:app_borda}) it is more stable across tasks than using individual methods on their own.
Most model-based methods (\himl{Classification Uncertainty}, \himl{Confident Learning}, \himl{Diverse Ensemble}, \himl{Label Aggregation}, \himl{Retag})  perform very well overall, but methods based on training dynamics that do not need cross-validation (\himl{Curriculum Spotter}, \himl{Datamap Confidence}, \himl{Leitner Spotter}) are on par or better.
Especially \himl{Datamap Confidence} shows a very solid performance and can keep up with the closely related \himl{Classification Uncertainty}, sometimes even outperforming it while not needing \ac{cv}.
\himl{Confident Learning} in particular has high precision for token and span labeling.

\citet{amiriSpottingSpuriousData2018} argue that prediction loss is not enough to detect incorrect instances because easy ones still can have a large loss.
Therefore, more intricate methods like \himl{Leitner Spotter} and \himl{Curriculum Spotter} are needed.
We do not observe a large difference between \himl{Classifier Uncertainty} and the two, though.
\himl{Datamap Confidence} as a more complicated sibling of \himl{Classification Uncertainty}, however, outperforms these from time to time, indicating that training dynamics offers an advantage over simply using class probabilities.

\himl{Variation n-grams} (VN) has high precision and tends to be conservative in flagging items, that is, exhibit low false positives, especially for span classification.
\himl{Weighted Discrepancy} works overall better than \himl{Label Entropy}, but both methods almost always perform worse than more intricate ones.
When manually analyzing their scores, they mostly assign a score of $0.0$ and rarely a different score (less than $10 \%$ from our observation, often even lower).
This is because there are only very few instances with both surface form overlap and different labels.
While the scores for \himl{Prediction Margin} appear to be not good, the original paper~\cite{dligachReducingNeedDouble2011}  reports a similarly low performance while their implementation of \himl{Retag} reaches scores that are around two times higher  (10\% vs. 23\% precision and 38\% vs. 60\% recall).
This is similar to our observations.
One potential reason why \himl{Classification Uncertainty} produces better results than the related \himl{Prediction Margin} is that the latter does not take the given label into account; it always uses the difference between the two most probable classes.
Using a formulation of \himl{Projection Ensemble} that uses the label did not improve results significantly, though.

Methods based on vector proximity --- \himl{k-Nearest Neighbor Entropy} (KNN) and \himl{Mean Distance} (MD) --- perform sub-par across tasks and datasets.
We attribute this to issues in distance calculation for high-dimensional data, as noted for instance by \citet{cuizhuExampleBasedRobustOutlier2005} in a related setting.
In high-dimensional vector spaces, everything can appear equidistant (curse of dimensionality).
Another performance-relevant issue is the embedding quality.
In \citet{grivasNotCuteStroke2020}, \him{KNN} is used with domain-specific embeddings for biomedical texts.
These could have potentially improved performance in their setting, but they do not report quantitative results, though, which makes a comparison difficult.
With regard to \himl{Mean Distance}, we only achieve $H=0.08$.
On real data for intent classification, \citet{larsonOutlierDetectionImproved2019} achieve an average precision of around $0.35$.
They report high recall and good average precision on datasets with random labels but do not report precision on its own.
Their datasets contain mainly paraphrased intents, which makes it potentially easier to achieve good performance.
This is similar to how \ac{aed} applied on our randomly perturbed \hids{ATIS} dataset resulted in high detection scores.
Code and data used in their original publication are not available anymore.
We were therefore not able to reproduce their reported performances with our implementation and on our data.

\himl{Item Response Theory} (IRT) does not perform well across datasets and tends to overly flag instances.
Therefore, it is preferable to use the model predictions in a \himl{Diverse Ensemble} which yields much better performance.
\him{IRT} is also relatively slow for larger corpora as it is optimized via variational inference and needs many iterations to converge.
Our hypothesis is that \himl{Item Response Theory} needs more subjects (in our case models) to better estimate discriminability.
Compared to our very few subjects (seven for text classification and four for token and span labeling), \citet{rodriguezEvaluationExamplesAre2021} used predictions of the SQuAD leaderboard with 161 development and 115 test subjects.
To validate this hypothesis, we rerun \himl{Item Response Theory} on the unaggregated predictions of \himl{Projected Ensemble}.
While this leads to slightly better performances, it is still not working as well as using predictions in \himl{Diverse Ensemble} or \himl{Projected Ensemble} directly.
As it is often unfeasible to have that many models providing predictions, we see \himl{Item Response Theory} only useful in very specific scenarios.

Regarding \himl{Dropout Uncertainty}, after extensive debugging with different models, datasets and formulations of the method, we were not able to achieve comparably good results to other \ac{aed} methods evaluated in this work.
On real data, \citet{amiriSpottingSpuriousData2018} also report relatively low performances similar to ours.
Our implementation delivers results similar to \citet{shelmanovHowCertainYour2021} on misclassification detection.
In their paper, the reported scores appear to be very high.
But we consider their reported scores an overestimate, as they use ROC AUC (which is overconfident for imbalanced datasets) and not AP to evaluate their experiments.
Even when applying the method on debug datasets with the most advantageous conditions that are solvable by other methods with perfect scores, \himl{Dropout Uncertainty} only achieves AP values of around $0.2$.
The main reason we see for the overall low scores for \himl{Dropout Uncertainty} is that the different repeated prediction probabilities are highly correlated and do not differ much overall.
This is similar to the observations of~\citet{shelmanovHowCertainYour2021}.

\paragraph{Qualitative Analysis}

To better understand for which kinds of errors methods work well or fail, we manually analyze the instances in \hids{CoNLL-2003}.
It is our dataset of choice for three reasons: 1) span labeling datasets potentially contain many different errors, 2) it is annotated and corrected by humans and 3) it is quite difficult for \ac{aed} to find errors in, based on our previous evaluation (see \cref{tab:best_wide}).
For spans whose noisy labels disagree with the correction, we annotate them as either being inconsistent, a true error, a incorrect correction, or a hallucinated entity.
Descriptions and examples for each type of error are given in the following:

\begin{description}
    \item[True errors] are labels which are unambiguously incorrect, for instance in the sentence \jexample{NATO military chiefs to visit Iberia},  the entity \jexample{Iberia} was annotated as \jlabel{org} but should be \jlabel{loc}, as it refers to the Iberian peninsula.
    \item[Inconsistencies] are instances which were assigned different labels in similar contexts.
    In \hids{CoNLL-2003}, these are mostly from sport teams that were sometimes annotated as \jlabel{loc} and sometimes as  \jlabel{org}.
    \item[Incorrect correction] In very few cases, the correction introduced a new error, e.g., \jexample{United Nations} was incorrectly corrected from \jlabel{org} to \jlabel{loc}.
    \item[Hallucinated entity] are spans that were labeled to contain an entity, but they should not have been annotated at all. For example, in the sentence \jexample{Returns on treasuries were also in negative territory}, \jexample{treasuries} was annotated as \jlabel{misc} but does not contain a named entity. Sometimes, entities that should consist of one span were annotated originally as two entities. This results in one unmatched entity after alignment. We consider this a hallucinated entity as well.
\end{description}
\begin{figure}[b]
    \centering
    \includegraphics[width=.7\textwidth]{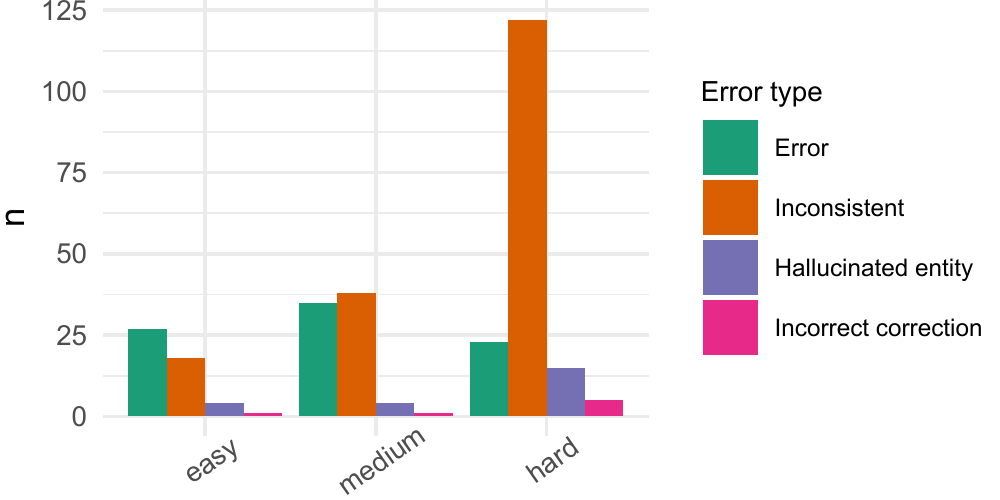}
    \caption{Error counts per difficulty level in \hids{CoNLL-2003}. It can be seen that the number of inconsistencies increases with the difficulty. This indicates that ``real'' annotation errors are easier to detect than inconsistencies.}
    \label{fig:conll2003_error_counts}
\end{figure}
\hids{CoNLL-2003} was corrected manually by \citet{reissIdentifyingIncorrectLabels2020}.
After aligning (see \cref{sec:span_alignment}), we find that there are in total 293 errors.
We group them by difficulty based on how often methods were able to detect them.
For scorers, we consider the instances with the highest 10\% scores as flagged, similarly to how we evaluate precision and recall.
For span labeling, we implemented a total of 16 methods.
The errors detected at least by half of the methods (50\%) are considered \jhi{easy}, the ones deteced by at least four methods (25\%) are considered \jhi{medium}, and the rest, \jhi{hard} (25\%).
This results in 50 easy, 78 medium and 165 hard instances.
The distribution of error types across difficulty levels is visualized in \cref{fig:conll2003_error_counts}.
It can be seen that true errors are easier to detect than inconsistencies by a significant margin: the easy partition consists only of 50\% inconsistencies, whereas in the hard partition, it consists of around 75\% inconsistent instances.
This can be intuitively explained by the fact that the inconsistencies are not rare, but make up a large fraction of all corrections.
It is therefore difficult for a method to learn that it should be flagged when it is only given noisy labels.

We further analyze how each method can deal with the different types of errors across difficulty levels.
The percentage of correctly detected errors per type and method is depicted in \cref{tab:conll2003_error_counts}.
It again can be seen that true errors are easier for methods to detect than inconsistencies;
inconsistencies of hard difficulty were almost never detected.
Interestingly, scorers that are not model-based (\himl{k-Nearest Neighbor Entropy} (KNN), \himl{Label Entropy} (LE) and \himl{Weighted Discrepancy} (WD)) are able to better detect inconsistencies of  medium and sometimes hard difficulty but fail at detecting most errors.
We explain this for \him{KNN} by the fact that it relies on semantic vector space embeddings that do not rely on the noisy label but on the semantics of its surface form in its context.
As neighbors in the space have the same meaning, it is possible to detect errors even though the label is inconsistent in many cases.
The same can be said about \him{WD} and \him{LE} which rely only on the surface form and how often it is annotated differently.
If the correct label is the majority, then it can still detect inconsistencies even if they are quite frequent.
But both still do not perform as well as other methods on easier instances, they only find 50\% of errors and inconsistencies whereas \himl{Classification Uncertainty} or \himl{Retag} detects almost all (but again fail to find inconsistencies on medium difficulty).
\himl{Variation n-grams} (VN), however, do not work well even for easy cases because it relies on contexts around annotations that need to match exactly, which is very rare in this dataset.

\begin{table*}[htb]
    \centering
    \caption{\% of errors and inconsistencies detected on \hids{CoNLL-2003} across methods and difficulty for flaggers \flaggercircle and scorers \scorercircle grouped by error types. It can be seen that real errors (E) are more often detected than inconsistencies (I). Some methods not relying on models (KNN, LE, WD) are sometimes better in spotting inconsistencies than errors, whereas for model-based method it is the opposite. Note that errors concerning incorrect corrections (IC) and hallucinated entities (HE) are quite rare and not reliable to draw conclusions from.}
    \begingroup
    \renewcommand*{\arraystretch}{1.2}
    \renewcommand*{\tabcolsep}{4pt}
    \begin{small}
        
\begin{tabular}{ccccccccccccccccc}
\toprule
\multicolumn{1}{c}{ } & \multicolumn{7}{c}{Flagger} & \multicolumn{9}{c}{Scorer} \\
\cmidrule(l{3pt}r{3pt}){2-8} \cmidrule(l{3pt}r{3pt}){9-17}
Error & CL & DE & IRT & LA & PE & RE & VN & BC & CU & DM & DU & KNN & LE & MD & WD & PM\\
\midrule
\addlinespace[0.3em]
\multicolumn{17}{l}{\textbf{Easy}}\\
\hspace{1em}E & \cellcolor[HTML]{FCC181}{66} & \cellcolor[HTML]{F2A648}{96} & \cellcolor[HTML]{F1A340}{100} & \cellcolor[HTML]{F1A340}{100} & \cellcolor[HTML]{F1A340}{100} & \cellcolor[HTML]{F1A340}{100} & \cellcolor[HTML]{FFFBF8}{3} & \cellcolor[HTML]{A196C7}{92} & \cellcolor[HTML]{998EC3}{100} & \cellcolor[HTML]{BBB2D7}{66} & \cellcolor[HTML]{E4E1EF}{25} & \cellcolor[HTML]{D5D0E7}{40} & \cellcolor[HTML]{E1DCED}{29} & \cellcolor[HTML]{F0EEF6}{14} & \cellcolor[HTML]{E1DCED}{29} & \cellcolor[HTML]{E4E1EF}{25}\\
\hspace{1em}I & \cellcolor[HTML]{FBBC76}{72} & \cellcolor[HTML]{F1A340}{100} & \cellcolor[HTML]{F1A340}{100} & \cellcolor[HTML]{F5AD57}{88} & \cellcolor[HTML]{F3A84B}{94} & \cellcolor[HTML]{F3A84B}{94} & \cellcolor[HTML]{FFF4E9}{11} & \cellcolor[HTML]{9F94C6}{94} & \cellcolor[HTML]{998EC3}{100} & \cellcolor[HTML]{C6BFDE}{55} & \cellcolor[HTML]{E2DEEE}{27} & \cellcolor[HTML]{C1B9DA}{61} & \cellcolor[HTML]{C6BFDE}{55} & \cellcolor[HTML]{F4F2F8}{11} & \cellcolor[HTML]{C6BFDE}{55} & \cellcolor[HTML]{E2DEEE}{27}\\
\hspace{1em}HE & \cellcolor[HTML]{FFE7CF}{25} & \cellcolor[HTML]{F1A340}{100} & \cellcolor[HTML]{F1A340}{100} & \cellcolor[HTML]{F1A340}{100} & \cellcolor[HTML]{F1A340}{100} & \cellcolor[HTML]{F1A340}{100} & \cellcolor[HTML]{FFFFFF}{0} & \cellcolor[HTML]{998EC3}{100} & \cellcolor[HTML]{998EC3}{100} & \cellcolor[HTML]{998EC3}{100} & \cellcolor[HTML]{998EC3}{100} & \cellcolor[HTML]{FFFFFF}{0} & \cellcolor[HTML]{FFFFFF}{0} & \cellcolor[HTML]{CCC5E1}{50} & \cellcolor[HTML]{FFFFFF}{0} & \cellcolor[HTML]{998EC3}{100}\\
\hspace{1em}IC & \cellcolor[HTML]{FFFFFF}{0} & \cellcolor[HTML]{F1A340}{100} & \cellcolor[HTML]{F1A340}{100} & \cellcolor[HTML]{F1A340}{100} & \cellcolor[HTML]{F1A340}{100} & \cellcolor[HTML]{F1A340}{100} & \cellcolor[HTML]{FFFFFF}{0} & \cellcolor[HTML]{998EC3}{100} & \cellcolor[HTML]{998EC3}{100} & \cellcolor[HTML]{998EC3}{100} & \cellcolor[HTML]{FFFFFF}{0} & \cellcolor[HTML]{FFFFFF}{0} & \cellcolor[HTML]{FFFFFF}{0} & \cellcolor[HTML]{998EC3}{100} & \cellcolor[HTML]{FFFFFF}{0} & \cellcolor[HTML]{FFFFFF}{0}\\
\addlinespace[0.3em]
\multicolumn{17}{l}{\textbf{Medium}}\\
\hspace{1em}E & \cellcolor[HTML]{FFD9B3}{40} & \cellcolor[HTML]{FFCF9D}{51} & \cellcolor[HTML]{FFCC98}{54} & \cellcolor[HTML]{F5AD57}{88} & \cellcolor[HTML]{F7B262}{82} & \cellcolor[HTML]{F7B262}{82} & \cellcolor[HTML]{FFFFFF}{0} & \cellcolor[HTML]{DFDAEC}{31} & \cellcolor[HTML]{9F94C6}{94} & \cellcolor[HTML]{DFDAEC}{31} & \cellcolor[HTML]{DCD7EA}{34} & \cellcolor[HTML]{E8E4F1}{22} & \cellcolor[HTML]{FFFFFF}{0} & \cellcolor[HTML]{F3F2F8}{11} & \cellcolor[HTML]{FFFFFF}{0} & \cellcolor[HTML]{E5E1F0}{25}\\
\hspace{1em}I & \cellcolor[HTML]{FFF3E6}{13} & \cellcolor[HTML]{FFE6CC}{26} & \cellcolor[HTML]{FFE1C2}{31} & \cellcolor[HTML]{FFE6CC}{26} & \cellcolor[HTML]{FFCE9B}{52} & \cellcolor[HTML]{FFE1C2}{31} & \cellcolor[HTML]{FFFFFF}{0} & \cellcolor[HTML]{EFECF6}{15} & \cellcolor[HTML]{D9D4E9}{36} & \cellcolor[HTML]{E7E3F1}{23} & \cellcolor[HTML]{E9E6F2}{21} & \cellcolor[HTML]{D1CBE4}{44} & \cellcolor[HTML]{BEB6D9}{63} & \cellcolor[HTML]{EFECF6}{15} & \cellcolor[HTML]{BEB6D9}{63} & \cellcolor[HTML]{F1EFF7}{13}\\
\hspace{1em}HE & \cellcolor[HTML]{FFFFFF}{0} & \cellcolor[HTML]{FAB971}{75} & \cellcolor[HTML]{FAB971}{75} & \cellcolor[HTML]{FFE7CF}{25} & \cellcolor[HTML]{FAB971}{75} & \cellcolor[HTML]{F1A340}{100} & \cellcolor[HTML]{FFFFFF}{0} & \cellcolor[HTML]{B2A9D2}{75} & \cellcolor[HTML]{CCC5E1}{50} & \cellcolor[HTML]{B2A9D2}{75} & \cellcolor[HTML]{E5E2F0}{25} & \cellcolor[HTML]{FFFFFF}{0} & \cellcolor[HTML]{FFFFFF}{0} & \cellcolor[HTML]{FFFFFF}{0} & \cellcolor[HTML]{FFFFFF}{0} & \cellcolor[HTML]{CCC5E1}{50}\\
\hspace{1em}IC & \cellcolor[HTML]{F1A340}{100} & \cellcolor[HTML]{F1A340}{100} & \cellcolor[HTML]{F1A340}{100} & \cellcolor[HTML]{F1A340}{100} & \cellcolor[HTML]{F1A340}{100} & \cellcolor[HTML]{F1A340}{100} & \cellcolor[HTML]{FFFFFF}{0} & \cellcolor[HTML]{FFFFFF}{0} & \cellcolor[HTML]{998EC3}{100} & \cellcolor[HTML]{FFFFFF}{0} & \cellcolor[HTML]{FFFFFF}{0} & \cellcolor[HTML]{FFFFFF}{0} & \cellcolor[HTML]{FFFFFF}{0} & \cellcolor[HTML]{FFFFFF}{0} & \cellcolor[HTML]{FFFFFF}{0} & \cellcolor[HTML]{FFFFFF}{0}\\
\addlinespace[0.3em]
\multicolumn{17}{l}{\textbf{Hard}}\\
\hspace{1em}E & \cellcolor[HTML]{FFFFFF}{0} & \cellcolor[HTML]{FFEFDD}{17} & \cellcolor[HTML]{FFF3E6}{13} & \cellcolor[HTML]{FFFFFF}{0} & \cellcolor[HTML]{FCC283}{65} & \cellcolor[HTML]{FFFFFF}{0} & \cellcolor[HTML]{FFFFFF}{0} & \cellcolor[HTML]{FFFFFF}{0} & \cellcolor[HTML]{FFFFFF}{0} & \cellcolor[HTML]{EDEBF5}{17} & \cellcolor[HTML]{FFFFFF}{0} & \cellcolor[HTML]{EDEBF5}{17} & \cellcolor[HTML]{FFFFFF}{0} & \cellcolor[HTML]{EDEBF5}{17} & \cellcolor[HTML]{FFFFFF}{0} & \cellcolor[HTML]{F2F0F7}{13}\\
\hspace{1em}I & \cellcolor[HTML]{FFFFFF}{0} & \cellcolor[HTML]{FFFEFD}{0} & \cellcolor[HTML]{FFFEFD}{0} & \cellcolor[HTML]{FFFFFF}{0} & \cellcolor[HTML]{FFF6EE}{9} & \cellcolor[HTML]{FFFFFF}{0} & \cellcolor[HTML]{FFFFFF}{0} & \cellcolor[HTML]{FCFBFD}{3} & \cellcolor[HTML]{FEFEFF}{0} & \cellcolor[HTML]{F9F8FC}{5} & \cellcolor[HTML]{FBFAFD}{4} & \cellcolor[HTML]{F0EEF6}{14} & \cellcolor[HTML]{FAF9FC}{4} & \cellcolor[HTML]{FDFDFE}{1} & \cellcolor[HTML]{FAF9FC}{4} & \cellcolor[HTML]{FBFAFD}{4}\\
\hspace{1em}HE & \cellcolor[HTML]{FFFFFF}{0} & \cellcolor[HTML]{FFF2E5}{13} & \cellcolor[HTML]{FFF2E5}{13} & \cellcolor[HTML]{FFFFFF}{0} & \cellcolor[HTML]{FFCD9A}{53} & \cellcolor[HTML]{FFF9F2}{6} & \cellcolor[HTML]{FFFFFF}{0} & \cellcolor[HTML]{FFFFFF}{0} & \cellcolor[HTML]{FFFFFF}{0} & \cellcolor[HTML]{FFFFFF}{0} & \cellcolor[HTML]{F8F7FB}{6} & \cellcolor[HTML]{FFFFFF}{0} & \cellcolor[HTML]{FFFFFF}{0} & \cellcolor[HTML]{F8F7FB}{6} & \cellcolor[HTML]{FFFFFF}{0} & \cellcolor[HTML]{FFFFFF}{0}\\
\hspace{1em}IC & \cellcolor[HTML]{FFFFFF}{0} & \cellcolor[HTML]{FFECD8}{20} & \cellcolor[HTML]{FFECD8}{20} & \cellcolor[HTML]{FFECD8}{20} & \cellcolor[HTML]{FFECD8}{20} & \cellcolor[HTML]{FFFFFF}{0} & \cellcolor[HTML]{FFFFFF}{0} & \cellcolor[HTML]{FFFFFF}{0} & \cellcolor[HTML]{FFFFFF}{0} & \cellcolor[HTML]{FFFFFF}{0} & \cellcolor[HTML]{EAE8F3}{20} & \cellcolor[HTML]{FFFFFF}{0} & \cellcolor[HTML]{FFFFFF}{0} & \cellcolor[HTML]{EAE8F3}{20} & \cellcolor[HTML]{FFFFFF}{0} & \cellcolor[HTML]{FFFFFF}{0}\\
\bottomrule
\end{tabular}

    \end{small}
    \endgroup

    \label{tab:conll2003_error_counts}
\end{table*}

\noindent
To summarize, the methods that worked best overall across tasks and datasets are \himl{Borda Count} (BC), \himl{Diverse Ensemble} (DE), \himl{Label Aggregation} (LA), and  \himl{Retag} (RE).
Inconsistencies appear to be more difficult to detect for most methods, especially for model-based ones.
Methods that do not rely on the noisy labels like \himl{k-Nearest Neighbor Entropy}, \himl{Label Entropy} and \himl{Weighted Discrepancy} were better in finding inconsistencies on more difficult instances when manually analyzing \hids{CoNLL-2003}.
\FloatBarrier

\subsection{RQ2 -- Does model calibration improve model-based method performance?}
\label{sec:calibration}

Several model-based \ac{aed} methods, for instance \himl{Classification Uncertainty}, directly leverage probability estimates provided by a machine learning model (\cref{sec:model_based}).
Therefore, it is of interest whether models output class probability distributions which are accurate.
For instance, if a model predicts 100 instances and states for all 80\% confidence, then the accuracy should be around $0.8$.
If this is the case for a model, then it is called \emph{calibrated}.
Previous works have shown that models are often not calibrated very well, especially neural networks~\citep{guoCalibrationModernNeural2017}.
To alleviate this issue, a number of calibration algorithms have been developed.
The most common approaches are post-hoc which means that they are applied after the model has already been trained.

Probabilities which are an under- or overestimate can lead to non-optimal \ac{aed} results.
The question arises whether model-based \ac{aed} methods can benefit from calibration and if so, to what extent.
We are only aware of one work mentioning calibration in context of annotation error detection.
\citet{northcuttConfidentLearningEstimating2021} claim that their approach does not require calibration to work well, but they did not evaluate it in detail.
We only evaluate whether calibration helps for approaches that directly use probabilities and can leverage \ac{cv}, as calibration needs to be trained on a holdout set.
This, for instance, excludes \himl{Curriculum Spotter}, \himl{Leitner Spotter}, and \himl{Datamap Confidence}.
Methods that can benefit are \himl{Confident Learning}, \himl{Classifier Uncertainty}, \himl{Dropout Uncertainty} and \himl{Prediction Margin}.

There are two groups of approaches for post-hoc calibration: \jhi{parametric} (e.g. Platt Scaling/Logistic Calibration~\citep{plattProbabilisticOutputsSupport1999} or Temperature Scaling~\citep{guoCalibrationModernNeural2017}) or \jhi{non-parametric} (e.g. Histogram Binning~\citep{zadroznyObtainingCalibratedProbability2001}, Isotonic Regression~\citep{zadroznyTransformingClassifierScores2002} or Bayesian Binning into Quantiles~\citep{naeiniObtainingWellCalibrated2015}).
On a holdout corpus we evaluate several calibration methods to determine which calibration method to use (see \cref{sec:appendix_calibration}).
As a result, we apply the best --- Platt Scaling --- for all experiments that leverage calibration.

Calibration is normally trained on a holdout set.
As we already perform cross-validation, we use the holdout set both for training the calibration and for predicting annotation errors.
While this would not be optimal if we are interested in generalizing calibrated probabilities to unseen data, we are more interested in downstream task performance.
Using an additional fold per round would be theoretically more sound.
But our preliminary experiments show that it has the issue of reducing the available training data and thereby hurts the error detection performance more than the calibration helps.
Using the same fold for both calibration and applying \ac{aed}, however, improves overall task performance which is what matters in our special task setting.
We do not leak the values for the downstream tasks (whether an instance is labeled incorrectly or not) but only the labels for the primary task.

\begin{figure}[t]
    \begin{subfigure}{.99\textwidth}
        \centering
        \includegraphics[width=\linewidth]{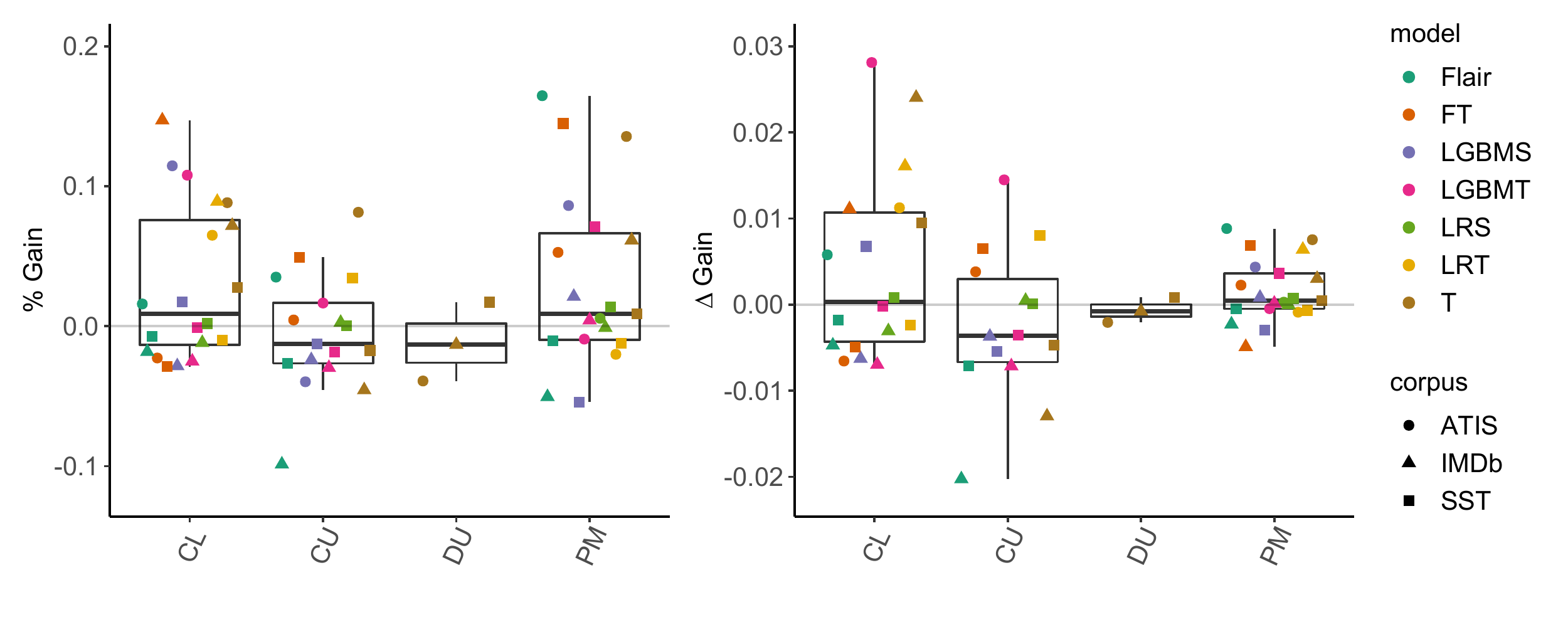}
        \caption{Text classification}
    \end{subfigure}
    \begin{subfigure}{.99\textwidth}
        \centering
        \includegraphics[width=\linewidth]{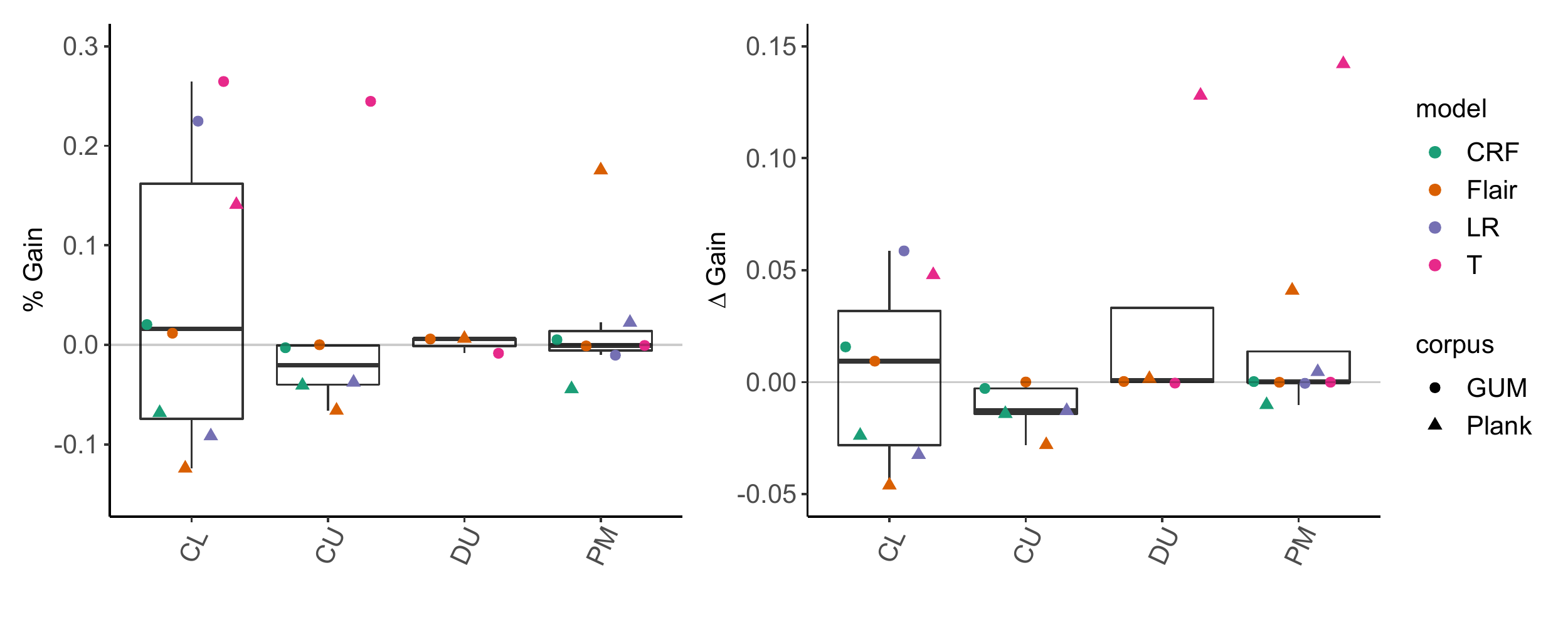}
        \caption{Token labeling}
    \end{subfigure}
    \begin{subfigure}{.99\textwidth}
        \centering
        \includegraphics[width=\linewidth]{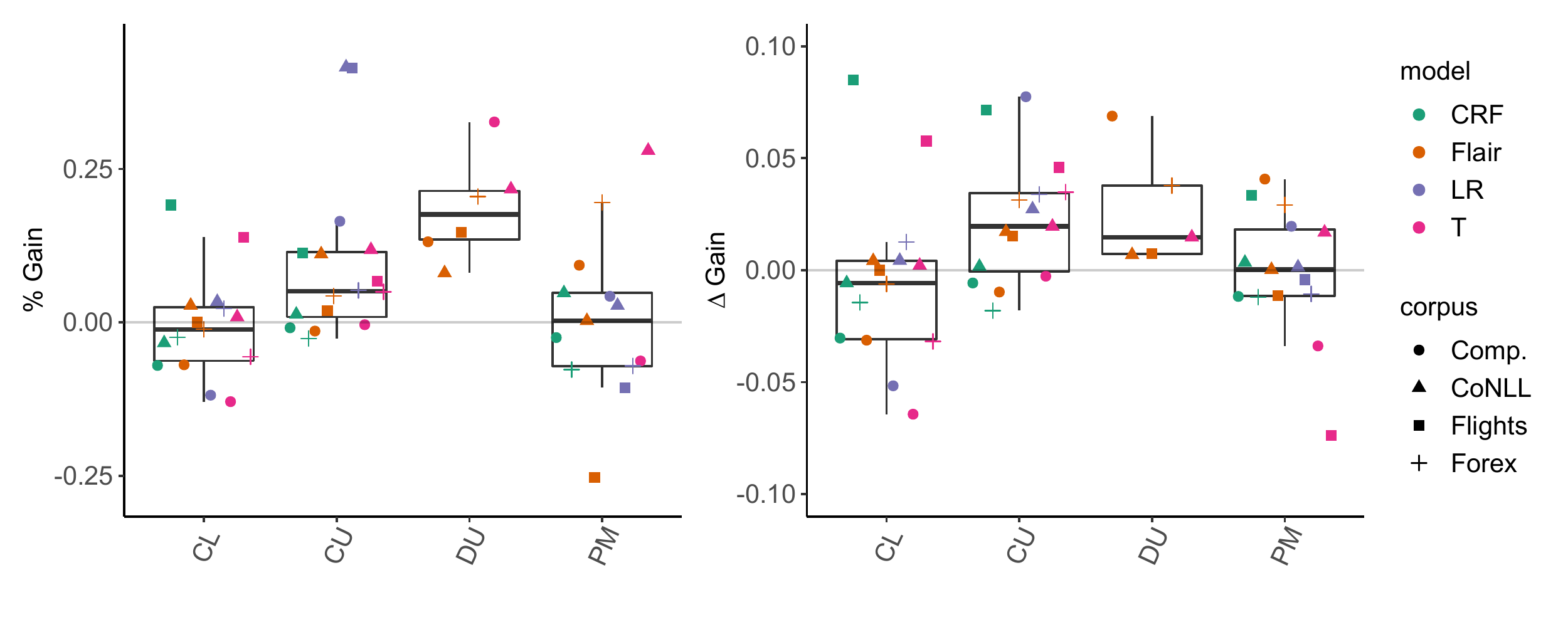}
        \caption{Span labeling}
    \end{subfigure}
    \caption{Relative and total improvement of model-based \ac{aed} methods over different corpora, methods, and models when calibrating probabilities. It can be seen that calibration can lead to good improvements, while on median mostly not hurting performance. This plot is best viewed in the electronic version of this paper. Not displayed are extreme (positive) outlier points.}
    \label{fig:r1_calibration}
\end{figure}

To evaluate whether calibration helps model-based methods that leverage probabilities, we train models with cross-validation and then evaluate each applicable method with and without calibration.
The same model and therefore the same initial probabilities are used for both.
We measure the relative and total improvement in F1 (for flaggers) and AP (for scorers), which are our main metrics.
The results are depicted in \cref{fig:r1_calibration}.
It can be seen that calibration has the potential of improving the performance of certain methods by quite a large margin.
For \himl{Confident Learning}, the absolute gain is up to 3 percentage points (pp) F1 on text classification, 5 pp for token labeling and up to 10 pp for span labeling.
On the latter two tasks, though, there are also many cases with  performance reductions.
A similar pattern can be seen for \himl{Classification Uncertainty} with up to 2 pp, no impact and up to 8 pp, respectively.
\himl{Dropout Uncertainty} and \himl{Prediction Margin} do not perform well to begin with.
But after calibration, they gain 5 to 10 pp AP especially for span and in some instances for token labeling.
In most cases on median, calibration does not hurt the overall performance.

In order to check whether the improvement using calibration is statistically significant, we also employ statistical testing.
We choose the Wilcoxon signed-rank test~\citep{wilcoxonIndividualComparisonsRanking1945} because the data is not normally distributed which is required by the more powerful paired t-test.
The alternative hypothesis is that calibration improves method performance, resulting in a one-sided test.

We do not perform a multiple-comparison correction as each experiment works on different data.
The p-values can be seen in \cref{tab:calibration_wilcox}.
It can be seen that calibration can improve performance significantly overall in two task and method combinations (text classification + \himl{Confident Learning} and span labeling + \himl{Classification Uncertainty}).
For text classification and token labeling, the absolute gain is relatively small.
For span labeling, \himl{Classification Uncertainty} benefits the most.
The gains for \himl{Dropout Uncertainty} and \himl{Prediction Margins} appear large, but these methods do not perform well in the first place.
Hence, our conclusion is that calibration can help model-based \ac{aed} performance but it is very task and dataset specific.

We do not see a clear tendency which models or datasets benefit the most from calibration.
More investigation is needed regarding which calibrator works best for which model and task.
We chose the one that reduces the calibration error the most which is not necessarily the best choice for each setting.

\begin{table}[t]
    \centering
    \caption{p-values for Wilcoxon signed-rank test. We check whether calibration improves AED performance on a statistically significant level.
        Underlined values are significant with $p < 0.05$. }
    
\begin{tabular}{llll}
\toprule
Method & Text & Token & Span\\
\midrule
Confident Learning (CL) & \underline{0.021} & 0.230 & 0.665\\
Classification Uncertainty (CU) & 0.121 & 0.320 & \underline{0.003}\\
Dropout Uncertainty (DU) & 0.750 & 0.188 & 0.320\\
Prediction Margin (PM) & 0.064 & 0.273 & 0.628\\
\bottomrule
\end{tabular}

    \label{tab:calibration_wilcox}
\end{table}

\clearpage

\subsection{RQ3 -- To what extent are model and detection performance correlated?}
\label{sec:r2_relationship}

Several \ac{aed} methods directly use model predictions or probabilities to detect potential annotation errors.
This raises the question of how model performance impacts \ac{aed} performance.
\citet{reissIdentifyingIncorrectLabels2020} state that they deliberately use simpler models to find more potential errors in \hids{CoNLL-2003} and therefore developed \himl{Projection Ensemble}, an ensemble of logistic regression classifiers that use BERT embeddings reduced by different Gaussian projections.
Their motivation is to obtain a diverse collection of predictions to have disagreements.
They conjecture that using very well-performing models might be detrimental to \ac{aed} performance as their predictions potentially would not differ that much from the noisy labels as the models learned predicting the noise.
In contrast to that, \citet{barnesSentimentAnalysisNot2019} use state-of-the-art models to find annotation errors in different sentiment datasets.
But neither \citet{reissIdentifyingIncorrectLabels2020} nor \citet{barnesSentimentAnalysisNot2019} directly evaluates AED performance but rather, use AED to clean noisy datasets for which the gold labels are unknown.
Therefore, the question of how much model and detection performance are correlated has not yet been thoroughly evaluated.

\begin{figure}[ht]
    \centering
    \subfloat[Text classification]{\includegraphics[width=.9\linewidth]{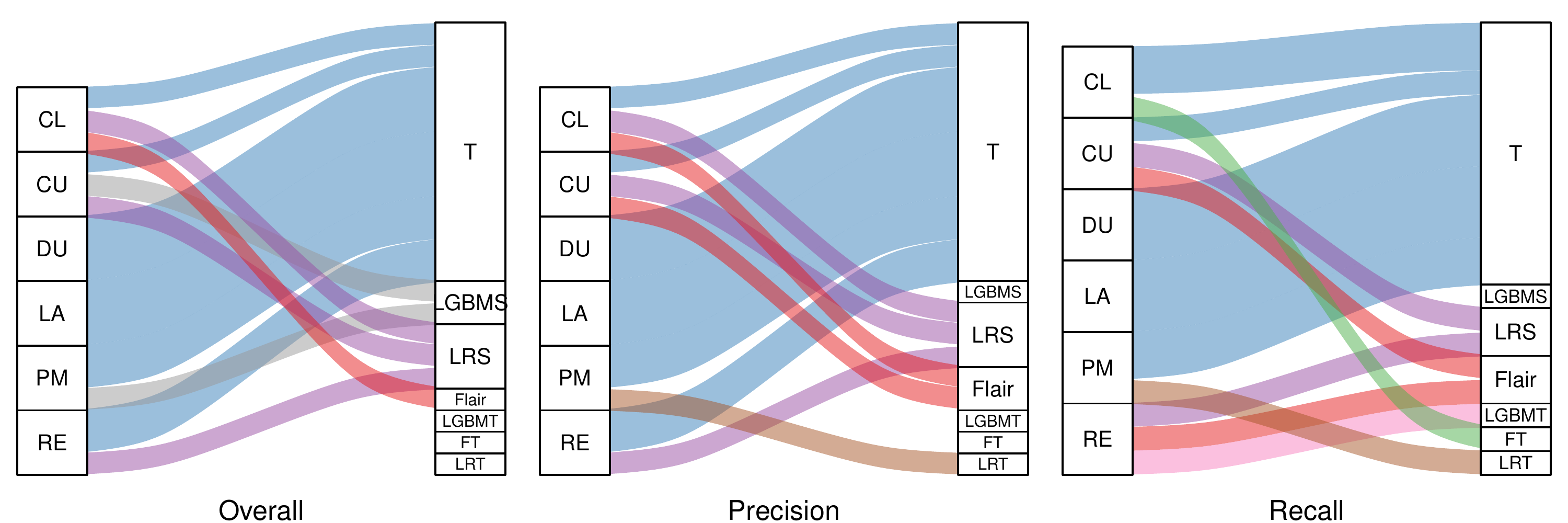}} \\
    \subfloat[Token labeling]{\includegraphics[width=.9\linewidth]{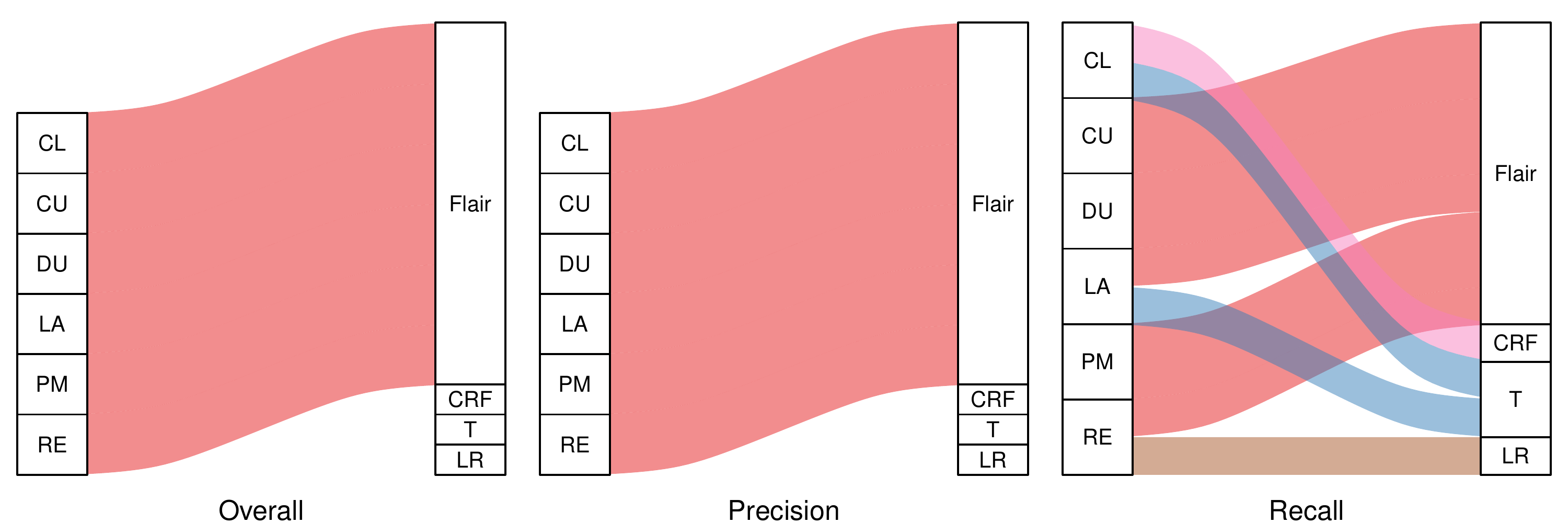}} \\
    \subfloat[Span labeling]{\includegraphics[width=.9\linewidth]{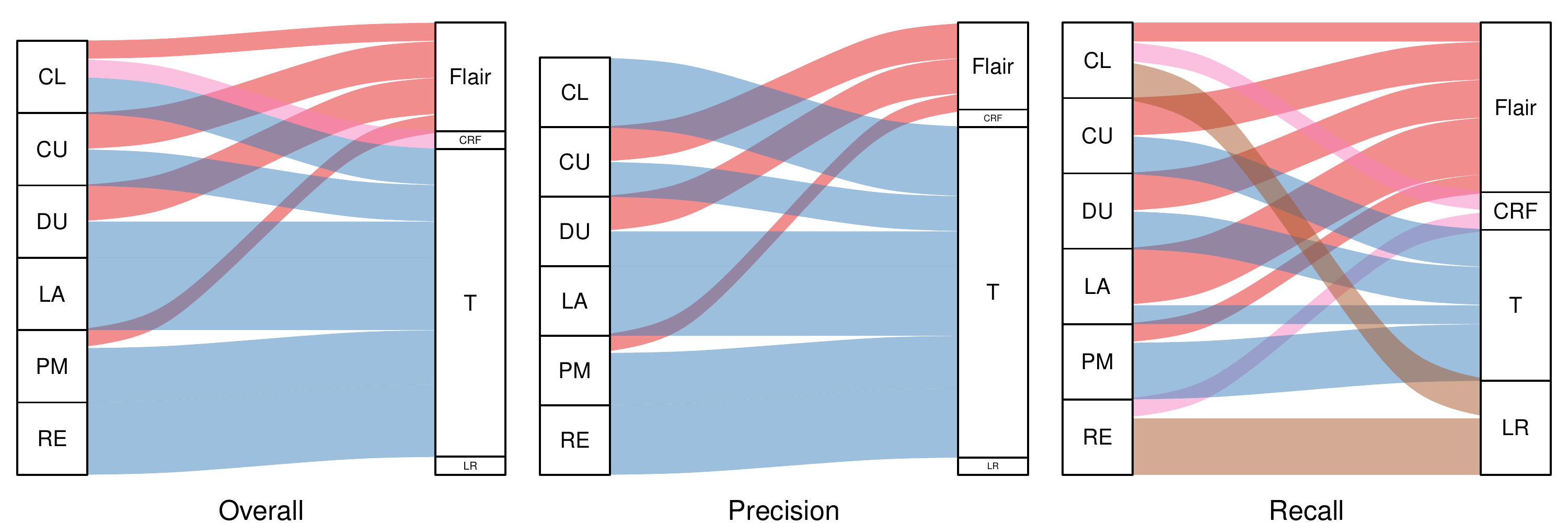}} \\
    \caption{Model-based methods and how often which model type leads to the best method performance with respect to overall, precision, and recall score.
        A connection from left to right between a method and a model indicates that using that method with outputs from that model leads to the best task performance.
        The color of the connection indicates the chosen model, for instance, Flair is \modelcircle{red}, Transformer \modelcircle{jpositive}.
        The model axis is presented in descending order by model performance, aggregated by Borda count across datasets. This figure is best viewed in color.}
    \label{fig:r2_alluvial}
\end{figure}

For answering this question, we leverage the fact that we implemented several models of varying performance for each task.
We use two complementary approaches to analyze this question.
First, we measure the correlation between model and task performances for the overall score, precision and recall.
Then, we analyze which models lead to the best \ac{aed} performance.

Throughout this section, scores for flaggers and scorers are coalesced; overall score corresponds to F1 and AP, precision to precision and precision@10\%, recall to recall and recall@10\%.
For reference, model performances are shown in \cref{fig:appendix_r2_model_performance_heatmap}.
We choose micro aggregation for measuring model performance, as we are interested in the overall  scores and not the scores per class.
Using macro aggregation yields qualitatively similar but less significant results.

\paragraph{Correlation} In order to determine whether there exists a positive or negative relationship between model and method performances, we compute Kendall's \texttau{} coefficient~\citep{kendallNewMeasureRank1938} for each method and dataset.
The results are depicted in \cref{tab:r2_kendall_tau}.
We see that when the test is significant with $p < 0.05$ then there is almost always a moderate to strong monotonic relationship.\footnote{$|\tau| > 0.07$ indicates a weak, $|\tau| > 0.21$ a moderate, $|\tau| > 0.35$ indicates a strong monotonic relation.}
\texttau{} is zero or positive for classification and token labeling, hinting that there is either no relationship or a positive one.
For span labeling we observe  negative correlation for precision and overall.
It is significant in one case only.

One issue with this test is its statistical power.
In our setting, it is quite low due to the few samples available per method and task.
It is therefore likely that the  null hypothesis --- in our case, the assumption that there is no relationship between model and method performances --- is not rejected even if it should have been.
Hence, we next perform additional analysis to see which models overall lead to the best model performances.

\begin{table}[ht]
    \centering
    \caption{Kendall's \texttau{} coefficient grouped by task and method measured across datasets.
    For $p$, the null hypothesis is $\tau = 0$ and the  alternative hypotheses is $\tau \neq 0$. Underlined are significant p-values with $p < 0.05$. Positive correlation is highlighted \positivecircle, negative correlation is highlighted \negativecircle.}
    \begin{small}
    \begingroup
    \renewcommand*{\arraystretch}{1.2}
        
\begin{tabular}{cSSSSSS}
\toprule
\multicolumn{1}{c}{ } & \multicolumn{2}{c}{Overall} & \multicolumn{2}{c}{Precision} & \multicolumn{2}{c}{Recall} \\
\cmidrule(l{3pt}r{3pt}){2-3} \cmidrule(l{3pt}r{3pt}){4-5} \cmidrule(l{3pt}r{3pt}){6-7}
Method & {\texttau} & {$p$} & {\texttau} & {$p$} & {\texttau} & {$p$}\\
\midrule
\addlinespace[0.3em]
\multicolumn{7}{l}{\textbf{Text}}\\
\hspace{1em}CL & \cellcolor[HTML]{BCCAFF}{\textcolor{black}{+0.495}} & \underline{0.002} & \cellcolor[HTML]{A3B9FF}{\textcolor{black}{+0.657}} & \underline{0.000} & \cellcolor[HTML]{E14F72}{\textcolor{black}{-0.373}} & \underline{0.018}\\
\hspace{1em}CU & \cellcolor[HTML]{BDCBFF}{\textcolor{black}{+0.486}} & \underline{0.002} & \cellcolor[HTML]{CAD4FF}{\textcolor{black}{+0.396}} & \underline{0.013} & \cellcolor[HTML]{6699FF}{\textcolor{black}{+0.705}} & \underline{0.000}\\
\hspace{1em}DU & \cellcolor[HTML]{D3DBFF}{\textcolor{black}{+0.333}} & 0.602 & \cellcolor[HTML]{D3DBFF}{\textcolor{black}{+0.333}} & 0.602 & \cellcolor[HTML]{C0CDFF}{\textcolor{black}{+0.333}} & 0.602\\
\hspace{1em}LA & \cellcolor[HTML]{D3DBFF}{\textcolor{black}{+0.333}} & 0.602 & \cellcolor[HTML]{6398FF}{\textcolor{black}{+1.000}} & 0.117 & \cellcolor[HTML]{C0CDFF}{\textcolor{black}{+0.333}} & 0.602\\
\hspace{1em}PM & \cellcolor[HTML]{ECEFFF}{\textcolor{black}{+0.143}} & 0.365 & \cellcolor[HTML]{E3E8FF}{\textcolor{black}{+0.211}} & 0.184 & \cellcolor[HTML]{D4DCFF}{\textcolor{black}{+0.230}} & 0.147\\
\hspace{1em}RE & \cellcolor[HTML]{B0C2FF}{\textcolor{black}{+0.571}} & \underline{0.000} & \cellcolor[HTML]{ACBFFF}{\textcolor{black}{+0.600}} & \underline{0.000} & \cellcolor[HTML]{B0C2FF}{\textcolor{black}{+0.412}} & \underline{0.011}\\
\addlinespace[0.3em]
\multicolumn{7}{l}{\textbf{Token}}\\
\hspace{1em}CL & \cellcolor[HTML]{739FFF}{\textcolor{black}{+0.929}} & \underline{0.001} & \cellcolor[HTML]{739FFF}{\textcolor{black}{+0.929}} & \underline{0.001} & \cellcolor[HTML]{D7DFFF}{\textcolor{black}{+0.214}} & 0.458\\
\hspace{1em}CU & \cellcolor[HTML]{9AB4FF}{\textcolor{black}{+0.714}} & \underline{0.013} & \cellcolor[HTML]{8EADFF}{\textcolor{black}{+0.786}} & \underline{0.006} & \cellcolor[HTML]{6398FF}{\textcolor{black}{+0.714}} & \underline{0.013}\\
\hspace{1em}DU & \cellcolor[HTML]{EC7A8F}{\textcolor{black}{-0.333}} & 0.497 & \cellcolor[HTML]{F7A8B3}{\textcolor{black}{-0.333}} & 0.497 & \cellcolor[HTML]{C0CDFF}{\textcolor{black}{+0.333}} & 0.497\\
\hspace{1em}LA & \cellcolor[HTML]{6398FF}{\textcolor{black}{+1.000}} & \underline{0.042} & \cellcolor[HTML]{A2B9FF}{\textcolor{black}{+0.667}} & 0.174 & \cellcolor[HTML]{729EFF}{\textcolor{black}{+0.667}} & 0.174\\
\hspace{1em}PM & \cellcolor[HTML]{FFFFFF}{\textcolor{black}{+0.000}} & 1.000 & \cellcolor[HTML]{FFFFFF}{\textcolor{black}{+0.000}} & 1.000 & \cellcolor[HTML]{FFFFFF}{\textcolor{black}{+0.000}} & 1.000\\
\hspace{1em}RE & \cellcolor[HTML]{81A6FF}{\textcolor{black}{+0.857}} & \underline{0.003} & \cellcolor[HTML]{9AB4FF}{\textcolor{black}{+0.714}} & \underline{0.013} & \cellcolor[HTML]{BBCAFF}{\textcolor{black}{+0.357}} & 0.216\\
\addlinespace[0.3em]
\multicolumn{7}{l}{\textbf{Span}}\\
\hspace{1em}CL & \cellcolor[HTML]{FFF2F3}{\textcolor{black}{-0.033}} & 0.857 & \cellcolor[HTML]{FDCFD5}{\textcolor{black}{-0.183}} & 0.322 & \cellcolor[HTML]{EDF0FF}{\textcolor{black}{+0.101}} & 0.588\\
\hspace{1em}CU & \cellcolor[HTML]{FDFDFF}{\textcolor{black}{+0.017}} & 0.928 & \cellcolor[HTML]{FABEC6}{\textcolor{black}{-0.250}} & 0.177 & \cellcolor[HTML]{C6D2FF}{\textcolor{black}{+0.300}} & 0.105\\
\hspace{1em}DU & \cellcolor[HTML]{E14F72}{\textcolor{black}{-0.429}} & 0.138 & \cellcolor[HTML]{F18E9E}{\textcolor{black}{-0.429}} & 0.138 & \cellcolor[HTML]{C9D4FF}{\textcolor{black}{+0.286}} & 0.322\\
\hspace{1em}LA & \cellcolor[HTML]{EA7088}{\textcolor{black}{-0.357}} & 0.216 & \cellcolor[HTML]{E14F72}{\textcolor{black}{-0.643}} & \underline{0.026} & \cellcolor[HTML]{E5E9FF}{\textcolor{black}{+0.143}} & 0.621\\
\hspace{1em}PM & \cellcolor[HTML]{EE8195}{\textcolor{black}{-0.317}} & 0.087 & \cellcolor[HTML]{F7ABB6}{\textcolor{black}{-0.319}} & 0.086 & \cellcolor[HTML]{DAE0FF}{\textcolor{black}{+0.202}} & 0.279\\
\hspace{1em}RE & \cellcolor[HTML]{FABEC6}{\textcolor{black}{-0.167}} & 0.368 & \cellcolor[HTML]{FCC6CD}{\textcolor{black}{-0.217}} & 0.242 & \cellcolor[HTML]{FCCED4}{\textcolor{black}{-0.109}} & 0.558\\
\bottomrule
\end{tabular}

    \endgroup
    \end{small}
    \label{tab:r2_kendall_tau}
\end{table}

\paragraph{Which models lead to the best method performances} In order to further analyze the relationship between model and \ac{aed} performances, we look which model leads to the best performance on a given dataset.
In \cref{fig:r2_alluvial} we show the results differentiated by overall, precision, and recall scores.
We observe that in the most cases, the best or second best models lead to the best method performances.
It is especially clear for token labeling, where using Flair leads to the best performance in all cases if we look at the overall and precision score.
Interestingly, Flair has better performance than transformers for span labeling but the latter is preferred by most methods.
Flair only leads to best method performances for most of \hids{CoNLL-2003} and parts of \hids{Flights}.
Besides the fact that better models  on average lead to better \ac{aed} performance, we do not see a consistent pattern that certain methods prefer certain models.

\noindent
A special case, however, is the recall of \himl{Retag}.
We indeed observe the assumption of \citet{reissIdentifyingIncorrectLabels2020} that the model with the lowest recall often leads to the highest \ac{aed} recall (see \cref{fig:r2_alluvial}).
This is especially pronounced for token and span labeling.
For these tasks, \himl{Retag} can use a low-recall model to flag a large fraction of tokens  because the model disagrees at many positions with the input labels.
This improves recall while being detrimental to precision.

To summarize, we overall see positive correlation between model and \ac{aed} performances.
Using a well-performing model is a good choice for most model-based \ac{aed} approaches.
Neural models perform especially well although they are more expensive to train.
We therefore use transformers for text classification as well as span labeling and Flair for token labeling.
Using a low-recall model for \himl{Retag} leads to higher recall for token and span labeling, as conjectured by \citet{reissIdentifyingIncorrectLabels2020}.
This however concurs with lower precision and excessive flagging and thus more annotations need to be inspected.

\subsection{RQ4 -- What performance impact does using (or not using) cross-validation have?}
\label{sec:r3_cross_validation}

\begin{table}[ht]
	\centering
	\caption{Performance delta of model-based methods when training models with and without \acf{cv}. Negative \negativecircle values indicate that not using \ac{cv} performs worse than using it, positive \positivecircle values the opposite. It can be seen that overall recall is strongly impacted when not using \ac{cv} but precision can improve. Flagger and scorer results are separated by a gap.}
	\begin{small}
		\begingroup
		\renewcommand*{\arraystretch}{1.2}
		\renewcommand*{\tabcolsep}{4pt}
		
\begin{tabular}{cccccccccc}
\toprule
\multicolumn{1}{c}{ } & \multicolumn{3}{c}{Text} & \multicolumn{2}{c}{Token} & \multicolumn{4}{c}{Span} \\
\cmidrule(l{3pt}r{3pt}){2-4} \cmidrule(l{3pt}r{3pt}){5-6} \cmidrule(l{3pt}r{3pt}){7-10}
Method & ATIS & IMDb & SST & GUM & Plank & Comp. & CoNLL & Flights & Forex\\
\midrule
\addlinespace[0.3em]
\multicolumn{10}{l}{\textbf{$\Delta$ Precision}}\\
\hspace{1em}CL & \cellcolor[HTML]{ABBFFF}{\textcolor{black}{+0.51}} & \cellcolor[HTML]{739FFF}{\textcolor{black}{+0.77}} & \cellcolor[HTML]{FDD3D8}{\textcolor{black}{-0.22}} & \cellcolor[HTML]{E6EAFF}{\textcolor{black}{+0.16}} & \cellcolor[HTML]{D4DCFF}{\textcolor{black}{+0.27}} & \cellcolor[HTML]{FCFDFF}{\textcolor{black}{+0.02}} & \cellcolor[HTML]{96B1FF}{\textcolor{black}{+0.61}} & \cellcolor[HTML]{F2F4FF}{\textcolor{black}{+0.08}} & \cellcolor[HTML]{D4DCFF}{\textcolor{black}{+0.27}}\\
\hspace{1em}DE & \cellcolor[HTML]{BCCAFF}{\textcolor{black}{+0.41}} & \cellcolor[HTML]{EDF0FF}{\textcolor{black}{+0.11}} & \cellcolor[HTML]{DDE3FF}{\textcolor{black}{+0.21}} & \cellcolor[HTML]{D5DDFF}{\textcolor{black}{+0.26}} & \cellcolor[HTML]{DAE1FF}{\textcolor{black}{+0.23}} & \cellcolor[HTML]{C8D3FF}{\textcolor{black}{+0.34}} & \cellcolor[HTML]{FCCAD0}{\textcolor{black}{-0.26}} & \cellcolor[HTML]{9BB5FF}{\textcolor{black}{+0.59}} & \cellcolor[HTML]{C4D0FF}{\textcolor{black}{+0.36}}\\
\hspace{1em}IRT & \cellcolor[HTML]{FCFCFF}{\textcolor{black}{+0.02}} & \cellcolor[HTML]{C2CFFF}{\textcolor{black}{+0.38}} & \cellcolor[HTML]{FAFBFF}{\textcolor{black}{+0.03}} & \cellcolor[HTML]{FEFEFF}{\textcolor{black}{+0.00}} & \cellcolor[HTML]{F9FAFF}{\textcolor{black}{+0.04}} & \cellcolor[HTML]{F4F6FF}{\textcolor{black}{+0.07}} & \cellcolor[HTML]{FCC9CF}{\textcolor{black}{-0.27}} & \cellcolor[HTML]{FBFBFF}{\textcolor{black}{+0.03}} & \cellcolor[HTML]{F8ADB7}{\textcolor{black}{-0.41}}\\
\hspace{1em}LA & \cellcolor[HTML]{D5DDFF}{\textcolor{black}{+0.26}} & \cellcolor[HTML]{F5F7FF}{\textcolor{black}{+0.06}} & \cellcolor[HTML]{D2DBFF}{\textcolor{black}{+0.28}} & \cellcolor[HTML]{FFFEFE}{\textcolor{black}{-0.00}} & \cellcolor[HTML]{DBE2FF}{\textcolor{black}{+0.23}} & \cellcolor[HTML]{FEFEFF}{\textcolor{black}{+0.01}} & \cellcolor[HTML]{FFF4F5}{\textcolor{black}{-0.05}} & \cellcolor[HTML]{D1DAFF}{\textcolor{black}{+0.29}} & \cellcolor[HTML]{CCD6FF}{\textcolor{black}{+0.32}}\\
\hspace{1em}PE & \cellcolor[HTML]{F8F9FF}{\textcolor{black}{+0.05}} & \cellcolor[HTML]{FFFFFF}{\textcolor{black}{+0.00}} & \cellcolor[HTML]{FEFEFF}{\textcolor{black}{+0.01}} & \cellcolor[HTML]{FEFEFF}{\textcolor{black}{+0.01}} & \cellcolor[HTML]{F9FAFF}{\textcolor{black}{+0.04}} & \cellcolor[HTML]{E8ECFF}{\textcolor{black}{+0.14}} & \cellcolor[HTML]{FDFEFF}{\textcolor{black}{+0.01}} & \cellcolor[HTML]{DDE3FF}{\textcolor{black}{+0.22}} & \cellcolor[HTML]{EDF0FF}{\textcolor{black}{+0.11}}\\
\hspace{1em}RE & \cellcolor[HTML]{D1DAFF}{\textcolor{black}{+0.29}} & \cellcolor[HTML]{719EFF}{\textcolor{black}{+0.78}} & \cellcolor[HTML]{E1E7FF}{\textcolor{black}{+0.19}} & \cellcolor[HTML]{E8ECFF}{\textcolor{black}{+0.15}} & \cellcolor[HTML]{E4E8FF}{\textcolor{black}{+0.17}} & \cellcolor[HTML]{FFF9F9}{\textcolor{black}{-0.03}} & \cellcolor[HTML]{FFEBED}{\textcolor{black}{-0.10}} & \cellcolor[HTML]{E1E7FF}{\textcolor{black}{+0.19}} & \cellcolor[HTML]{D0D9FF}{\textcolor{black}{+0.29}}\\
\addlinespace[1em]\hspace{1em}CU & \cellcolor[HTML]{FFF8F9}{\textcolor{black}{-0.03}} & \cellcolor[HTML]{FEE6E9}{\textcolor{black}{-0.12}} & \cellcolor[HTML]{FDD7DB}{\textcolor{black}{-0.20}} & \cellcolor[HTML]{FFFFFF}{\textcolor{black}{-0.00}} & \cellcolor[HTML]{FFF5F6}{\textcolor{black}{-0.05}} & \cellcolor[HTML]{FFF4F5}{\textcolor{black}{-0.05}} & \cellcolor[HTML]{FEE5E8}{\textcolor{black}{-0.13}} & \cellcolor[HTML]{FFEEF0}{\textcolor{black}{-0.08}} & \cellcolor[HTML]{FBC5CC}{\textcolor{black}{-0.29}}\\
\hspace{1em}DU & \cellcolor[HTML]{C8D3FF}{\textcolor{black}{+0.34}} & \cellcolor[HTML]{FFFDFD}{\textcolor{black}{-0.01}} & \cellcolor[HTML]{ECEFFF}{\textcolor{black}{+0.12}} & \cellcolor[HTML]{F8F9FF}{\textcolor{black}{+0.04}} & \cellcolor[HTML]{E9EDFF}{\textcolor{black}{+0.14}} & \cellcolor[HTML]{FFFFFF}{\textcolor{black}{+0.00}} & \cellcolor[HTML]{FFF8F9}{\textcolor{black}{-0.03}} & \cellcolor[HTML]{F5F6FF}{\textcolor{black}{+0.07}} & \cellcolor[HTML]{E9EDFF}{\textcolor{black}{+0.14}}\\
\hspace{1em}PM & \cellcolor[HTML]{C9D4FF}{\textcolor{black}{+0.34}} & \cellcolor[HTML]{FFF6F7}{\textcolor{black}{-0.05}} & \cellcolor[HTML]{F8F9FF}{\textcolor{black}{+0.05}} & \cellcolor[HTML]{FAFBFF}{\textcolor{black}{+0.03}} & \cellcolor[HTML]{E9EDFF}{\textcolor{black}{+0.14}} & \cellcolor[HTML]{FEDCE0}{\textcolor{black}{-0.17}} & \cellcolor[HTML]{FFFBFB}{\textcolor{black}{-0.02}} & \cellcolor[HTML]{F8F9FF}{\textcolor{black}{+0.05}} & \cellcolor[HTML]{F5F6FF}{\textcolor{black}{+0.07}}\\
\addlinespace[0.3em]
\multicolumn{10}{l}{\textbf{$\Delta$ Recall}}\\
\hspace{1em}CL & \cellcolor[HTML]{FFF7F7}{\textcolor{black}{-0.04}} & \cellcolor[HTML]{ED7F93}{\textcolor{black}{-0.63}} & \cellcolor[HTML]{E25375}{\textcolor{black}{-0.82}} & \cellcolor[HTML]{FFF0F2}{\textcolor{black}{-0.07}} & \cellcolor[HTML]{FCCDD3}{\textcolor{black}{-0.25}} & \cellcolor[HTML]{FEDBDF}{\textcolor{black}{-0.18}} & \cellcolor[HTML]{FEDCE0}{\textcolor{black}{-0.17}} & \cellcolor[HTML]{FCCED4}{\textcolor{black}{-0.24}} & \cellcolor[HTML]{F6A3AF}{\textcolor{black}{-0.46}}\\
\hspace{1em}DE & \cellcolor[HTML]{FBC5CC}{\textcolor{black}{-0.29}} & \cellcolor[HTML]{FCCCD2}{\textcolor{black}{-0.25}} & \cellcolor[HTML]{F7ACB7}{\textcolor{black}{-0.41}} & \cellcolor[HTML]{FFECEE}{\textcolor{black}{-0.10}} & \cellcolor[HTML]{FABBC3}{\textcolor{black}{-0.34}} & \cellcolor[HTML]{FBBFC7}{\textcolor{black}{-0.31}} & \cellcolor[HTML]{FBC2C9}{\textcolor{black}{-0.30}} & \cellcolor[HTML]{F8B0BA}{\textcolor{black}{-0.39}} & \cellcolor[HTML]{F9B5BE}{\textcolor{black}{-0.37}}\\
\hspace{1em}IRT & \cellcolor[HTML]{B7C7FF}{\textcolor{black}{+0.44}} & \cellcolor[HTML]{F6F7FF}{\textcolor{black}{+0.06}} & \cellcolor[HTML]{93B0FF}{\textcolor{black}{+0.63}} & \cellcolor[HTML]{F0F3FF}{\textcolor{black}{+0.09}} & \cellcolor[HTML]{C3CFFF}{\textcolor{black}{+0.37}} & \cellcolor[HTML]{D6DEFF}{\textcolor{black}{+0.26}} & \cellcolor[HTML]{FBC0C7}{\textcolor{black}{-0.31}} & \cellcolor[HTML]{77A1FF}{\textcolor{black}{+0.76}} & \cellcolor[HTML]{FFEAEC}{\textcolor{black}{-0.11}}\\
\hspace{1em}LA & \cellcolor[HTML]{FCC9CF}{\textcolor{black}{-0.27}} & \cellcolor[HTML]{ED7D91}{\textcolor{black}{-0.64}} & \cellcolor[HTML]{E15072}{\textcolor{black}{-0.83}} & \cellcolor[HTML]{FFFEFF}{\textcolor{black}{-0.00}} & \cellcolor[HTML]{FBC2C9}{\textcolor{black}{-0.30}} & \cellcolor[HTML]{FEDBDF}{\textcolor{black}{-0.18}} & \cellcolor[HTML]{FCC9D0}{\textcolor{black}{-0.27}} & \cellcolor[HTML]{FABAC3}{\textcolor{black}{-0.34}} & \cellcolor[HTML]{F49CAA}{\textcolor{black}{-0.49}}\\
\hspace{1em}PE & \cellcolor[HTML]{FFFFFF}{\textcolor{black}{+0.00}} & \cellcolor[HTML]{FFFFFF}{\textcolor{black}{+0.00}} & \cellcolor[HTML]{FFFEFE}{\textcolor{black}{-0.00}} & \cellcolor[HTML]{FFFFFF}{\textcolor{black}{-0.00}} & \cellcolor[HTML]{FEE2E5}{\textcolor{black}{-0.14}} & \cellcolor[HTML]{FEE5E7}{\textcolor{black}{-0.13}} & \cellcolor[HTML]{FFF3F4}{\textcolor{black}{-0.06}} & \cellcolor[HTML]{FFFFFF}{\textcolor{black}{+0.00}} & \cellcolor[HTML]{FFEAEC}{\textcolor{black}{-0.11}}\\
\hspace{1em}RE & \cellcolor[HTML]{FABEC6}{\textcolor{black}{-0.32}} & \cellcolor[HTML]{ED7C91}{\textcolor{black}{-0.64}} & \cellcolor[HTML]{E25375}{\textcolor{black}{-0.82}} & \cellcolor[HTML]{FFFEFE}{\textcolor{black}{-0.00}} & \cellcolor[HTML]{F9B7C0}{\textcolor{black}{-0.35}} & \cellcolor[HTML]{FEDCE0}{\textcolor{black}{-0.17}} & \cellcolor[HTML]{FBC2CA}{\textcolor{black}{-0.30}} & \cellcolor[HTML]{F6A6B2}{\textcolor{black}{-0.44}} & \cellcolor[HTML]{F18F9F}{\textcolor{black}{-0.55}}\\
\addlinespace[1em]\hspace{1em}CU & \cellcolor[HTML]{FFF1F2}{\textcolor{black}{-0.07}} & \cellcolor[HTML]{EE8195}{\textcolor{black}{-0.62}} & \cellcolor[HTML]{F8AEB8}{\textcolor{black}{-0.40}} & \cellcolor[HTML]{FFFFFF}{\textcolor{black}{-0.00}} & \cellcolor[HTML]{FFF7F7}{\textcolor{black}{-0.04}} & \cellcolor[HTML]{FFFCFD}{\textcolor{black}{-0.01}} & \cellcolor[HTML]{FCCED4}{\textcolor{black}{-0.24}} & \cellcolor[HTML]{FDD3D8}{\textcolor{black}{-0.22}} & \cellcolor[HTML]{FBC6CC}{\textcolor{black}{-0.28}}\\
\hspace{1em}DU & \cellcolor[HTML]{82A6FF}{\textcolor{black}{+0.71}} & \cellcolor[HTML]{FFF4F6}{\textcolor{black}{-0.05}} & \cellcolor[HTML]{D8DFFF}{\textcolor{black}{+0.25}} & \cellcolor[HTML]{F1F3FF}{\textcolor{black}{+0.09}} & \cellcolor[HTML]{EDF0FF}{\textcolor{black}{+0.12}} & \cellcolor[HTML]{FFFFFF}{\textcolor{black}{+0.00}} & \cellcolor[HTML]{FFF2F3}{\textcolor{black}{-0.06}} & \cellcolor[HTML]{E4E9FF}{\textcolor{black}{+0.17}} & \cellcolor[HTML]{E9EDFF}{\textcolor{black}{+0.14}}\\
\hspace{1em}PM & \cellcolor[HTML]{83A7FF}{\textcolor{black}{+0.71}} & \cellcolor[HTML]{FDD1D6}{\textcolor{black}{-0.23}} & \cellcolor[HTML]{F1F3FF}{\textcolor{black}{+0.09}} & \cellcolor[HTML]{F5F7FF}{\textcolor{black}{+0.06}} & \cellcolor[HTML]{EDF0FF}{\textcolor{black}{+0.12}} & \cellcolor[HTML]{FFF7F8}{\textcolor{black}{-0.04}} & \cellcolor[HTML]{FFF7F8}{\textcolor{black}{-0.04}} & \cellcolor[HTML]{ECEFFF}{\textcolor{black}{+0.12}} & \cellcolor[HTML]{F5F6FF}{\textcolor{black}{+0.07}}\\
\addlinespace[0.3em]
\multicolumn{10}{l}{\textbf{$\Delta$ \% Flagged}}\\
\hspace{1em}CL & \cellcolor[HTML]{FFFBFB}{\textcolor{black}{-0.02}} & \cellcolor[HTML]{FFF4F5}{\textcolor{black}{-0.06}} & \cellcolor[HTML]{FED9DD}{\textcolor{black}{-0.19}} & \cellcolor[HTML]{FFFCFC}{\textcolor{black}{-0.02}} & \cellcolor[HTML]{FFF1F3}{\textcolor{black}{-0.07}} & \cellcolor[HTML]{FFECEE}{\textcolor{black}{-0.09}} & \cellcolor[HTML]{FFFAFB}{\textcolor{black}{-0.02}} & \cellcolor[HTML]{FFFDFD}{\textcolor{black}{-0.01}} & \cellcolor[HTML]{FFF2F4}{\textcolor{black}{-0.06}}\\
\hspace{1em}DE & \cellcolor[HTML]{FFF5F6}{\textcolor{black}{-0.05}} & \cellcolor[HTML]{FFF6F7}{\textcolor{black}{-0.05}} & \cellcolor[HTML]{FEE1E5}{\textcolor{black}{-0.15}} & \cellcolor[HTML]{FFF9F9}{\textcolor{black}{-0.03}} & \cellcolor[HTML]{FFEAEC}{\textcolor{black}{-0.10}} & \cellcolor[HTML]{FBC1C9}{\textcolor{black}{-0.31}} & \cellcolor[HTML]{FFF3F4}{\textcolor{black}{-0.06}} & \cellcolor[HTML]{FFF3F4}{\textcolor{black}{-0.06}} & \cellcolor[HTML]{FFEAEC}{\textcolor{black}{-0.11}}\\
\hspace{1em}IRT & \cellcolor[HTML]{F5F7FF}{\textcolor{black}{+0.06}} & \cellcolor[HTML]{DB3A67}{\textcolor{black}{-0.91}} & \cellcolor[HTML]{E5EAFF}{\textcolor{black}{+0.16}} & \cellcolor[HTML]{FAFBFF}{\textcolor{black}{+0.03}} & \cellcolor[HTML]{EBEEFF}{\textcolor{black}{+0.13}} & \cellcolor[HTML]{E8ECFF}{\textcolor{black}{+0.15}} & \cellcolor[HTML]{FFF2F4}{\textcolor{black}{-0.06}} & \cellcolor[HTML]{F4F6FF}{\textcolor{black}{+0.07}} & \cellcolor[HTML]{6E9CFF}{\textcolor{black}{+0.79}}\\
\hspace{1em}LA & \cellcolor[HTML]{FFF9F9}{\textcolor{black}{-0.03}} & \cellcolor[HTML]{FFF3F5}{\textcolor{black}{-0.06}} & \cellcolor[HTML]{FED9DD}{\textcolor{black}{-0.19}} & \cellcolor[HTML]{FFFFFF}{\textcolor{black}{+0.00}} & \cellcolor[HTML]{FFEBED}{\textcolor{black}{-0.10}} & \cellcolor[HTML]{FFEBED}{\textcolor{black}{-0.10}} & \cellcolor[HTML]{FFF5F6}{\textcolor{black}{-0.05}} & \cellcolor[HTML]{FFF9FA}{\textcolor{black}{-0.03}} & \cellcolor[HTML]{FFEDEF}{\textcolor{black}{-0.09}}\\
\hspace{1em}PE & \cellcolor[HTML]{FFFCFC}{\textcolor{black}{-0.01}} & \cellcolor[HTML]{FFFEFE}{\textcolor{black}{-0.00}} & \cellcolor[HTML]{FFFDFE}{\textcolor{black}{-0.01}} & \cellcolor[HTML]{FFFFFF}{\textcolor{black}{-0.00}} & \cellcolor[HTML]{FFF6F7}{\textcolor{black}{-0.04}} & \cellcolor[HTML]{FEE1E4}{\textcolor{black}{-0.15}} & \cellcolor[HTML]{FFF9FA}{\textcolor{black}{-0.03}} & \cellcolor[HTML]{FFF0F1}{\textcolor{black}{-0.08}} & \cellcolor[HTML]{FFF5F6}{\textcolor{black}{-0.05}}\\
\hspace{1em}RE & \cellcolor[HTML]{FFF7F8}{\textcolor{black}{-0.04}} & \cellcolor[HTML]{FFF3F4}{\textcolor{black}{-0.06}} & \cellcolor[HTML]{FDD8DD}{\textcolor{black}{-0.19}} & \cellcolor[HTML]{FFFBFB}{\textcolor{black}{-0.02}} & \cellcolor[HTML]{FFEBED}{\textcolor{black}{-0.10}} & \cellcolor[HTML]{FFEEF0}{\textcolor{black}{-0.08}} & \cellcolor[HTML]{FFF5F6}{\textcolor{black}{-0.05}} & \cellcolor[HTML]{FFF9F9}{\textcolor{black}{-0.03}} & \cellcolor[HTML]{FFECEE}{\textcolor{black}{-0.09}}\\
\bottomrule
\end{tabular}

		\endgroup    \end{small}

	\label{tab:cv_needed_wide}
\end{table}

Model-based \ac{aed} approaches are typically used together with \ac{cv} \citep[e.g.,][]{amiriSpottingSpuriousData2018, larsonInconsistenciesCrowdsourcedSlotFilling2020, reissIdentifyingIncorrectLabels2020}.
\citet{northcuttConfidentLearningEstimating2021} explicitly state that \himl{Confident Learning} should only be applied to  out-of-sample predicted probabilities.
\citet{amiriSpottingSpuriousData2018} do not mention that they used \ac{cv} for \himl{Dropout Uncertainty}, \himl{Label Aggregation}, or \himl{Classification Uncertainty}.

When using \ac{aed} with \ac{cv}, models are trained on $k-1$ splits and then detection is done on the remaining $k$-th set.
After all unique folds are processed all instances have been checked.
\ac{cv} is often used in supervised learning where the goal is to find a model configuration as well as hyper-parameters that generalize on unseen data.
The goal of \ac{aed}, however, is to find errors in the data at hand.
Resulting models are just an instrument and not used afterwards.
They therefore will not be applied to unseen data and need not to generalize to data other than the one to clean.
Hence, the question arises whether \ac{cv} is really necessary for \ac{aed}, which has not been analyzed as of yet.
Not using \ac{cv} has the advantage of being much faster and using less energy, since using \ac{cv} increases training time linearly with the number of folds.
In the typical setup with 10-fold \ac{cv}, this means an increase of training time by 10$\times$.

To answer this question we train a single model on all instances and then predict on the very same data.
Then we use the resulting outputs to rerun methods that used \ac{cv} before, which are \himl{Classification Uncertainty}, \himl{Confident Learning}, \himl{Diverse Ensemble}, \himl{Dropout Uncertainty}, \himl{Item Response Theory}, \himl{Label Aggregation}, \himl{Prediction Margin}, \himl{Projection Ensemble}, and \himl{Retag}.
The results are listed in \cref{tab:cv_needed_wide}.
Overall, it can be seen that not using \ac{cv} massively degrades recall for model-based methods while the precision improves.
This can be intuitively explained by the fact that if the underlying models have already seen all the data, then they overfit to it and hence can re-predict it well.
Due to the positive relationship between model and method performances (see \cref{sec:r2_relationship}) this is also reflected downstream; less instances are predicted differently than the original labels.
This reduces recall and thereby the chance of making errors, thus increasing precision.
This can be seen by the reduction in the percentage of flagged instances for flaggers.
Interestingly, \himl{Dropout Uncertainty} and \himl{Prediction Margin} are not impacted as much and sometimes even improve when not using \ac{cv} across all scores, especially for easier datasets.
Recall of \himl{Item Response Theory} also improves at the cost of more flagged items and a reduction in precision.
\himl{Prediction ensemble} for text classification is relatively unaffected and for token and span labeling, the performance difference is around $\pm 0.10$ pp.
Therefore, it might be a good tradeoff to not use \ac{cv} with this method as it is already expensive due to its ensembling.

To summarize, not using cross-validation (CV) can negatively impact performance -- in particular, degrading recall. We therefore recommend the use of CV, even though it increases runtime by the number of folds (in our case, by a factor of ten).
In settings where this is an issue we recommend using methods that inherently do not need \ac{cv}.
These include most heuristics and well-performing approaches like \himl{Datamap Confidence}, \himl{Leitner Spotter}, or \himl{Curriculum Spotter}.
If precision is more important than recall then not using \ac{cv} might be taken into consideration.

    \section{Takeaways and recommendations}
\label{sec:takeaways}

This article has probed several questions related to \acl{aed}.
Our findings show that it is usually better to use well-performing models for model-based methods as they yield better detection performance on average.
Using a worse model for \himl{Retag} improves recall at the cost of lower precision.
For detection these models should be trained via cross-validation, otherwise the recall of downstream methods is heavily degraded (while the precision improves).
Calibration can improve these model-based annotation error detection methods, but more research is needed to determine when exactly it can be useful.
Some model-method combinations achieved relatively large gains after calibration while others did not improve.

Methods that are used frequently in practice --- \himl{Retag} and \himl{Classification Uncertainty} --- performed well in our experiments.
Others did not perform particularly well, especially \himl{Dropout Uncertainty}, \himl{Item Response Theory}, \himl{k-Nearest Neighbor Entropy}, \himl{Mean Distance}, and \himl{Prediction Margin}.
For \himl{Mean Distance} in particular, \citet{larsonOutlierDetectionImproved2019} reported AP   of  $>0.6$  and recall $>0.8$ on corpora with artificial noise, which we could not reproduce.
Experiments with \himl{Dropout Uncertainty} disseminated in \citet{amiriSpottingSpuriousData2018} reached similar high scores as using \himl{Curriculum Spotter}, \himl{Leitner Spotter}, or \himl{Classification Uncertainty}, but we were not able to make \himl{Dropout Uncertainty} reach similar high scores as the others.
\himl{Label Aggregation}, though, which uses the same inputs performs exceedingly well.
For the others, either no scores were reported or they were similarly low as in our experiments.

Experiments on actual corpora have shown that \ac{aed} methods still have room for improvement.
While looking promising on artificial corpora, there is a large performance drop when applying them in practice.
Overall, the methods that worked best are \himl{Classification Uncertainty}, \himl{Confident Learning}, \himl{Curriculum Spotter}, \himl{Datamap Confidence}, \himl{Diverse Ensemble}, \himl{Label Aggregation}, \himl{Leitner Spotter}, \himl{Projection Ensemble}, and  \himl{Retag}.
More complicated methods are not necessarily better.
For instance \himl{Classification Uncertainty} and \himl{Retag} perform well across tasks and datasets while being easy to implement.
Model-based methods require $k$-fold cross-validation.
Therefore, if runtime is a concern, then \himl{Datamap Confidence} is a good alternative.
It performs well while only needing to train one model instead of $k$.
In case the data or its corresponding task to correct is not suitable for machine learning, methods like \himl{Label Entropy}, \himl{K-Nearest-Neighbor Entropy} or \himl{Variation n-grams} still can be applied.
As the latter usually has high precision it is often worth to apply it whenever the data is suitable for it; that is, if the data has sufficient surface form overlap.
Individual scorer scores can be aggregated via \himl{Borda Count} but it tremendously increases runtime.
While not yielding significantly better results in our experiments, results aggregated that way were much more stable across datasets and tasks while individual scorers sometimes had performance drops in certain settings.

Manual analysis of \hids{CoNLL-2003} showed that finding inconsistencies is often more difficult than finding annotation errors. While model-based methods were often quite good in the latter, they performed poorly when detecting inconsistencies.
Methods that do not rely on the noisy labels but on the surface form or semantics like \himl{k-Nearest Neighbor Entropy}, \himl{Label Entropy} and \himl{Weighted Discrepancy} have shown the opposite behaviour.
They each have their own strengths and it can be worth to combine both types of methods.

    \section{Conclusion}

Having annotated corpora with high-quality labels is imperative for many branches of science and for the training of well-performing and generalizing models.
Previous work has shown that even commonly used benchmark corpora contain non-negligible numbers of annotation errors.
In order to assist human annotators with detecting and correcting these errors, many different methods for annotation error detection have been developed.
To date, however, methods have not been compared, so it has been unclear what method to choose under what circumstances.
To close this gap, we surveyed the field of annotation error detection, reimplemented 18 methods, collected and generated 9 datasets for text classification, token labeling, and span labeling and evaluated method performance in different settings.
Our results show that \ac{aed} can already be useful in real use cases to support data cleaning efforts.
But especially for more difficult datasets, the performance ceiling is far from reached yet.

In the past, the focus of most works researching or using \ac{aed} was to clean data and not to develop a method.
The method was only a means to achieve a cleaned corpus and not the target in itself.
Also, several works proposed algorithms for different use cases and \ac{aed} was one application to it just mentioned briefly at the end without in-depth evaluation, rendering it unclear how well the method performs.
We therefore strongly encourage authors that introduce new \ac{aed} methods to compare themselves to previous work and on the same corpora to foster reproducibility and to bring the performance of new methods into context.
This article surveys, standardizes and answers several fundamental questions regarding \ac{aed} so that future work has a stable footing for their research.
For this, we also make our implementation and datasets publicly available.

\paragraph{Limitations and future work}

While we thoroughly investigated many available methods on different datasets and tasks, there are some limitations to our work.
First, the datasets that we used were only in English.
Therefore, it would be interesting to investigate \ac{aed} on different languages.
One first step could be the work by \citet{hedderichAnalysingNoiseModel2021} who created a corpus  for named entity recognition in Estonian with natural noise patterns.
Hand-curated datasets with explicitly annotated errors are rare.
We therefore also used existing, clean datasets and injected random noise, similarly to previous works.
These datasets with artificial errors have been shown to overestimate the ability of \ac{aed} methods, but are still a good estimator for the maximal performance of methods.
The next step is to create benchmark corpora that are designed from the ground up for the evaluation of annotation error detection.
As creating these requires effort and is costly, a cheaper way is to aggregate raw crowdsourcing data.
This is often not published along with adjudicated corpora so we urge researchers to also publish these alongside the final corpus.

\ac{aed} was evaluated on three different tasks with nine \ac{nlp} datasets.
The tasks were chosen, based on the number of datasets and model types available to answer our research questions.
Most \ac{aed} methods are task agnostic; previous work for instance investigated question answering~\citep{amiriSpottingSpuriousData2018} or relation classification~\citep{altTACREDRevisitedThorough2020, stoicaReTACREDAddressingShortcomings2021}.
Hence, \ac{aed} can and has been applied in different fields like computer vision~\citep{northcuttPervasiveLabelErrors2021}.
But these works are plagued by the same issues that most previous \ac{aed} works have (e.g., only limited comparison to other works and  quantitive analysis, code and data not available).
Having several fundamental questions answered in this article, future work can now readily apply and investigate \ac{aed} on many different tasks, domains and in many different settings while leveraging our findings (which are summarized in \cref{sec:takeaways}).
It would especially be interesting to evaluate and apply \ac{aed} on more hierarchical and difficult tasks, such as semantic role labeling or natural language inference.

While we investigated many relevant research questions, these were mostly about model-based methods as well as flaggers.
To date, scorers have been treated as a black box, so it would be worth investigating what makes a good scorer -- for example, what makes \himl{Classification Uncertainty} better than \himl{Prediction Margin}.
Also, leveraging scorers as uncertainty estimates for correction is a promising application, similar to the works of \citet{dligachReducingNeedDouble2011} or \citet{angleAutomatedErrorCorrection2018}.

This work also only focuses on errors, inconsistencies and ambiguities related to instance labels.
Some datasets also benefit from finding errors concerning tokenization, sentence splitting,  or missing entities~\citep[e.g.,][]{reissIdentifyingIncorrectLabels2020}.
We also did not investigate the specific kinds of errors made.
This can be useful information and could be leveraged by human annotators during manual corrections.
It would be especially interesting to investigate the kinds of errors certain models and configurations were able to correct.
For instance whether using no cross-validation finds more obvious errors but with a higher precision.
We leave detection of errors other than incorrect labels or error kind detection for future work because we did not find a generic way to do it across the wide range of evaluated datasets and tasks used in this article.

Finally, we implemented each method as described and performed only basic hyper-parameter tuning.
We did not tune them further due to the prohibitive costs for our large-scale setup.
This is especially true for the considered machine learning models, where we kept the parameters mostly default for all regardless of the dataset and domain.
We are sure that one can certainly improve scores for each method, but our implementations should still serve as a lower bound.
However, we do not expect large gains from further optimization and no large shifts in ranking between the methods.
    \appendix

\appendixsection{Hyper-parameter tuning and implementation details}

In this section we briefly describe implementation details for the different \ac{aed} methods used throughout this article.
As the tuning data we select one corpus for each task type.
For text classification we subsample 5000 instances from the training split of \hids{AG News} \citep{zhangCharacterlevelConvolutionalNetworks2015}; the number of samples is chosen as it is around the same data size as our other datasets.
As the corpus for token labeling we choose the English part of \hids{ParTUT}~\citep{sanguinettiPartTUTTurinUniversity2015} and their POS annotations and inject 5\% random noise.
For span labeling we use \hids{CoNLL-2003} of \cite{reissIdentifyingIncorrectLabels2020} to which we apply 5\% flipped label noise.

\subsection{Aggregating probabilities and embeddings for span labeling}
\label{sec:app_aggregation}

When converting BIO-tagged sequences to spans for alignment (see \cref{sec:span_alignment}) consisting only of start and end position as well as its label, the probabilities assigned to each BIO-tag representing the span need to be aggregated.
The same needs to be done for creating span embeddings from token embeddings.
As an example, consider named entity recognition for persons and locations with a tagset of \texttt{B-PER}, \texttt{I-PER}, \texttt{B-LOC}, \texttt{I-LOC}.
It has to be aggregated so that spans have labels \texttt{PER} or \texttt{LOC}.
Look at a span of two tokens that has been tagged as \texttt{B-PER, I-PER}.
Then the probability for \texttt{PER} needs to be aggregated from the \texttt{B-PER} and \texttt{I-PER} tags.
We evaluate on our \hids{CoNLL-2003} tuning data.
We use a Maxent sequence tagger to evaluate \himl{Confident Learning} with 10-fold cross-validation for this hyper-parameter selection.
In addition, for \himl{k-Nearest-Neighbor Entropy} we evaluate aggregation schemes to create span embeddings from individual token embeddings.
Overall, we do not observe a large difference between max, mean or median aggregation.
The results can be seen in \cref{tab:appendix_span_alignment_aggregation}.
We choose aggregating via arithmetic mean because it is slightly better in terms of F1 and AP than the other methods.

\begin{table}[thb]
    \centering
    \caption{Impact of different aggregation functions for span alignment and embeddings.}
    \begin{tabular}{@{}lSSSSSS@{}}
        \toprule
        &     \multicolumn{3}{c}{CL}     &          \multicolumn{3}{c}{KNN}          \\
        \cmidrule(lr){2-4}\cmidrule(lr){5-7}
        {Aggregation}                       & {P}      & {R}      & {F1}     & {AP}     & {P@10\%} & {R@10\%} \\ \midrule
        min                                 & 0.148829 & 0.593627 & 0.237991 & 0.307023 & 0.324935 & 0.326495 \\
        max                                 & 0.761025 & 0.881275 & 0.816748 & 0.307649 & 0.3245   & 0.326058 \\
        \rowcolor{jck-gray}mean             & 0.765804 & 0.877783 & 0.817978 & 0.318326 & 0.331451 & 0.333042 \\
        median                              & 0.765065 & 0.8756   & 0.816609 & 0.316105 & 0.330148 & 0.331733 \\ \bottomrule
    \end{tabular}
    \label{tab:appendix_span_alignment_aggregation}
\end{table}

\subsection{Method details}

In the following we describe the choices we made when implementing the various \ac{aed} methods evaluated in this article.

\begin{description}[itemsep=5pt,wide, labelwidth=!, labelindent=0pt]
\item[Diverse Ensemble]
Our diverse ensemble uses the predictions of all different model types trained for the task and dataset, similarly to~\citet{loftssonCorrectingPOSTaggedCorpus2009, altTACREDRevisitedThorough2020, barnesSentimentAnalysisNot2019}.

\item[Spotter and Datamap Confidence]
The implementations of \himl{Datamap Confidence} as well as \himl{Curriculum Spotter} and \himl{Leitner Spotter} require callbacks or a similar functionality to obtain predictions for every epoch which only Hugging Face Transformers provide. That is why we only evaluate these methods in combination with a transformer.

\item[Dropout Uncertainty]
In our implementation we use mean entropy which we observed in preliminary experiments to perform slightly better overall than the other version evaluated by \citet{shelmanovHowCertainYour2021}.

\item[Variation n-grams]
We follow \citet{wisniewskiErratorToolHelp2018} for our implementation and use generalized suffix trees to find repetitions.
If there are repetitions of length more than one in the surface forms that are tagged differently we look up the respective tag sequence that occurs most often in the corpus and flag the positions of all other repetitions where they disagree with the majority tags.
We convert tokens and sentences to lower case to slightly increase recall while slightly reducing precision.
We do not flag an instance if its label is the most common label.
This yields far better results as the most common label is most often correct and should not be flagged.
When using \himl{Variation n-grams} for span labeling we use a context of one token to the left and right of the span, similarly to \citet{larsonInconsistenciesCrowdsourcedSlotFilling2020}.

\item[Projection ensemble]
In our implementation we flag an instance if the majority label of the ensemble disagrees with the given one.

\item[Label aggregation]
In the original work that evaluated using \himl{Label Aggregation} for \ac{aed} \citep{amiriSpottingSpuriousData2018},  MACE~\citep{hovyLearningWhomTrust2013} was used.
We use Dawid-Skene~\citep{dawidMaximumLikelihoodEstimation1979}, which has similar performance as MACE~\citep{paunComparingBayesianModels2018} but many more available implementations (we use \citet{ustalovGeneralpurposeCrowdsourcingComputational2021}).
The difference between the two --- MACE modeling annotator spam --- is not relevant here.

\item[Mean distance]
\label{sec:app_mean_distance}
We compare different embedding methods and metrics for \himl{Mean Distance}.
For that we use the Sentence Transformers\footnote{\url{https://www.sbert.net/}} library  and evaluate S-BERT embeddings~\citep{reimersSentenceBERTSentenceEmbeddings2019a}, Universal Sentence Encoder~\citep{cerUniversalSentenceEncoder2018} and average GloVe embeddings~\citep{penningtonGloVeGlobalVectors2014}.
We evaluate on our \hids{AG News} tuning data.
As our Universal Sentence Encoder implementation we use \texttt{distiluse-base-multilingual-cased-v1} from Sentence Transformers.
The Universal Sentence Encoder embeddings as used in the original implementation of \himl{Mean Distance}  \citep{larsonOutlierDetectionImproved2019} overall perform not better than all S-BERT embeddings.
\texttt{lof} refers to Local Outlier Factor, a clustering metric proposed by \citet{breunigLOFIdentifyingDensitybased2000}.
Using \emph{all-mpnet-base-v2} together with euclidean distance works best here and we will use this throughout our experiments.

\item[Item Response Theory]
We use the setup from \citet{rodriguezEvaluationExamplesAre2021}, that is a 2P IRT model that is optimized via variational inference and the original code of the authors.
We optimize for $10,000$ iterations.
\himl{Item Response Theory} uses the collected predictions of all models for the respective task, similarly to \himl{Diverse Ensemble}.

\item[Label Entropy and Weighted Discrepancy]
We implement equations $(2)$ and $(3)$ in \citet{hollensteinInconsistencyDetectionSemantic2016} but assign the minimum score (meaning no error) if the current label is the most common label.
This yields far better results because the most common label is most often correct and should not be downranked.

\item[K-Nearest-Neighbor Entropy]
\label{sec:app_details_knn}
To evaluate which embedding aggregation over transformer layers works best for \himl{k-Nearest-Neighbor Entropy}, we evaluate several different configurations on our \hids{ParTUT} tuning data.
We chose this task and not span labeling as span labeling requires an additional aggregation step to combine token embeddings to span embeddings (see \cref{sec:app_aggregation}).
The transformer of choice is RoBERTa \citep{liuRoBERTaRobustlyOptimized2019}, as it has better performance than BERT while still being fast enough.
We also compare with several non-transformer embeddings: Glove 6B~\citep{penningtonGloVeGlobalVectors2014} , Byte-Pair Encoding~\citep{heinzerlingBPEmbTokenizationfreePretrained2018} and a concatenation of both.
We follow \citet{devlinBERTPretrainingDeep2019a} regarding which configurations to try.
The results can be seen in \cref{tab:app_knn_partut}.
The best scoring embedder is RoBERTa using only the last layer that will be the configuration used throughout this work for obtaining token and span embeddings.
For sentence embeddings we will use \emph{all-mpnet-base-v2} (see \cref{sec:app_mean_distance}).
To compute the KNN entropy, we use the code of the original authors. %

\begin{table}[tbh]
	\centering
	\caption{Evaluation the performance impact of using different embedding types and configurations for KNN entropy on UD ParTUT.}
	\small{
		\begin{tabular}{@{}lSSS@{}}
			\toprule
			{Embedder}                      & {AP}     & {P@10\%} & {R@10\%} \\ \midrule
			\rowcolor{jck-gray} Last Hidden & 0.265451 & 0.255168 & 0.255696 \\
			Sum All Layers                  & 0.236867 & 0.25379  & 0.254315 \\
			First Hidden                    & 0.230412 & 0.269178 & 0.269735 \\
			Sum Last 4 Hidden               & 0.227318 & 0.245751 & 0.24626  \\
			Second-to-Last Hidden           & 0.224009 & 0.243684 & 0.244189 \\
			Concat Last 4 Hidden            & 0.148866 & 0.193156 & 0.193556 \\
			Glove                           & 0.148349 & 0.16881  & 0.16916  \\
			Glove + BPE                     & 0.148245 & 0.175011 & 0.175374 \\
			BPE                             & 0.146647 & 0.169729 & 0.170081 \\ \bottomrule
	\end{tabular}}
	\label{tab:app_knn_partut}
\end{table}

\item[Borda Count]
\label{sec:app_borda}
In order to evaluate which scores to aggregate via \himl{Borda Count} we evaluate three settings on our \hids{AG News} tuning data.
We either aggregate the five best, three best, or all scorer outputs.
As the underlying model for model-based methods we use transformers, because we need repeated probabilities for \himl{Dropout Uncertainty}.
The results are listed in \cref{tab:appendix_borda}.
It can be seen that aggregating only the three best scores leads to far superior performance. Hence, we choose this setting when evaluating \himl{Borda Count} aggregation during our experiments.
\end{description}

\begin{table}[h!t]
    \centering
    \caption{Evaluation of scorers and their aggregation via Borda count for text classification and span labeling. Highlighted in gray are the runs of Borda count aggregation.}
    \begin{subfigure}[t]{0.48\textwidth}
        \adjustbox{valign=t, max width=0.8\textwidth}{
        \begin{tabular}{@{}lSSS@{}}
        \toprule
        {Method}                                  & {AP}      & {P@10\%} & {R@10\%}  \\ \midrule
        \rowcolor{jck-gray}BC\textsubscript{top3} & 0.847598  & 0.46     & 0.946502  \\
        \rowcolor{jck-gray}BC\textsubscript{top2} & 0.823535  & 0.456    & 0.938272  \\
        DM                                        & 0.819075  & 0.448    & 0.921811  \\
        \rowcolor{jck-gray}BC\textsubscript{top5} & 0.79447   & 0.454    & 0.934156  \\
        LS                                        & 0.705893  & 0.448    & 0.921811  \\
        CU                                        & 0.521347  & 0.426    & 0.876543  \\
        MD                                        & 0.421569  & 0.344    & 0.707819  \\
        CS                                        & 0.295746  & 0.39     & 0.802469  \\
        \rowcolor{jck-gray}BC\textsubscript{all}  & 0.268355  & 0.244    & 0.502058  \\
        KNN                                       & 0.0547579 & 0.062    & 0.127572  \\
        DU                                        & 0.0546097 & 0.062    & 0.127572  \\
        PM                                        & 0.0503699 & 0.046    & 0.0946502 \\ \bottomrule
        \end{tabular}}
        \caption{Text}
    \end{subfigure}
    \hfill
    \begin{subfigure}[t]{0.48\textwidth}
        \adjustbox{valign=t, max width=0.83\textwidth}{
        \begin{tabular}{@{}lSSS}
        \toprule
        {Method}                                  & {AP}  & {P@10\%} & {R@10\%} \\ \midrule
        DM                                        & 0.963 & 0.932    & 0.934    \\
        \rowcolor{jck-gray}BC\textsubscript{top3} & 0.897 & 0.863    & 0.865    \\
        CU                                        & 0.881 & 0.837    & 0.839    \\
        \rowcolor{jck-gray}BC\textsubscript{top2} & 0.881 & 0.837    & 0.839    \\
        \rowcolor{jck-gray}BC\textsubscript{top5} & 0.716 & 0.625    & 0.626    \\
        WD                                        & 0.665 & 0.632    & 0.633    \\
        LE                                        & 0.567 & 0.579    & 0.580    \\
        \rowcolor{jck-gray}BC\textsubscript{all}  & 0.350 & 0.378    & 0.379    \\
        MD                                        & 0.206 & 0.231    & 0.232    \\
        DU                                        & 0.103 & 0.104    & 0.104    \\
        PM                                        & 0.102 & 0.104    & 0.104    \\
        KNN                                       & 0.100 & 0.101    & 0.101    \\ \bottomrule
        \end{tabular}}
        \caption{Span}
    \end{subfigure}
    \label{tab:appendix_borda}
\end{table}

\FloatBarrier

\appendixsection{Calibration}
\label{sec:appendix_calibration}

\begin{figure}[ht]
	\begin{minipage}{0.48\textwidth}
		\centering
		\includegraphics*[width=0.99\textwidth]{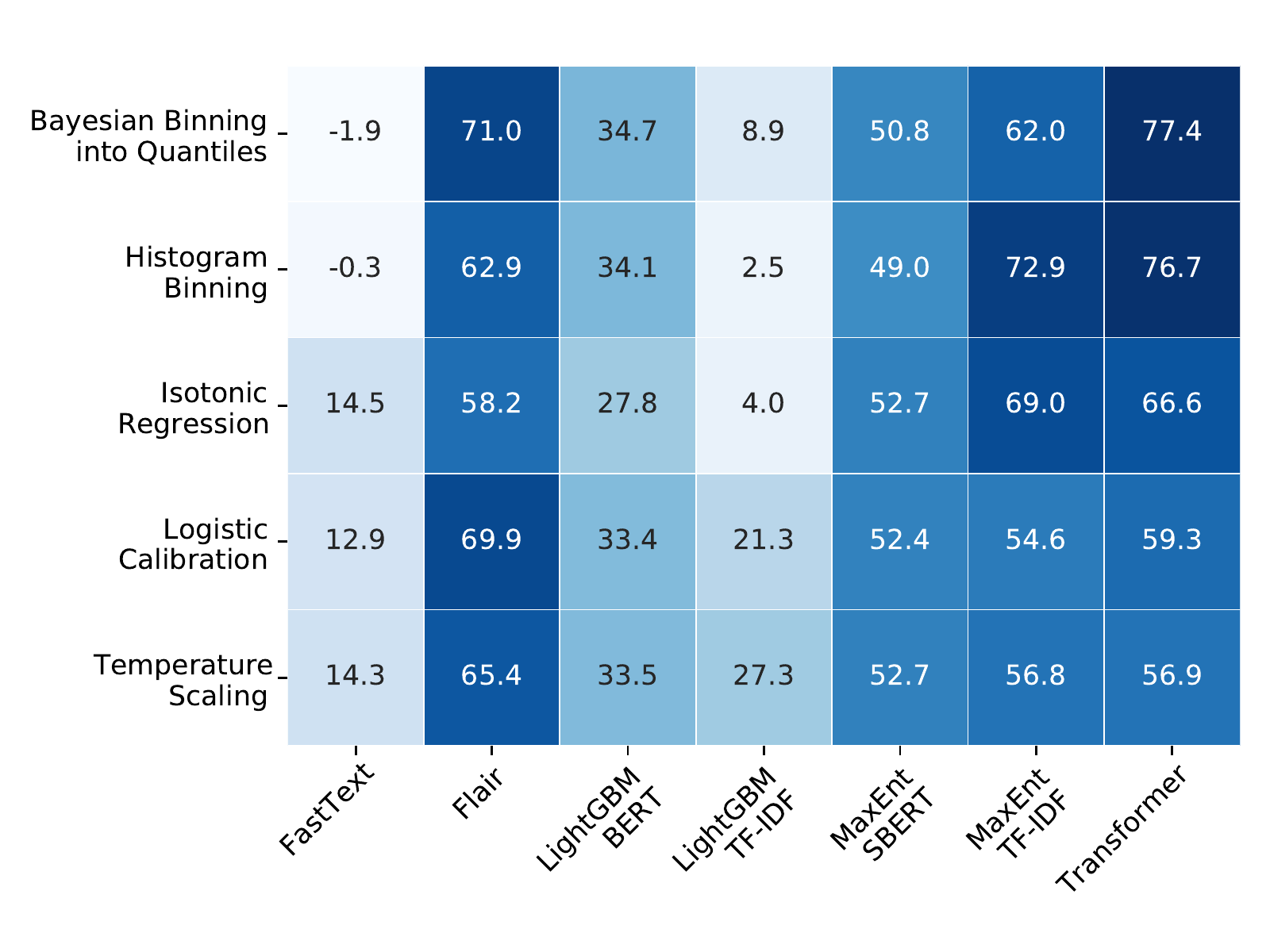}
		\caption{Percentage decrease of the Expected Calibration Error (ECE) after calibration, when training models on our \hids{AG news} tuning data. Higher is better.}
		\label{tab:appendix_calibration}
	\end{minipage}
	\hfil
	\begin{minipage}{0.48\textwidth}
		\centering
		\includegraphics*[width=0.99\textwidth]{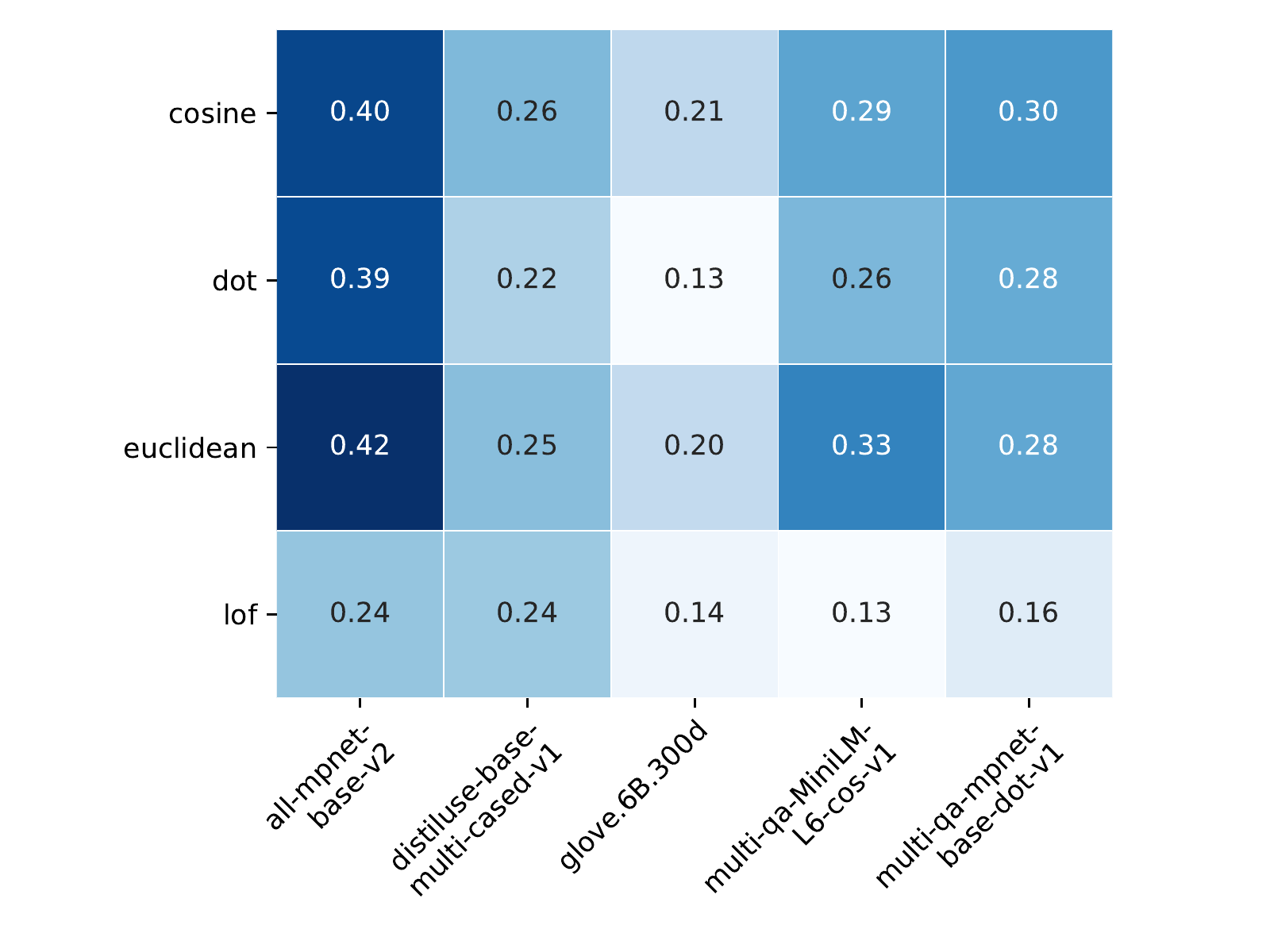}
		\label{tab:appendix_mean_distance_tuning}
		\caption{Average Precision of using \himl{Mean Distance} with different embedders and similarity metrics on our \hids{AG News} tuning data.}
	\end{minipage}
\end{figure}

From the most common calibration methods we select the best method by calibrating probabilities for models trained on our \hids{AG News} tuning data.
We use 10-fold cross-validation where eight parts are used for training models, one for calibrating and one for evaluating the calibration.
The results can be seen in \cref{tab:appendix_calibration}.
We follow \citet{guoCalibrationModernNeural2017} and use the Expected Calibration Error (ECE) \citep{naeiniObtainingWellCalibrated2015} as the metric for calibration quality.
We decide to use one method for all task types and finally choose \jhi{Logistic Calibration} (also known as Platt Scaling), which performs well across tasks.
We use the implementations of \citet{kuppersMultivariateConfidenceCalibration2020}.

\clearpage
\appendixsection{Best scores}
\label[appendix]{appendix:app:best}

\begin{figure}[h]
	\centering
	\subfloat[Text classification]{\includegraphics[width=.65\linewidth]{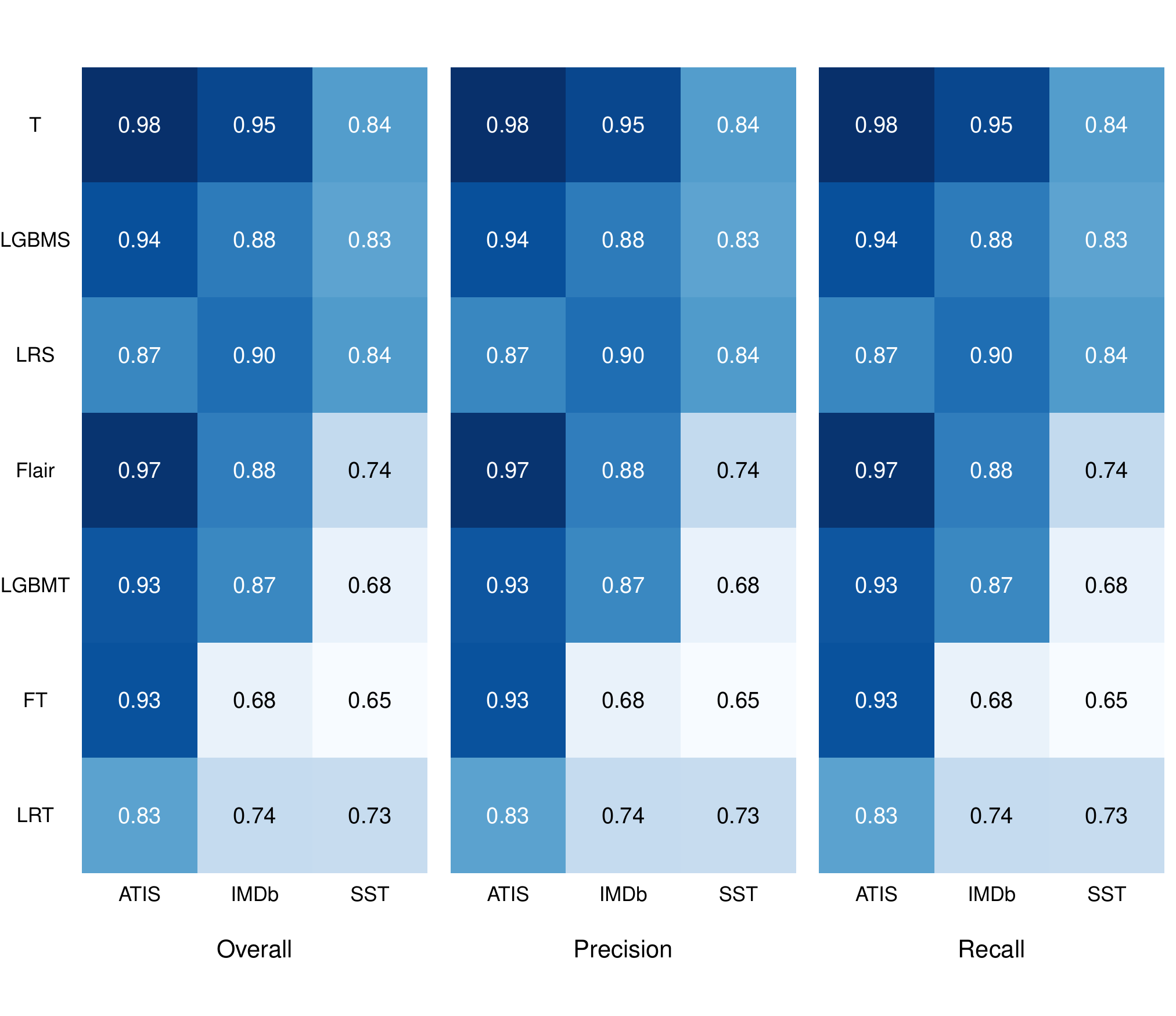}}
\end{figure}
\begin{figure}[h]\ContinuedFloat
    \begin{subfigure}{.99\textwidth}
        \centering
        \subfloat[Token labeling]{\includegraphics[height=.17\textheight]{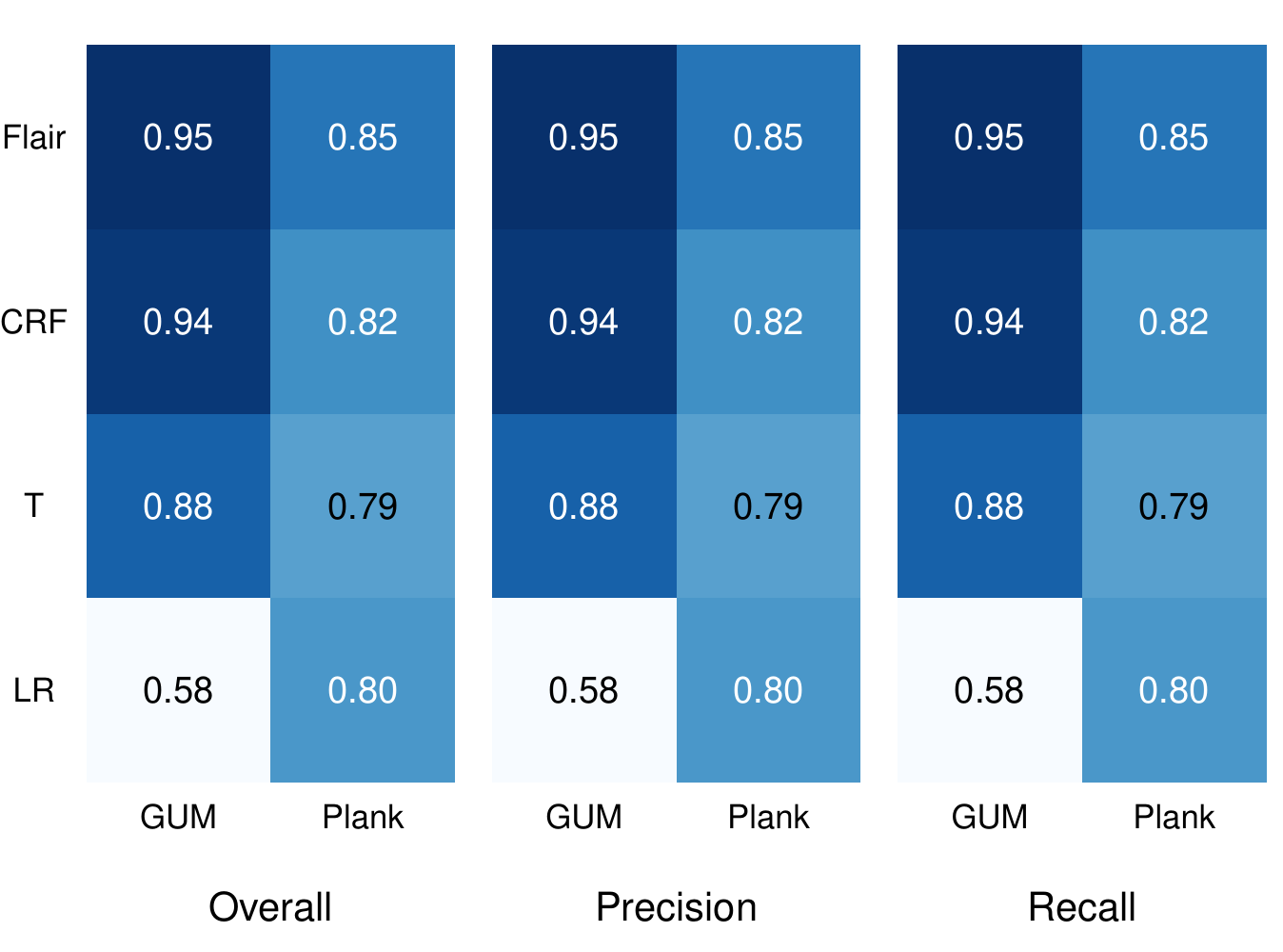}} \\
        \subfloat[Sequence labeling]{\includegraphics[height=.17\textheight]{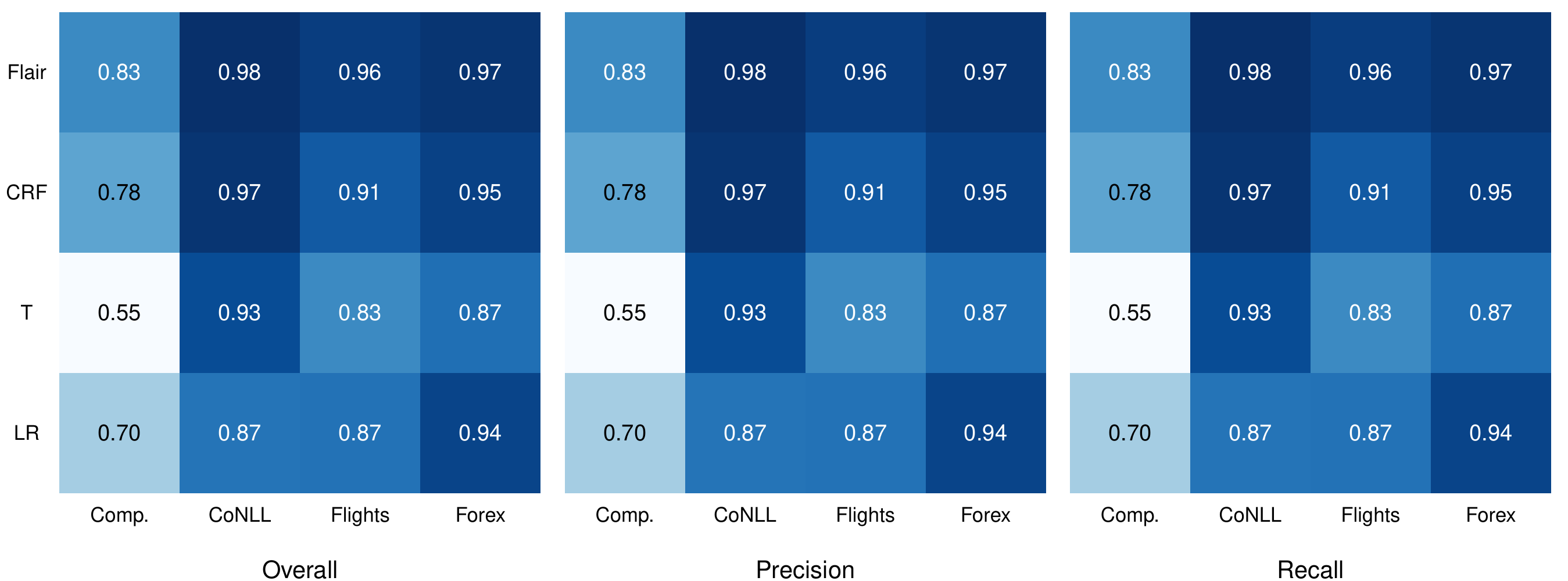}}
    \end{subfigure}
	\caption{Model performances across tasks and datasets. The model axis is ordered in descending order by the respective models' overall performance via Borda Count. }
\label{fig:appendix_r2_model_performance_heatmap}
\end{figure}

\begin{figure}
\begingroup
\begin{subfigure}{.5\textwidth}
    \centering
    \scalebox{.5}{
        \begingroup
        \renewcommand*{\arraystretch}{1.2}
        \begin{minipage}[b]{.96\linewidth}
            
\begin{tabular}{ccrrrr}
\toprule
{C} & {M} & {P} & {R} & {F1} & {\%F}\\
\midrule
 & CL & \cellcolor[HTML]{FFD6AC}{\textcolor{black}{0.43}} & \cellcolor[HTML]{FFE3C6}{\textcolor{black}{0.30}} & \cellcolor[HTML]{FFDEBB}{\textcolor{black}{0.35}} & \cellcolor[HTML]{FFFCF9}{\textcolor{black}{0.03}}\\

 & DE & \cellcolor[HTML]{FECB95}{\textcolor{black}{0.56}} & \cellcolor[HTML]{F1A340}{\textcolor{black}{1.00}} & \cellcolor[HTML]{FBBC77}{\textcolor{black}{0.72}} & \cellcolor[HTML]{FFF7EE}{\textcolor{black}{0.09}}\\

 & IRT & \cellcolor[HTML]{FFFFFF}{\textcolor{black}{0.00}} & \cellcolor[HTML]{FFFFFF}{\textcolor{black}{0.00}} & \cellcolor[HTML]{FFFFFF}{\textcolor{black}{0.00}} & \cellcolor[HTML]{F5AB53}{\textcolor{black}{0.91}}\\

 & LA & \cellcolor[HTML]{FBBD79}{\textcolor{black}{0.71}} & \cellcolor[HTML]{F1A340}{\textcolor{black}{1.00}} & \cellcolor[HTML]{F7B262}{\textcolor{black}{0.83}} & \cellcolor[HTML]{FFF9F2}{\textcolor{black}{0.07}}\\

 & PE & \cellcolor[HTML]{FFDCB8}{\textcolor{black}{0.37}} & \cellcolor[HTML]{F1A340}{\textcolor{black}{1.00}} & \cellcolor[HTML]{FFCD98}{\textcolor{black}{0.54}} & \cellcolor[HTML]{FFF3E6}{\textcolor{black}{0.13}}\\

\multirow{-6}{*}{\centering\arraybackslash ATIS} & RE & \cellcolor[HTML]{FCC07F}{\textcolor{black}{0.67}} & \cellcolor[HTML]{F1A340}{\textcolor{black}{1.00}} & \cellcolor[HTML]{F8B467}{\textcolor{black}{0.81}} & \cellcolor[HTML]{FFF8F1}{\textcolor{black}{0.07}}\\
\addlinespace
 & CL & \cellcolor[HTML]{FFE9D3}{\textcolor{black}{0.23}} & \cellcolor[HTML]{FDC488}{\textcolor{black}{0.63}} & \cellcolor[HTML]{FFDFBF}{\textcolor{black}{0.33}} & \cellcolor[HTML]{FFFAF4}{\textcolor{black}{0.06}}\\

 & DE & \cellcolor[HTML]{FFEDDB}{\textcolor{black}{0.19}} & \cellcolor[HTML]{FABB74}{\textcolor{black}{0.73}} & \cellcolor[HTML]{FFE3C6}{\textcolor{black}{0.30}} & \cellcolor[HTML]{FFF8F0}{\textcolor{black}{0.08}}\\

 & IRT & \cellcolor[HTML]{FFFEFE}{\textcolor{black}{0.01}} & \cellcolor[HTML]{FFE2C5}{\textcolor{black}{0.30}} & \cellcolor[HTML]{FFFEFC}{\textcolor{black}{0.01}} & \cellcolor[HTML]{F4AA4F}{\textcolor{black}{0.93}}\\

 & LA & \cellcolor[HTML]{FFEAD4}{\textcolor{black}{0.22}} & \cellcolor[HTML]{FDC386}{\textcolor{black}{0.64}} & \cellcolor[HTML]{FFE0BF}{\textcolor{black}{0.33}} & \cellcolor[HTML]{FFFAF4}{\textcolor{black}{0.06}}\\

 & PE & \cellcolor[HTML]{FFF5EB}{\textcolor{black}{0.10}} & \cellcolor[HTML]{FDC68A}{\textcolor{black}{0.62}} & \cellcolor[HTML]{FFEEDD}{\textcolor{black}{0.18}} & \cellcolor[HTML]{FFF4E8}{\textcolor{black}{0.12}}\\

\multirow{-6}{*}{\centering\arraybackslash IMDb} & RE & \cellcolor[HTML]{FFEAD5}{\textcolor{black}{0.22}} & \cellcolor[HTML]{FDC385}{\textcolor{black}{0.64}} & \cellcolor[HTML]{FFE0C0}{\textcolor{black}{0.33}} & \cellcolor[HTML]{FFF9F4}{\textcolor{black}{0.06}}\\
\addlinespace
 & CL & \cellcolor[HTML]{FFEBD5}{\textcolor{black}{0.22}} & \cellcolor[HTML]{F8B364}{\textcolor{black}{0.82}} & \cellcolor[HTML]{FFDFBE}{\textcolor{black}{0.34}} & \cellcolor[HTML]{FFEDDB}{\textcolor{black}{0.19}}\\

 & DE & \cellcolor[HTML]{FFEBD7}{\textcolor{black}{0.21}} & \cellcolor[HTML]{F8B363}{\textcolor{black}{0.82}} & \cellcolor[HTML]{FFE0BF}{\textcolor{black}{0.33}} & \cellcolor[HTML]{FFECD9}{\textcolor{black}{0.20}}\\

 & IRT & \cellcolor[HTML]{FFFEFD}{\textcolor{black}{0.01}} & \cellcolor[HTML]{FFF0E1}{\textcolor{black}{0.16}} & \cellcolor[HTML]{FFFDFB}{\textcolor{black}{0.02}} & \cellcolor[HTML]{F8B364}{\textcolor{black}{0.82}}\\

 & LA & \cellcolor[HTML]{FFEAD4}{\textcolor{black}{0.22}} & \cellcolor[HTML]{F7B261}{\textcolor{black}{0.84}} & \cellcolor[HTML]{FFDEBC}{\textcolor{black}{0.35}} & \cellcolor[HTML]{FFEDDB}{\textcolor{black}{0.19}}\\

 & PE & \cellcolor[HTML]{FFEBD5}{\textcolor{black}{0.22}} & \cellcolor[HTML]{F7B261}{\textcolor{black}{0.84}} & \cellcolor[HTML]{FFDFBD}{\textcolor{black}{0.34}} & \cellcolor[HTML]{FFEDDA}{\textcolor{black}{0.19}}\\

\multirow{-6}{*}{\centering\arraybackslash SST} & RE & \cellcolor[HTML]{FFEBD6}{\textcolor{black}{0.21}} & \cellcolor[HTML]{F8B363}{\textcolor{black}{0.82}} & \cellcolor[HTML]{FFDFBE}{\textcolor{black}{0.34}} & \cellcolor[HTML]{FFEDDA}{\textcolor{black}{0.19}}\\
\bottomrule
\end{tabular}

        \end{minipage}%
        \hfil%
        \begin{minipage}[b]{.96\linewidth}
            
\begin{tabular}{ccrrr}
\toprule
{C} & {M} & {AP} & {P@10\%} & {R@10\%}\\
\midrule
 & BC & \cellcolor[HTML]{9B90C4}{\textcolor{black}{0.98}} & \cellcolor[HTML]{CEC8E2}{\textcolor{black}{0.48}} & \cellcolor[HTML]{998EC3}{\textcolor{black}{1.00}}\\

 & CS & \cellcolor[HTML]{9C91C5}{\textcolor{black}{0.97}} & \cellcolor[HTML]{CEC8E2}{\textcolor{black}{0.48}} & \cellcolor[HTML]{998EC3}{\textcolor{black}{1.00}}\\

 & CU & \cellcolor[HTML]{A69CCB}{\textcolor{black}{0.87}} & \cellcolor[HTML]{CEC8E2}{\textcolor{black}{0.48}} & \cellcolor[HTML]{998EC3}{\textcolor{black}{1.00}}\\

 & DM & \cellcolor[HTML]{9B90C4}{\textcolor{black}{0.98}} & \cellcolor[HTML]{CEC8E2}{\textcolor{black}{0.48}} & \cellcolor[HTML]{998EC3}{\textcolor{black}{1.00}}\\

 & DU & \cellcolor[HTML]{FAF9FC}{\textcolor{black}{0.05}} & \cellcolor[HTML]{F9F8FB}{\textcolor{black}{0.06}} & \cellcolor[HTML]{F2F0F7}{\textcolor{black}{0.13}}\\

 & KNN & \cellcolor[HTML]{F2F0F7}{\textcolor{black}{0.13}} & \cellcolor[HTML]{F2F0F7}{\textcolor{black}{0.13}} & \cellcolor[HTML]{E3E0EF}{\textcolor{black}{0.27}}\\

 & LS & \cellcolor[HTML]{A297C8}{\textcolor{black}{0.91}} & \cellcolor[HTML]{CEC8E2}{\textcolor{black}{0.48}} & \cellcolor[HTML]{998EC3}{\textcolor{black}{1.00}}\\

 & MD & \cellcolor[HTML]{F0EEF6}{\textcolor{black}{0.14}} & \cellcolor[HTML]{EEEBF5}{\textcolor{black}{0.17}} & \cellcolor[HTML]{DBD6EA}{\textcolor{black}{0.35}}\\

\multirow{-9}{*}{\centering\arraybackslash ATIS} & PM & \cellcolor[HTML]{F9F8FC}{\textcolor{black}{0.06}} & \cellcolor[HTML]{F9F8FB}{\textcolor{black}{0.06}} & \cellcolor[HTML]{F2F0F7}{\textcolor{black}{0.13}}\\
\addlinespace
 & BC & \cellcolor[HTML]{DBD6EA}{\textcolor{black}{0.35}} & \cellcolor[HTML]{F0EEF6}{\textcolor{black}{0.14}} & \cellcolor[HTML]{B7AED5}{\textcolor{black}{0.71}}\\

 & CS & \cellcolor[HTML]{E1DDED}{\textcolor{black}{0.29}} & \cellcolor[HTML]{F2F0F7}{\textcolor{black}{0.13}} & \cellcolor[HTML]{BEB5D9}{\textcolor{black}{0.64}}\\

 & CU & \cellcolor[HTML]{E2DEEE}{\textcolor{black}{0.28}} & \cellcolor[HTML]{F0EDF6}{\textcolor{black}{0.15}} & \cellcolor[HTML]{B3AAD2}{\textcolor{black}{0.74}}\\

 & DM & \cellcolor[HTML]{E5E2F0}{\textcolor{black}{0.25}} & \cellcolor[HTML]{F2F0F7}{\textcolor{black}{0.13}} & \cellcolor[HTML]{BFB7D9}{\textcolor{black}{0.63}}\\

 & DU & \cellcolor[HTML]{F9F8FB}{\textcolor{black}{0.06}} & \cellcolor[HTML]{F7F6FA}{\textcolor{black}{0.08}} & \cellcolor[HTML]{D7D2E8}{\textcolor{black}{0.39}}\\

 & KNN & \cellcolor[HTML]{FAF9FC}{\textcolor{black}{0.05}} & \cellcolor[HTML]{F9F9FC}{\textcolor{black}{0.05}} & \cellcolor[HTML]{E4E0EF}{\textcolor{black}{0.27}}\\

 & LS & \cellcolor[HTML]{DFDBEC}{\textcolor{black}{0.31}} & \cellcolor[HTML]{F1EFF7}{\textcolor{black}{0.13}} & \cellcolor[HTML]{BBB2D7}{\textcolor{black}{0.67}}\\

 & MD & \cellcolor[HTML]{FCFBFD}{\textcolor{black}{0.03}} & \cellcolor[HTML]{FBFBFD}{\textcolor{black}{0.04}} & \cellcolor[HTML]{EDEAF4}{\textcolor{black}{0.18}}\\

\multirow{-9}{*}{\centering\arraybackslash IMDb} & PM & \cellcolor[HTML]{FAF9FC}{\textcolor{black}{0.05}} & \cellcolor[HTML]{F8F7FB}{\textcolor{black}{0.07}} & \cellcolor[HTML]{DBD6EA}{\textcolor{black}{0.35}}\\
\addlinespace
 & BC & \cellcolor[HTML]{CCC5E1}{\textcolor{black}{0.50}} & \cellcolor[HTML]{D9D4E9}{\textcolor{black}{0.37}} & \cellcolor[HTML]{B4ABD3}{\textcolor{black}{0.74}}\\

 & CS & \cellcolor[HTML]{E9E6F2}{\textcolor{black}{0.21}} & \cellcolor[HTML]{E4E0EF}{\textcolor{black}{0.27}} & \cellcolor[HTML]{C8C1DF}{\textcolor{black}{0.54}}\\

 & CU & \cellcolor[HTML]{E3DFEF}{\textcolor{black}{0.27}} & \cellcolor[HTML]{E1DDED}{\textcolor{black}{0.29}} & \cellcolor[HTML]{C3BBDC}{\textcolor{black}{0.59}}\\

 & DM & \cellcolor[HTML]{CCC6E1}{\textcolor{black}{0.49}} & \cellcolor[HTML]{DDD8EB}{\textcolor{black}{0.33}} & \cellcolor[HTML]{BBB2D7}{\textcolor{black}{0.67}}\\

 & DU & \cellcolor[HTML]{FAF9FC}{\textcolor{black}{0.05}} & \cellcolor[HTML]{FBFAFC}{\textcolor{black}{0.04}} & \cellcolor[HTML]{F6F5FA}{\textcolor{black}{0.09}}\\

 & KNN & \cellcolor[HTML]{F4F2F8}{\textcolor{black}{0.11}} & \cellcolor[HTML]{F4F2F9}{\textcolor{black}{0.11}} & \cellcolor[HTML]{E9E6F2}{\textcolor{black}{0.22}}\\

 & LS & \cellcolor[HTML]{D0C9E3}{\textcolor{black}{0.46}} & \cellcolor[HTML]{DED9EB}{\textcolor{black}{0.33}} & \cellcolor[HTML]{BCB4D8}{\textcolor{black}{0.65}}\\

 & MD & \cellcolor[HTML]{F6F5FA}{\textcolor{black}{0.08}} & \cellcolor[HTML]{F5F3F9}{\textcolor{black}{0.10}} & \cellcolor[HTML]{EAE8F3}{\textcolor{black}{0.20}}\\

\multirow{-9}{*}{\centering\arraybackslash SST} & PM & \cellcolor[HTML]{FAF9FC}{\textcolor{black}{0.05}} & \cellcolor[HTML]{FAF9FC}{\textcolor{black}{0.05}} & \cellcolor[HTML]{F5F4F9}{\textcolor{black}{0.10}}\\
\bottomrule
\end{tabular}

        \end{minipage}%
        \endgroup}
    \caption{Text classification}
\end{subfigure}%
\hfill
\begin{subfigure}{.5\textwidth}
    \centering
    \scalebox{.5}{
        \begingroup
        \renewcommand*{\arraystretch}{1.2}
        \begin{minipage}[b]{.96\linewidth}
            
\begin{tabular}{ccrrrr}
\toprule
{C} & {M} & {P} & {R} & {F1} & {\%F}\\
\midrule
 & CL & \cellcolor[HTML]{F7B261}{\textcolor{black}{0.83}} & \cellcolor[HTML]{FFDEBB}{\textcolor{black}{0.35}} & \cellcolor[HTML]{FFD0A0}{\textcolor{black}{0.50}} & \cellcolor[HTML]{FFEEDC}{\textcolor{black}{0.18}}\\

 & DE & \cellcolor[HTML]{FECA93}{\textcolor{black}{0.57}} & \cellcolor[HTML]{FEC991}{\textcolor{black}{0.58}} & \cellcolor[HTML]{FEC992}{\textcolor{black}{0.57}} & \cellcolor[HTML]{FFD7AD}{\textcolor{black}{0.43}}\\

 & IRT & \cellcolor[HTML]{FFE1C3}{\textcolor{black}{0.32}} & \cellcolor[HTML]{FEC890}{\textcolor{black}{0.59}} & \cellcolor[HTML]{FFD8B1}{\textcolor{black}{0.41}} & \cellcolor[HTML]{F9B66A}{\textcolor{black}{0.79}}\\

 & LA & \cellcolor[HTML]{F9B76C}{\textcolor{black}{0.78}} & \cellcolor[HTML]{FFD2A3}{\textcolor{black}{0.48}} & \cellcolor[HTML]{FEC88E}{\textcolor{black}{0.59}} & \cellcolor[HTML]{FFE6CD}{\textcolor{black}{0.26}}\\

 & PE & \cellcolor[HTML]{FEC88E}{\textcolor{black}{0.59}} & \cellcolor[HTML]{FFCE9A}{\textcolor{black}{0.53}} & \cellcolor[HTML]{FECB95}{\textcolor{black}{0.56}} & \cellcolor[HTML]{FFDCB7}{\textcolor{black}{0.38}}\\

 & RE & \cellcolor[HTML]{F8B467}{\textcolor{black}{0.80}} & \cellcolor[HTML]{FFCE9B}{\textcolor{black}{0.53}} & \cellcolor[HTML]{FDC486}{\textcolor{black}{0.64}} & \cellcolor[HTML]{FFE5CA}{\textcolor{black}{0.28}}\\

\multirow{-7}{*}{\centering\arraybackslash Comp.} & VN & \cellcolor[HTML]{FBBE7C}{\textcolor{black}{0.69}} & \cellcolor[HTML]{FFF9F3}{\textcolor{black}{0.06}} & \cellcolor[HTML]{FFF4E9}{\textcolor{black}{0.11}} & \cellcolor[HTML]{FFFBF8}{\textcolor{black}{0.04}}\\
\addlinespace
 & CL & \cellcolor[HTML]{FFDBB5}{\textcolor{black}{0.39}} & \cellcolor[HTML]{FFEEDD}{\textcolor{black}{0.18}} & \cellcolor[HTML]{FFE8D0}{\textcolor{black}{0.24}} & \cellcolor[HTML]{FFFDFA}{\textcolor{black}{0.02}}\\

 & DE & \cellcolor[HTML]{FFE6CC}{\textcolor{black}{0.26}} & \cellcolor[HTML]{FFE2C5}{\textcolor{black}{0.30}} & \cellcolor[HTML]{FFE4C9}{\textcolor{black}{0.28}} & \cellcolor[HTML]{FFF9F3}{\textcolor{black}{0.06}}\\

 & IRT & \cellcolor[HTML]{FFE6CB}{\textcolor{black}{0.27}} & \cellcolor[HTML]{FFE1C3}{\textcolor{black}{0.31}} & \cellcolor[HTML]{FFE4C7}{\textcolor{black}{0.29}} & \cellcolor[HTML]{FFF9F3}{\textcolor{black}{0.06}}\\

 & LA & \cellcolor[HTML]{FFE4C8}{\textcolor{black}{0.29}} & \cellcolor[HTML]{FFE1C3}{\textcolor{black}{0.31}} & \cellcolor[HTML]{FFE3C5}{\textcolor{black}{0.30}} & \cellcolor[HTML]{FFF9F4}{\textcolor{black}{0.06}}\\

 & PE & \cellcolor[HTML]{FFEFDF}{\textcolor{black}{0.17}} & \cellcolor[HTML]{FFD3A6}{\textcolor{black}{0.47}} & \cellcolor[HTML]{FFE8D0}{\textcolor{black}{0.25}} & \cellcolor[HTML]{FFF1E2}{\textcolor{black}{0.15}}\\

 & RE & \cellcolor[HTML]{FFE2C5}{\textcolor{black}{0.30}} & \cellcolor[HTML]{FFE0C0}{\textcolor{black}{0.33}} & \cellcolor[HTML]{FFE1C2}{\textcolor{black}{0.32}} & \cellcolor[HTML]{FFFAF4}{\textcolor{black}{0.06}}\\

\multirow{-7}{*}{\centering\arraybackslash CoNLL} & VN & \cellcolor[HTML]{FEC78D}{\textcolor{black}{0.60}} & \cellcolor[HTML]{FFFEFD}{\textcolor{black}{0.01}} & \cellcolor[HTML]{FFFDFB}{\textcolor{black}{0.02}} & \cellcolor[HTML]{FFFFFF}{\textcolor{black}{0.00}}\\
\addlinespace
 & CL & \cellcolor[HTML]{F4AA51}{\textcolor{black}{0.92}} & \cellcolor[HTML]{FFE6CB}{\textcolor{black}{0.27}} & \cellcolor[HTML]{FFD8B0}{\textcolor{black}{0.42}} & \cellcolor[HTML]{FFFEFD}{\textcolor{black}{0.01}}\\

 & DE & \cellcolor[HTML]{FFD8B0}{\textcolor{black}{0.41}} & \cellcolor[HTML]{F8B467}{\textcolor{black}{0.80}} & \cellcolor[HTML]{FFCC97}{\textcolor{black}{0.55}} & \cellcolor[HTML]{FFF8F0}{\textcolor{black}{0.07}}\\

 & IRT & \cellcolor[HTML]{FFFEFD}{\textcolor{black}{0.01}} & \cellcolor[HTML]{FFEDD9}{\textcolor{black}{0.20}} & \cellcolor[HTML]{FFFEFC}{\textcolor{black}{0.02}} & \cellcolor[HTML]{F4A94E}{\textcolor{black}{0.93}}\\

 & LA & \cellcolor[HTML]{FEC78D}{\textcolor{black}{0.60}} & \cellcolor[HTML]{FABB74}{\textcolor{black}{0.73}} & \cellcolor[HTML]{FCC282}{\textcolor{black}{0.66}} & \cellcolor[HTML]{FFFBF6}{\textcolor{black}{0.05}}\\

 & PE & \cellcolor[HTML]{FFEEDB}{\textcolor{black}{0.18}} & \cellcolor[HTML]{FCBF7E}{\textcolor{black}{0.68}} & \cellcolor[HTML]{FFE4C7}{\textcolor{black}{0.29}} & \cellcolor[HTML]{FFF2E4}{\textcolor{black}{0.14}}\\

 & RE & \cellcolor[HTML]{FDC68B}{\textcolor{black}{0.61}} & \cellcolor[HTML]{FABB74}{\textcolor{black}{0.73}} & \cellcolor[HTML]{FCC181}{\textcolor{black}{0.67}} & \cellcolor[HTML]{FFFBF6}{\textcolor{black}{0.05}}\\

\multirow{-7}{*}{\centering\arraybackslash Flights} & VN & \cellcolor[HTML]{F1A340}{\textcolor{black}{1.00}} & \cellcolor[HTML]{FFEFDE}{\textcolor{black}{0.17}} & \cellcolor[HTML]{FFE3C7}{\textcolor{black}{0.29}} & \cellcolor[HTML]{FFFEFE}{\textcolor{black}{0.01}}\\
\addlinespace
 & CL & \cellcolor[HTML]{FABB75}{\textcolor{black}{0.73}} & \cellcolor[HTML]{FFD4A7}{\textcolor{black}{0.46}} & \cellcolor[HTML]{FECA93}{\textcolor{black}{0.57}} & \cellcolor[HTML]{FFF9F3}{\textcolor{black}{0.06}}\\

 & DE & \cellcolor[HTML]{FFCE9B}{\textcolor{black}{0.52}} & \cellcolor[HTML]{F8B465}{\textcolor{black}{0.81}} & \cellcolor[HTML]{FDC486}{\textcolor{black}{0.64}} & \cellcolor[HTML]{FFF0E1}{\textcolor{black}{0.16}}\\

 & IRT & \cellcolor[HTML]{FFD1A2}{\textcolor{black}{0.49}} & \cellcolor[HTML]{F6B05D}{\textcolor{black}{0.85}} & \cellcolor[HTML]{FDC589}{\textcolor{black}{0.62}} & \cellcolor[HTML]{FFEEDD}{\textcolor{black}{0.18}}\\

 & LA & \cellcolor[HTML]{FCC283}{\textcolor{black}{0.65}} & \cellcolor[HTML]{FAB970}{\textcolor{black}{0.76}} & \cellcolor[HTML]{FBBE7A}{\textcolor{black}{0.70}} & \cellcolor[HTML]{FFF4E8}{\textcolor{black}{0.12}}\\

 & PE & \cellcolor[HTML]{FFD4A8}{\textcolor{black}{0.45}} & \cellcolor[HTML]{FBBC78}{\textcolor{black}{0.72}} & \cellcolor[HTML]{FECB95}{\textcolor{black}{0.56}} & \cellcolor[HTML]{FFF0E0}{\textcolor{black}{0.16}}\\

 & RE & \cellcolor[HTML]{FCC181}{\textcolor{black}{0.67}} & \cellcolor[HTML]{FABA71}{\textcolor{black}{0.75}} & \cellcolor[HTML]{FBBD7A}{\textcolor{black}{0.70}} & \cellcolor[HTML]{FFF4E9}{\textcolor{black}{0.11}}\\

\multirow{-7}{*}{\centering\arraybackslash Forex} & VN & \cellcolor[HTML]{F1A340}{\textcolor{black}{1.00}} & \cellcolor[HTML]{FFF8F1}{\textcolor{black}{0.07}} & \cellcolor[HTML]{FFF2E5}{\textcolor{black}{0.14}} & \cellcolor[HTML]{FFFEFE}{\textcolor{black}{0.01}}\\
\bottomrule
\end{tabular}

        \end{minipage}%
        \begin{minipage}[b]{.96\linewidth}
            
\begin{tabular}{ccrrr}
\toprule
{C} & {M} & {AP} & {P@10\%} & {R@10\%}\\
\midrule
 & BC & \cellcolor[HTML]{B9B0D6}{\textcolor{black}{0.68}} & \cellcolor[HTML]{ABA1CD}{\textcolor{black}{0.83}} & \cellcolor[HTML]{EBE8F3}{\textcolor{black}{0.20}}\\

 & CU & \cellcolor[HTML]{B8AFD5}{\textcolor{black}{0.70}} & \cellcolor[HTML]{A69CCB}{\textcolor{black}{0.87}} & \cellcolor[HTML]{EAE7F3}{\textcolor{black}{0.21}}\\

 & DM & \cellcolor[HTML]{BBB3D7}{\textcolor{black}{0.66}} & \cellcolor[HTML]{ABA2CE}{\textcolor{black}{0.82}} & \cellcolor[HTML]{EBE8F3}{\textcolor{black}{0.19}}\\

 & DU & \cellcolor[HTML]{D3CDE5}{\textcolor{black}{0.43}} & \cellcolor[HTML]{CCC5E1}{\textcolor{black}{0.50}} & \cellcolor[HTML]{F3F1F8}{\textcolor{black}{0.12}}\\

 & KNN & \cellcolor[HTML]{C1B9DA}{\textcolor{black}{0.61}} & \cellcolor[HTML]{B2A9D2}{\textcolor{black}{0.75}} & \cellcolor[HTML]{EDEAF4}{\textcolor{black}{0.18}}\\

 & LE & \cellcolor[HTML]{D5D0E7}{\textcolor{black}{0.41}} & \cellcolor[HTML]{D8D3E8}{\textcolor{black}{0.38}} & \cellcolor[HTML]{F6F4FA}{\textcolor{black}{0.09}}\\

 & MD & \cellcolor[HTML]{C8C1DF}{\textcolor{black}{0.54}} & \cellcolor[HTML]{C3BBDC}{\textcolor{black}{0.59}} & \cellcolor[HTML]{F1EEF7}{\textcolor{black}{0.14}}\\

 & PM & \cellcolor[HTML]{C8C1DF}{\textcolor{black}{0.54}} & \cellcolor[HTML]{BDB5D8}{\textcolor{black}{0.64}} & \cellcolor[HTML]{EFEDF6}{\textcolor{black}{0.15}}\\

\multirow{-9}{*}{\centering\arraybackslash Comp.} & WD & \cellcolor[HTML]{D1CBE4}{\textcolor{black}{0.45}} & \cellcolor[HTML]{CEC8E2}{\textcolor{black}{0.48}} & \cellcolor[HTML]{F3F2F8}{\textcolor{black}{0.11}}\\
\addlinespace
 & BC & \cellcolor[HTML]{F0EEF6}{\textcolor{black}{0.14}} & \cellcolor[HTML]{F2F0F7}{\textcolor{black}{0.13}} & \cellcolor[HTML]{E6E3F0}{\textcolor{black}{0.24}}\\

 & CU & \cellcolor[HTML]{EEECF5}{\textcolor{black}{0.17}} & \cellcolor[HTML]{ECE9F4}{\textcolor{black}{0.18}} & \cellcolor[HTML]{DCD7EA}{\textcolor{black}{0.34}}\\

 & DM & \cellcolor[HTML]{F1EFF7}{\textcolor{black}{0.14}} & \cellcolor[HTML]{F2F1F8}{\textcolor{black}{0.12}} & \cellcolor[HTML]{E8E4F1}{\textcolor{black}{0.23}}\\

 & DU & \cellcolor[HTML]{F8F7FB}{\textcolor{black}{0.07}} & \cellcolor[HTML]{F7F6FA}{\textcolor{black}{0.08}} & \cellcolor[HTML]{F0EDF6}{\textcolor{black}{0.15}}\\

 & KNN & \cellcolor[HTML]{F3F1F8}{\textcolor{black}{0.12}} & \cellcolor[HTML]{F2F0F7}{\textcolor{black}{0.13}} & \cellcolor[HTML]{E7E3F1}{\textcolor{black}{0.24}}\\

 & LE & \cellcolor[HTML]{ECE9F4}{\textcolor{black}{0.19}} & \cellcolor[HTML]{EDEBF5}{\textcolor{black}{0.17}} & \cellcolor[HTML]{DED9EC}{\textcolor{black}{0.32}}\\

 & MD & \cellcolor[HTML]{F9F8FB}{\textcolor{black}{0.06}} & \cellcolor[HTML]{FAF9FC}{\textcolor{black}{0.05}} & \cellcolor[HTML]{F6F4F9}{\textcolor{black}{0.09}}\\

 & PM & \cellcolor[HTML]{F9F8FB}{\textcolor{black}{0.06}} & \cellcolor[HTML]{F7F6FB}{\textcolor{black}{0.07}} & \cellcolor[HTML]{F1EFF7}{\textcolor{black}{0.14}}\\

\multirow{-9}{*}{\centering\arraybackslash CoNLL} & WD & \cellcolor[HTML]{EEECF5}{\textcolor{black}{0.16}} & \cellcolor[HTML]{EDEBF5}{\textcolor{black}{0.17}} & \cellcolor[HTML]{DED9EC}{\textcolor{black}{0.32}}\\
\addlinespace
 & BC & \cellcolor[HTML]{CDC6E2}{\textcolor{black}{0.49}} & \cellcolor[HTML]{E6E3F0}{\textcolor{black}{0.24}} & \cellcolor[HTML]{BEB6D9}{\textcolor{black}{0.63}}\\

 & CU & \cellcolor[HTML]{B9B0D6}{\textcolor{black}{0.68}} & \cellcolor[HTML]{E1DDEE}{\textcolor{black}{0.29}} & \cellcolor[HTML]{B2A8D2}{\textcolor{black}{0.76}}\\

 & DM & \cellcolor[HTML]{DBD6EA}{\textcolor{black}{0.35}} & \cellcolor[HTML]{E5E1F0}{\textcolor{black}{0.25}} & \cellcolor[HTML]{BCB3D7}{\textcolor{black}{0.66}}\\

 & DU & \cellcolor[HTML]{ECE9F4}{\textcolor{black}{0.18}} & \cellcolor[HTML]{F4F3F9}{\textcolor{black}{0.10}} & \cellcolor[HTML]{E3E0EF}{\textcolor{black}{0.27}}\\

 & KNN & \cellcolor[HTML]{F7F6FB}{\textcolor{black}{0.07}} & \cellcolor[HTML]{F7F6FB}{\textcolor{black}{0.07}} & \cellcolor[HTML]{EBE8F3}{\textcolor{black}{0.20}}\\

 & LE & \cellcolor[HTML]{F4F3F9}{\textcolor{black}{0.10}} & \cellcolor[HTML]{F4F3F9}{\textcolor{black}{0.10}} & \cellcolor[HTML]{E3E0EF}{\textcolor{black}{0.27}}\\

 & MD & \cellcolor[HTML]{F8F7FB}{\textcolor{black}{0.07}} & \cellcolor[HTML]{F3F2F8}{\textcolor{black}{0.11}} & \cellcolor[HTML]{E1DDED}{\textcolor{black}{0.29}}\\

 & PM & \cellcolor[HTML]{F3F1F8}{\textcolor{black}{0.12}} & \cellcolor[HTML]{F2F1F8}{\textcolor{black}{0.12}} & \cellcolor[HTML]{DEDAEC}{\textcolor{black}{0.32}}\\

\multirow{-9}{*}{\centering\arraybackslash Flights} & WD & \cellcolor[HTML]{F4F2F9}{\textcolor{black}{0.11}} & \cellcolor[HTML]{F2F1F8}{\textcolor{black}{0.12}} & \cellcolor[HTML]{DEDAEC}{\textcolor{black}{0.32}}\\
\addlinespace
 & BC & \cellcolor[HTML]{C8C1DF}{\textcolor{black}{0.54}} & \cellcolor[HTML]{CDC6E1}{\textcolor{black}{0.49}} & \cellcolor[HTML]{CDC6E2}{\textcolor{black}{0.49}}\\

 & CU & \cellcolor[HTML]{B7AED5}{\textcolor{black}{0.70}} & \cellcolor[HTML]{B6ADD4}{\textcolor{black}{0.71}} & \cellcolor[HTML]{B7AED5}{\textcolor{black}{0.71}}\\

 & DM & \cellcolor[HTML]{C1B9DA}{\textcolor{black}{0.61}} & \cellcolor[HTML]{BCB4D8}{\textcolor{black}{0.66}} & \cellcolor[HTML]{BDB4D8}{\textcolor{black}{0.65}}\\

 & DU & \cellcolor[HTML]{DED9EC}{\textcolor{black}{0.32}} & \cellcolor[HTML]{DEDAEC}{\textcolor{black}{0.32}} & \cellcolor[HTML]{DEDAEC}{\textcolor{black}{0.32}}\\

 & KNN & \cellcolor[HTML]{EEECF5}{\textcolor{black}{0.16}} & \cellcolor[HTML]{F1EFF7}{\textcolor{black}{0.14}} & \cellcolor[HTML]{F1EFF7}{\textcolor{black}{0.14}}\\

 & LE & \cellcolor[HTML]{F3F2F8}{\textcolor{black}{0.11}} & \cellcolor[HTML]{F6F4FA}{\textcolor{black}{0.09}} & \cellcolor[HTML]{F6F4FA}{\textcolor{black}{0.09}}\\

 & MD & \cellcolor[HTML]{F1EFF7}{\textcolor{black}{0.14}} & \cellcolor[HTML]{EAE7F3}{\textcolor{black}{0.20}} & \cellcolor[HTML]{EAE7F3}{\textcolor{black}{0.20}}\\

 & PM & \cellcolor[HTML]{E5E2F0}{\textcolor{black}{0.25}} & \cellcolor[HTML]{E0DCED}{\textcolor{black}{0.30}} & \cellcolor[HTML]{E0DCED}{\textcolor{black}{0.30}}\\

\multirow{-9}{*}{\centering\arraybackslash Forex} & WD & \cellcolor[HTML]{F1EFF7}{\textcolor{black}{0.14}} & \cellcolor[HTML]{EAE7F3}{\textcolor{black}{0.20}} & \cellcolor[HTML]{EAE7F3}{\textcolor{black}{0.20}}\\
\bottomrule
\end{tabular}

        \end{minipage}%
        \endgroup}
    \caption{Span labeling}
\end{subfigure}\vspace{1em}
\begin{subfigure}{\textwidth}
    \centering
    \scalebox{.5}{
        \begingroup
        \renewcommand*{\arraystretch}{1.2}
        \begin{minipage}[b]{.48\linewidth}
            
\begin{tabular}{ccrrrr}
\toprule
{C} & {M} & {P} & {R} & {F1} & {\%F}\\
\midrule
 & CL & \cellcolor[HTML]{FABC76}{\textcolor{black}{0.73}} & \cellcolor[HTML]{F5AC55}{\textcolor{black}{0.90}} & \cellcolor[HTML]{F8B567}{\textcolor{black}{0.80}} & \cellcolor[HTML]{FFF9F3}{\textcolor{black}{0.06}}\\

 & DE & \cellcolor[HTML]{FEC88F}{\textcolor{black}{0.59}} & \cellcolor[HTML]{F1A340}{\textcolor{black}{1.00}} & \cellcolor[HTML]{FABA73}{\textcolor{black}{0.74}} & \cellcolor[HTML]{FFF7EF}{\textcolor{black}{0.08}}\\

 & IRT & \cellcolor[HTML]{FFFFFF}{\textcolor{black}{0.00}} & \cellcolor[HTML]{FFFFFF}{\textcolor{black}{0.00}} & \cellcolor[HTML]{FFFFFF}{\textcolor{black}{0.00}} & \cellcolor[HTML]{F4AB52}{\textcolor{black}{0.91}}\\

 & LA & \cellcolor[HTML]{FFCF9D}{\textcolor{black}{0.51}} & \cellcolor[HTML]{F1A341}{\textcolor{black}{1.00}} & \cellcolor[HTML]{FCC07E}{\textcolor{black}{0.68}} & \cellcolor[HTML]{FFF6EC}{\textcolor{black}{0.10}}\\

 & PE & \cellcolor[HTML]{FFD9B1}{\textcolor{black}{0.41}} & \cellcolor[HTML]{F1A341}{\textcolor{black}{1.00}} & \cellcolor[HTML]{FEC991}{\textcolor{black}{0.58}} & \cellcolor[HTML]{FFF3E8}{\textcolor{black}{0.12}}\\

 & RE & \cellcolor[HTML]{FFCD9A}{\textcolor{black}{0.53}} & \cellcolor[HTML]{F1A340}{\textcolor{black}{1.00}} & \cellcolor[HTML]{FBBE7C}{\textcolor{black}{0.69}} & \cellcolor[HTML]{FFF6ED}{\textcolor{black}{0.09}}\\

\multirow{-7}{*}{\centering\arraybackslash GUM} & VN & \cellcolor[HTML]{FFD2A5}{\textcolor{black}{0.47}} & \cellcolor[HTML]{FCC282}{\textcolor{black}{0.66}} & \cellcolor[HTML]{FFCB96}{\textcolor{black}{0.55}} & \cellcolor[HTML]{FFF8F2}{\textcolor{black}{0.07}}\\
\addlinespace
 & CL & \cellcolor[HTML]{FFD3A5}{\textcolor{black}{0.47}} & \cellcolor[HTML]{FFE2C4}{\textcolor{black}{0.31}} & \cellcolor[HTML]{FFDCB8}{\textcolor{black}{0.37}} & \cellcolor[HTML]{FFF8F0}{\textcolor{black}{0.08}}\\

 & DE & \cellcolor[HTML]{FFD5A9}{\textcolor{black}{0.45}} & \cellcolor[HTML]{FFCF9E}{\textcolor{black}{0.51}} & \cellcolor[HTML]{FFD2A4}{\textcolor{black}{0.48}} & \cellcolor[HTML]{FFF2E5}{\textcolor{black}{0.13}}\\

 & IRT & \cellcolor[HTML]{FFF9F2}{\textcolor{black}{0.07}} & \cellcolor[HTML]{FFD2A5}{\textcolor{black}{0.47}} & \cellcolor[HTML]{FFF4E8}{\textcolor{black}{0.12}} & \cellcolor[HTML]{F7B15F}{\textcolor{black}{0.84}}\\

 & LA & \cellcolor[HTML]{FFD4A8}{\textcolor{black}{0.46}} & \cellcolor[HTML]{FFCE9B}{\textcolor{black}{0.53}} & \cellcolor[HTML]{FFD1A2}{\textcolor{black}{0.49}} & \cellcolor[HTML]{FFF2E5}{\textcolor{black}{0.14}}\\

 & PE & \cellcolor[HTML]{FFD2A3}{\textcolor{black}{0.48}} & \cellcolor[HTML]{FFCD9A}{\textcolor{black}{0.53}} & \cellcolor[HTML]{FFD09F}{\textcolor{black}{0.50}} & \cellcolor[HTML]{FFF3E6}{\textcolor{black}{0.13}}\\

 & RE & \cellcolor[HTML]{FFD3A5}{\textcolor{black}{0.47}} & \cellcolor[HTML]{FFCE9C}{\textcolor{black}{0.52}} & \cellcolor[HTML]{FFD1A1}{\textcolor{black}{0.49}} & \cellcolor[HTML]{FFF3E6}{\textcolor{black}{0.13}}\\

\multirow{-7}{*}{\centering\arraybackslash Plank} & VN & \cellcolor[HTML]{FFCB96}{\textcolor{black}{0.55}} & \cellcolor[HTML]{FFEBD7}{\textcolor{black}{0.21}} & \cellcolor[HTML]{FFE3C5}{\textcolor{black}{0.30}} & \cellcolor[HTML]{FFFBF6}{\textcolor{black}{0.04}}\\
\bottomrule
\end{tabular}

        \end{minipage}%
        \begin{minipage}[b]{.48\linewidth}
            
\begin{tabular}{ccrrr}
\toprule
{C} & {M} & {AP} & {P@10\%} & {R@10\%}\\
\midrule
 & BC & \cellcolor[HTML]{A196C8}{\textcolor{black}{0.92}} & \cellcolor[HTML]{CFC8E3}{\textcolor{black}{0.47}} & \cellcolor[HTML]{9E93C6}{\textcolor{black}{0.95}}\\

 & CU & \cellcolor[HTML]{9B91C4}{\textcolor{black}{0.98}} & \cellcolor[HTML]{CCC6E1}{\textcolor{black}{0.50}} & \cellcolor[HTML]{998EC3}{\textcolor{black}{1.00}}\\

 & DM & \cellcolor[HTML]{9E93C6}{\textcolor{black}{0.95}} & \cellcolor[HTML]{CDC7E2}{\textcolor{black}{0.49}} & \cellcolor[HTML]{9B90C4}{\textcolor{black}{0.98}}\\

 & DU & \cellcolor[HTML]{FAF9FC}{\textcolor{black}{0.05}} & \cellcolor[HTML]{FAF9FC}{\textcolor{black}{0.05}} & \cellcolor[HTML]{F5F3F9}{\textcolor{black}{0.10}}\\

 & KNN & \cellcolor[HTML]{E9E6F2}{\textcolor{black}{0.21}} & \cellcolor[HTML]{ECE9F4}{\textcolor{black}{0.19}} & \cellcolor[HTML]{D8D3E8}{\textcolor{black}{0.38}}\\

 & LE & \cellcolor[HTML]{C2BADB}{\textcolor{black}{0.60}} & \cellcolor[HTML]{DCD7EA}{\textcolor{black}{0.34}} & \cellcolor[HTML]{B8B0D5}{\textcolor{black}{0.69}}\\

 & MD & \cellcolor[HTML]{F2F1F8}{\textcolor{black}{0.12}} & \cellcolor[HTML]{F0EEF6}{\textcolor{black}{0.14}} & \cellcolor[HTML]{E1DDEE}{\textcolor{black}{0.29}}\\

 & PM & \cellcolor[HTML]{FAF9FC}{\textcolor{black}{0.05}} & \cellcolor[HTML]{FAF9FC}{\textcolor{black}{0.05}} & \cellcolor[HTML]{F4F2F9}{\textcolor{black}{0.11}}\\

\multirow{-9}{*}{\centering\arraybackslash GUM} & WD & \cellcolor[HTML]{C8C1DF}{\textcolor{black}{0.53}} & \cellcolor[HTML]{D7D1E7}{\textcolor{black}{0.39}} & \cellcolor[HTML]{AFA5D0}{\textcolor{black}{0.79}}\\
\addlinespace
 & BC & \cellcolor[HTML]{D8D2E8}{\textcolor{black}{0.38}} & \cellcolor[HTML]{D4CEE6}{\textcolor{black}{0.42}} & \cellcolor[HTML]{DBD6EA}{\textcolor{black}{0.36}}\\

 & CU & \cellcolor[HTML]{D3CEE5}{\textcolor{black}{0.42}} & \cellcolor[HTML]{CBC4E0}{\textcolor{black}{0.51}} & \cellcolor[HTML]{D3CDE5}{\textcolor{black}{0.43}}\\

 & DM & \cellcolor[HTML]{E3DFEF}{\textcolor{black}{0.27}} & \cellcolor[HTML]{D9D4E9}{\textcolor{black}{0.37}} & \cellcolor[HTML]{DFDAEC}{\textcolor{black}{0.31}}\\

 & DU & \cellcolor[HTML]{E6E3F1}{\textcolor{black}{0.24}} & \cellcolor[HTML]{E2DEEE}{\textcolor{black}{0.28}} & \cellcolor[HTML]{E6E3F1}{\textcolor{black}{0.24}}\\

 & KNN & \cellcolor[HTML]{DFDBEC}{\textcolor{black}{0.31}} & \cellcolor[HTML]{D7D1E7}{\textcolor{black}{0.39}} & \cellcolor[HTML]{DDD8EB}{\textcolor{black}{0.33}}\\

 & LE & \cellcolor[HTML]{E8E5F2}{\textcolor{black}{0.22}} & \cellcolor[HTML]{E6E2F0}{\textcolor{black}{0.24}} & \cellcolor[HTML]{EAE7F3}{\textcolor{black}{0.21}}\\

 & MD & \cellcolor[HTML]{EFEDF6}{\textcolor{black}{0.16}} & \cellcolor[HTML]{ECE9F4}{\textcolor{black}{0.19}} & \cellcolor[HTML]{EFEDF6}{\textcolor{black}{0.16}}\\

 & PM & \cellcolor[HTML]{E7E4F1}{\textcolor{black}{0.23}} & \cellcolor[HTML]{E2DEEE}{\textcolor{black}{0.29}} & \cellcolor[HTML]{E6E3F0}{\textcolor{black}{0.24}}\\

\multirow{-9}{*}{\centering\arraybackslash Plank} & WD & \cellcolor[HTML]{D7D2E8}{\textcolor{black}{0.39}} & \cellcolor[HTML]{D3CDE5}{\textcolor{black}{0.43}} & \cellcolor[HTML]{D9D4E9}{\textcolor{black}{0.37}}\\
\bottomrule
\end{tabular}

        \end{minipage}%
        \endgroup}
    \caption{Token labeling}
\end{subfigure}
\caption{\ac{aed} results achieved with using the respective best models across all flaggers \flaggercircle and scorers \scorercircle for text classification, span and token labeling.}
\endgroup
\end{figure}
    \clearpage
    \begin{acknowledgments}
We thank Andreas R{\"u}ckl{\'e}, Edwin Simpson, Falko Helm, Ivan Habernal, Ji-Ung Lee, Michael Bugert, Nafise Sadat Moosavi, Nils Dycke,  Richard Eckart de Castilho, Tobias Mayer, and Yevgeniy Puzikov and our anonymous reviewers for the fruitful discussions and helpful feedback that improved this article.
We are especially grateful for Michael Bugert and his implementation of the span matching via linear sum assignment and Amir Zeldes, Andreas Grivas, Beatrice Alex, Hadi Amiri, Jeremy Barnes as well as Stefan Larson for answering our questions regarding their publications and for making datasets and code available.
This research work has been funded by the ``Data Analytics for the Humanities'' grant by the Hessian Ministry of Higher Education, Research, Science and the Arts.
\end{acknowledgments}

    \begin{multicols}{2}
        \bibliography{bibliography}
    \end{multicols}

\end{document}